%% file: main.tex
\documentclass[11pt,a4paper]{article}
\usepackage[T2A,T1]{fontenc}
\usepackage[utf8]{inputenc}
\usepackage[russian,main=english]{babel}
\usepackage[margin=2.5cm]{geometry}
\usepackage{amsmath,amssymb}
\usepackage{textcomp}
\usepackage{booktabs}
\usepackage{tabularx}
\usepackage{longtable}
\usepackage{rotating}
\usepackage{xurl}
\usepackage{ragged2e}
\renewcommand{\_}{\textunderscore\allowbreak}
\usepackage{graphicx}
\usepackage{microtype}
\usepackage[hidelinks,unicode]{hyperref}
\DeclareUnicodeCharacter{00B7}{\ensuremath{\cdot}}
\DeclareUnicodeCharacter{00D7}{\ensuremath{\times}}
\DeclareUnicodeCharacter{0394}{\ensuremath{\Delta}}
\DeclareUnicodeCharacter{03B1}{\ensuremath{\alpha}}
\DeclareUnicodeCharacter{2074}{\ensuremath{^{4}}}
\DeclareUnicodeCharacter{2076}{\ensuremath{^{6}}}
\DeclareUnicodeCharacter{2077}{\ensuremath{^{7}}}
\DeclareUnicodeCharacter{207B}{\ensuremath{^{-}}}
\DeclareUnicodeCharacter{2212}{\ensuremath{-}}
\DeclareUnicodeCharacter{2013}{\textendash\allowbreak}

\title{Same Origin, Different Verdicts: Text Production Conditions Confound Machine-Generated Text Detection in Russian}
\author{Stanislav Kirichenko\\
\small Independent Researcher\\
\small \texttt{staurus86@gmail.com}\\
\small ORCID: \href{https://orcid.org/0009-0001-2914-9541}{0009-0001-2914-9541}}
\date{}
\begin{document}
\maketitle
\begin{abstract}
Automated assessment of whether a text was machine-generated is applied where
the verdict carries a cost for the author, yet origin is a property of the
production process, not an observable property of the text. The conditions
under which a text was produced — how rigid the task was, the editorial rules,
the platform template — push it in the same direction in which a generative
model is presumed to push it.

This study measures the contribution of these conditions while origin is held
fixed. The Russian-language corpus of 1882 documents comprises a machine part,
generated by four generation channels — each backed by its own model family —
across 45 tasks in three task modes, and an
archival human part annotated for source regimentation level before the texts
were read. The primary contrast is computed within the machine corpus: both
halves of a pair are generated by the same model for the same task and differ
only in task wording. Machine-likeness is measured by four operationalisations
— an interpretable index with weights fixed before data collection, a trained
classifier, a zero-shot metric on an open model, and a judge model.

The main result is disagreement between procedures. Changing the task mode
shifted the scores of all four, but the direction did not match: three gave a
negative shift, the judge a positive one, and none of the four intervals
covers zero. The cross-procedure comparison plan and the specifications of
procedures 2–4 were fixed after the result of procedure 1 had been obtained
but before procedures 2–4 were run; the disagreement survived correction of a
corpus defect and of the nested validation.

Two observations explain why such scores cannot be read as measurements of
authorship. A baseline on a single feature — the source identifier — reaches a
median AUROC of 1.0000 across the thirteen group splits where AUROC is
defined, meaning that the origin
class is confounded with how the corpus was assembled. Removing emphasis
markers did not change the full text by a single character, yet it moved the
boundary between heading and prose, changed the format counters, and flipped a
confident machine verdict to human for every applicable model.

The scores should be interpreted as properties of a specific measurement
system and corpus, not as a direct measurement of text origin. The system
distinguishes not only linguistic style but traces of data collection,
repetition, formatting, and preprocessing decisions.
\end{abstract}
\par\bigskip
{\footnotesize Assembled 2026-08-02T19:40:42+00:00 following \texttt{08-\allowbreak{}paper/paper-\allowbreak{}assembly-\allowbreak{}manifest.md}. Sources verified against their hashes; the exclusions of manifest §2 applied. English translation of the frozen Russian manuscript (sha256 \texttt{3c9f8fadb68e82175d93cb688c264ecebb967f1d9fe999ab235dcd4c4f14a34c}), prepared 2026-08-02; the Russian text remains the source of record.\par}

\section{Introduction}

\subsection{Why this matters}

Automated assessment of text origin is applied where the decision carries a cost:
checking student work, moderating platforms, screening contractors, editorial
policies. A verdict of "machine-written" carries consequences for the author, and
so the question of what exactly a detector measures stops being a technical one.

The cost of error is asymmetric. A missed machine text is cheap; a false
accusation against a human is expensive, and the author cannot defend against it:
proving one's own authorship is harder than disputing it. In the accusatory scenario,
therefore, what decides is not average accuracy but the detector's behaviour at a
low false positive rate — and on this characteristic systems diverge dramatically.

\subsection{Machine-likeness is not directly observable}

The origin of a text is a property of the process, not of the text itself. We observe
only the result, and any machine-likeness score is an operationalisation: a set of
features, a way of weighting them, a decision threshold.

An uncomfortable consequence follows: different operationalisations may diverge,
and this is not necessarily an error on the part of either. Until they are shown to
measure the same thing, the question "which detector is right" is posed before it
can be answered.

This property of the process depends not only on who wrote the text but also on the
conditions under which it was written. A rigid technical specification narrows the
choice of wording; editorial regimentation standardises the structure; a platform
template dictates the formatting. All three push the text in the same direction in
which the machine is presumed to act.

\subsection{How authorship is confounded in practice}

A corpus for detector evaluation is almost always assembled so that the origin class
coincides with something else.

Human texts are taken from platforms, machine texts are generated through channels —
and the source becomes a proxy for the class. Human texts come from genres where one
regimentation regime applies, machine texts are written to tasks of a different kind —
and the genre becomes a proxy for the class. Human texts keep the publication's
formatting, machine texts arrive from an API without it — and the formatting becomes
a proxy for the class. Finally, both classes pass through a data preparation
pipeline, and its decisions — what counts as a heading, what counts as
prose — change the model input without changing a single word.

Each of these confounds is known in isolation. What happens to a machine-likeness
score when production conditions change while origin stays the same is a separate
question, and it is rarely asked.

\subsection{What the literature review established}

A systematic sample of 73 papers shows that studies predominantly evaluate detector
accuracy and robustness to editing, while the human side of the comparison
is assembled opportunistically, without control over the task, the regimentation, or
the production channel.

In this sample, no studies were found that separately measure the effect of
production conditions on machine-likeness scores with origin held fixed. Nor were
any estimates found of the false positive rate for formal and translated
Russian-language prose at the same time. Both claims are narrowed to the scope of
the search performed.

\subsection{Design}

The study compares production conditions under fixed tasks.

The machine part of the corpus was generated by four generation channels — each
backed by its own model family — on 45 tasks in three modes: from a neutral
technical specification to a task requiring the avoidance of typical
machine-text features, with two replicates. The main contrast is computed within the machine corpus: both
halves of a pair are generated by the same model for the same task, and only the
formulation differs. Origin does not change; the conditions do.

The human part consists of archival texts published before 2022, annotated by
source regimentation level before the texts themselves were read. It includes the
stratum on which the cost of error is measured: formal academic prose and translations.

Machine-likeness is measured by four procedures with different inputs and different
provenance of weights: an interpretable index with weights fixed before data
collection; a trained classifier on the same feature matrix with group splits; a
zero-shot metric on an open Russian-language model; a judge model with a rubric.
None is declared the gold standard.

The design, hypotheses, and analysis rules were fixed by preregistration before the
dependent variable was computed; deviations are published as amendments with dates
and reasons.

\subsection{Research questions}

\begin{enumerate}
\item[1.] Do text production conditions change its machine-likeness score under a fixed
   task?
\item[2.] Do the direction and magnitude of the effect agree across the four
   operationalisations?
\item[3.] How strongly do the results depend on the source, genre, register, formatting,
   and transformations of the text?
\end{enumerate}

\subsection{Contributions}

The study delivers four things.

\textbf{A measurement of the effect of conditions with origin held fixed} — on a
Russian-language corpus, with a fixed task and a preregistered contrast.

\textbf{A test of the agreement of the four operationalisations} instead of an
assumption of it, with an explicit rule: matching signs count as convergence of
operationalisations, not as independent replication.

\textbf{An assessment of how the result depends on conditions that usually stay in the
background} — the source, genre, formal register, formatting, and ten
transformations of the text, including cases where the decision flips while the
text remains unchanged.

\textbf{Recorded limitations of transferability} — which of the reported numbers do not
transfer to applications where both classes come from a single source, and why.

The study does not propose a new detector and does not improve existing ones. Its
subject is what the procedures already in use actually measure when text production
conditions change while origin remains the same.

\section{Research gap}

\subsection{1. What the literature measures}

Research predominantly evaluates detector accuracy and robustness to editing.

Two findings in this area are firmly established. The first: a universal binary
detector transfers poorly to new models, domains, and attacks — RAID assembled six
million generations across eleven models and eight domains and showed performance
collapsing under a change of decoding strategy and under adversarial attacks (Dugan et al., 2024); on
unfamiliar domains and generators, TPR at 1\% FPR falls as low as zero (Tufts et al., 2025); M4
(Wang et al., 2024a) and MULTITuDE across eleven languages (Macko et al., 2023) give the same result. The second:
average accuracy is unfit for the accusatory scenario — in an audit of four
detectors on 1992 excerpts, open systems mistook up to 78\% of human text for
machine text, and at an FPR ceiling of 0.5\% only one commercial system retained
detection (Jabarian and Imas, 2025).

Borile and Abrate, 2025 name the cause of poor transfer: linguistic and domain confounders
create spurious correlations that do not carry beyond the training distribution.
The question of which specific conditions give rise to these correlations is not posed there.

\subsection{2. How the human side of the comparison is assembled}

The human reference is often collected opportunistically — from whatever is available —
without control over the task, the regimentation, or the production channel.

This is not a complaint about individual studies but a consequence of how the field
is set up: to measure a detector, any corpus of demonstrably human texts will do, and
ready-made collections are the most convenient choice. Under such selection, the
conditions in which these texts were written are not recorded and do not enter the analysis.

The cost is visible in Fredrick and Craven, 2025: the direction of differences in lexical diversity changes
with the composition of the human control group. If a machine-likeness score
depends on the conditions under which the reference was written, then a threshold
calibrated on one corpus carries over to another with no guarantees.

\subsection{3. What the sample lacks}

\textbf{Within the included sample of 73 studies, no research was found that separately
assessed how text production conditions — task mode, regimentation level,
and generation channel — change a text's machine-likeness scores
while its origin remains unchanged.}

Three groups of studies come close, and each stops short. Studies of
editing prevalence count how much text has passed through a model (Liang et al., 2024,
Kobak et al., 2025) but do not ask how much machine-likeness the task itself creates. Work on
hybrid authorship varies the mode of editing (Artemova et al., 2025, Su et al., 2025, Zeng et al., 2024, Baumler et al., 2026) — yet
the human reference in these studies is written freely, without regimentation. Fairness studies
show that the score varies across subgroups of writers (Liang et al., 2023, Stowe et al., 2026, Basu et al., 2026), not across the
conditions under which the text was produced.

CoAT comes closest — a Russian-language corpus of 246 thousand texts and
thirteen generators (Shamardina et al., 2024). It shows directly that the mode of production changes
detectability: machine-translated texts are recognised with an accuracy of 0.621, while
free generation is recognised with 0.99. The study stops
at the same point as the others: the generator changes together with the task type,
so the effect of conditions is not separated from the effect of origin, and the human
side is drawn from the same domains without control over task or regimentation.

Close by stands the study by Gritsay and colleagues (Gritsay et al., 2023): detecting machine-generated fragments in
Russian-language scientific documents yields an F1 of 0.968 with multi-domain training and
paraphrase augmentation, against 0.825 in English. The authors conclude outright
that a universal detector for all domains and languages is unlikely to be achievable — a conclusion
that matches ours in substance and was obtained on different data. There, too, production
conditions are not varied with origin held unchanged.

To this is added a language gap. The matrix contains five Russian-language primary studies,
and none varies production conditions with origin unchanged: RuATD
collects generators (Shamardina et al., 2022), LLMTrace annotates mixed authorship character by character
(Tolstykh et al., 2025), AINL-Eval measures transfer to unseen scientific fields (Batura et al., 2025), CoAT
compares generative task types together with a change of generator (Shamardina et al., 2024), and the study
by Gritsay and colleagues measures transfer between domains of scientific documents (Gritsay et al., 2023).

\textbf{Among the included studies, no FPR estimates were found that cover
formal and translated Russian-language prose at the same time.}

Both claims are worded to stay within the scope of the search performed: they speak
of what is absent from the included sample, not of what is absent from the literature at large.

\textbf{Where the sample is thinner, this is stated outright.} A targeted follow-up search on 1 August 2026 found two
peer-reviewed Russian-language studies missed by the original search: both were published outside
arXiv and outside international conferences. The original protocol searched mostly by
English keywords in English-language databases, so sample completeness for
Russian-language journal publications is lower than for preprints and conference proceedings.
The follow-up protocol and the analysis of the missed studies are in \texttt{01-\allowbreak{}literature/targeted-\allowbreak{}journal-\allowbreak{}sweep-\allowbreak{}2026-\allowbreak{}08-\allowbreak{}01.md}.

\subsection{4. What the present study does}

This study closes the gap by comparing production conditions under fixed
tasks and with several non-coinciding operationalisations of machine-likeness.

The task is held fixed while its mode varies — from a rigid technical specification
to a free formulation. Origin does not change: both halves of the
compared pair are produced by the same model from the same task. The human part
is annotated by source regimentation level, and the annotation is done from the
source before the texts are read.

Machine-likeness is measured by four procedures with different inputs and different ways of
obtaining weights: an interpretable index with weights set before data collection;
a trained classifier on the same feature matrix; a zero-shot metric on an
open Russian-language model; and a judge model scoring against a rubric. None is treated as a
gold standard, and the agreement of their signs is tested rather than assumed.

What remains background in the studies listed above is measured separately: FPR on the
formal register and translated prose, stability of the decision under text transformations,
and the contribution of source and genre compared with the contribution of style.

A limitation of the method is named upfront: word lists have a shelf life, because
authors adapt to published markers (Geng and Trotta, 2025). This is an argument for
features not reducible to a word list, and it was taken into account in feature selection.

*Source: \texttt{01-\allowbreak{}literature/literature-\allowbreak{}review.md} §5, \texttt{evidence-\allowbreak{}matrix.csv},
\texttt{protocol-\allowbreak{}group-\allowbreak{}crosswalk.md}.*

\section{Related work}

\subsection{1. The detection task and the limits of its solution}

A universal binary detector does not transfer beyond the conditions in which it
was tested. RAID assembled six million generations across eleven models and eight
domains and showed that detectors break under a change of decoding strategy and
under adversarial edits (Dugan et al., 2024). A direct test of seven systems yields TPR at FPR 1\%
down to zero on unfamiliar domains and generators (Tufts et al., 2025). The same result is
reproduced by M4 (Wang et al., 2024a), MultiSocial on 22 languages (Macko et al., 2025) and MULTITuDE on
eleven (Macko et al., 2023). Borile and Abrate, 2025 name the cause: the detector picks up linguistic
and domain confounders, and these correlations do not transfer. Attempts to make
the decision explainable follow the same path — combining probability curvature
with stylometry and readability metrics, decomposing feature contributions via
SHAP (Marchenko Breneur et al., 2026).

Estimates of how far the task is solved at all diverge. Studies reporting
99.0 F1 cross-domain (Verma et al., 2024) and AUROC 94.92\% in OOD scenarios (Chen et al., 2025) measure
quality on the aggregate; studies that break the estimate down by condition see
drops to zero (Tufts et al., 2025) and to 35\% under attacks (Wang et al., 2024b). Both groups are right
within their own design, and this is the first thing that gets in the way of
comparing results across papers.

A recurring design limitation: the human side of the comparison is almost always
collected opportunistically. The corpus is taken from ready sources — news,
essays, Wikipedia — and the machine part is generated to match it. The conditions
under which the human text was written are neither fixed nor varied.

Hence the first step towards our question: if detector quality changes with
domain and generator, the next question is whether it changes with the
conditions under which the human was writing.

\subsection{2. Interpretable features and neural detectors}

The line-up of methods is set by three classes. Zero-shot detectors on likelihood
statistics: DetectGPT computes probability curvature and raises AUROC from 0.81 to
0.95 on GPT-NeoX news (Mitchell et al., 2023), Fast-DetectGPT replaces perturbation with sampling
and speeds up the computation 340-fold with a relative gain of about 75\% (Bao et al., 2024), Binoculars
catches over 90\% of ChatGPT generations at FPR 0.01\% without training (Hans et al., 2024), GPT-who
measures the uniformity of information density (Venkatraman et al., 2024), DivEye — diversity (Basani et al., 2025).
Trained classifiers: Ghostbuster (Verma et al., 2024), RepreGuard on hidden
representations (Chen et al., 2025), feature-augmented transformers (Mady et al., 2026). The third class is
interpretable features, and it is the main one for us.

The feature-based approach works and requires no access to the model. Stylometry
over lexis, grammar, syntax and punctuation yields MCC .87 in a seven-class task
on ten-sentence descriptions of terms (Przystalski et al., 2026). On
American Chemical Society papers, a classifier on off-the-shelf features separates
human text from GPT-3.5 with accuracy above 99\% (Desaire et al., 2023), and interpretable
features give AUC 0.988–1.000 under transfer across prompts on news (Vera-Jimenez et al., 2026).
The lexical diversity measures such studies rely on have themselves been
validated: MTLD is the only one of the compared indices independent of text
length (McCarthy and Jarvis, 2010), MATTR solves the same problem with a moving window (Covington and McFall, 2010). The intrinsic
dimensionality of embeddings provides an invariant: about 9 for alphabetic
languages and about 7 for Chinese, with machine text roughly 1.5 lower (Tulchinskii et al., 2023). Grammatical and
rhetorical differences persist as models grow and intensify in
instruction-tuned versions (Reinhart et al., 2025). Formal syntactic parsing distinguishes the
distributions of HPSG grammar types across six models and humans (Zamaraeva et al., 2025). On 225
argumentative texts, the model shows an excess of performed uncertainty
markers: form-meaning divergence is higher in machine texts at p < 0.001 and
$\Delta$ = 0.68 (Bakhshi, 2026). The discriminating power also holds on phonological,
morphological and lexical features of essays of equal length (Georgiou, 2025).

Which features are transferable is where studies diverge. A test of 284 features on 27
models and ten domains left a single class stable — measures of lexical
density (El Attar et al., 2026). Studies that selected features within one type of
text report high quality on that same type (Przystalski et al., 2026, Georgiou, 2025) — transferability they did not
test. A separate line: the direction of differences in lexical diversity
changes with the composition of the human control group (Ter\v{c}on and Dobrovoljc, 2025 versus Fredrick and Craven, 2025).

A recurring design limitation: a feature is validated on class separation, not
on stability across production conditions. A paper reports that the feature
separates human from machine, and almost never what a change of genre,
regimentation or the writer's level of preparation does to it.

For our task the second point matters more. A feature that reverses direction
with the composition of the comparison is unfit for the role of a universal
threshold, and the question of what exactly it measures remains open.

\subsection{3. Hybrid and edited authorship}

Human editing breaks detection harder than a change of model. On 6.5 thousand
expert-edited texts, an expert's edit moves machine text away from detection,
while text edited by a model is not recognised as human (Artemova et al., 2025).
Under word-level co-authorship, metric methods yield a mean F1 of 0.462
(Su et al., 2025). On 1445 hybrid CoAuthor texts, the best pairing of segmenter and
classifier gives 0.5166 Kappa, and the authors name the cause: almost half of the
texts have a mean homogeneous span of three sentences or less (Zeng et al., 2024).
Tools that distinguish four categories instead of two, including "machine, then
humanised" and "human, then polished by a model", appeared in response to the
same problem (Abassy et al., 2024). The adjacent task of authorship attribution degrades
on jointly written text compared with purely human baselines:
embeddings are stronger in binary detection, n-gram models are more stable in
verification (Richburg et al., 2024). The volume of the edit itself is measured separately from its result —
LZ77 compression correlates with editing time at 0.81 against 0.59–0.68 for
Levenshtein distance (Devatine and Abraham, 2024).

What editing does to style is where assessments diverge. A preregistered
experiment with 81 participants showed a shift towards the writer's own style (g=0.55) — and
at the same time greater retained closeness to the model than to the same
person's unassisted text (g=$-$1.43), plus a rise in stylistic homogeneity
(g=1.42) (Baumler et al., 2026). That is, editing simultaneously brings the text closer to its author and
leaves it machine-like by measurement; the conclusion depends on which of the two
quantities is taken as the target.

A recurring design limitation: the mode of editing is varied, but not the
conditions of the original writing. The human reference in these studies is
written freely — the participant writes as they wish, with no brief, no mandatory
sections and no length requirement. CoAuthor records interaction traces (Lee et al., 2022), but not regimentation.

Here the gap is most clearly visible. The literature measures what editing does
to a text, and does not measure what the task does to it.

\subsection{4. Attacks and robustness}

Evading a detector is cheap. Substituting homoglyphs for characters drops the mean MCC
of seven systems from 0.64 to $-$0.01 (Creo and Pudasaini, 2025). Paraphrasing through DIPPER lowers
DetectGPT accuracy from 70.3\% to 4.6\%, while a defence by retrieval over a base of 15 million generations
recovers 80–97\% at FPR 1\% on human texts (Krishna et al., 2023). Recursive paraphrasing of
passages of around 300 tokens reduces detection with little loss of quality (Sadasivan et al., 2023).
Stylistic humanisation gives a mean ASR of 92\% against five detectors (Gu et al., 2026).
A stress test across four categories — editing, paraphrase, prompting, joint
generation — drops quality by 35\%, and no system holds against all of them
(Wang et al., 2024b). Changing a few characters flips the decision (Huang et al., 2024).

Whether this can be fixed is where conclusions diverge. Some studies propose
architectural remedies and show gains of 6.5–18.25\% absolute accuracy under attack (Huang et al., 2024),
inference-time adaptation from 24.1\% to 90.5\% on adversarial material (Ren et al., 2026), or
separation of content and expression (Bao et al., 2025). Others record that with an FPR ceiling of
0.5\%, only one of four tested systems retains detection (Jabarian and Imas, 2025).
The disagreement is explained not by method but by the FPR at which quality is measured.

A recurring design limitation: the attack is modelled as a hostile action
after the text is written. Editing to evade and editing for quality are not
separated in these studies, though mechanically they differ: the former targets
the detector, the latter the reader.

Our question lies nearby, but in a different plane. Regimentation, a template
and a length requirement change the text before it is produced, with no intent
whatsoever to deceive a detector, and in the terms of these studies such editing
is not an attack at all.

\subsection{5. Fairness and the cost of a false positive}

Detector error is unevenly distributed across groups of writers. Detectors
systematically assign English texts by non-native speakers to the machine class
while classifying native speakers' texts correctly (Liang et al., 2023). Across sixteen systems,
essays by learners of English receive the machine label more often (Stowe et al., 2026). The BAID benchmark measures recall on
texts of under-represented groups across seven categories, including dialect and
formality (Basu et al., 2026). An audit of four detectors on 1992 excerpts from six everyday
genres: open systems mistake up to 78\% of human text for machine text, and with
an FPR ceiling of 0.5\% only one commercial system retains detection (Jabarian and Imas, 2025).

Assessments of human judgement diverge. Five annotators with daily
LLM experience made one error on 300 English-language non-fiction articles and beat most
detectors (Russell et al., 2025). Participants in a Japanese web survey correctly identified human
text in 31.5\% of cases, whereas stylometry on the same material gave 99.8\%
(Zaitsu et al., 2025). This is a range of results from different tasks: the language, the genre, the
amount of material presented and the judge's preparation all differ. The overall
conclusion is more modest than either of the two originals — the accuracy of
human judgement about origin depends strongly on the conditions of the
experiment. Separately, Zhu et al., 2024 shows that the "AI" label moves
human evaluation more than the text itself: a machine passage labelled
"human" receives 68.4\% preference against 1.537–1.650 in blind comparison.

A recurring design limitation: inequality is sought along attributes of the
writer — language status, demographics, dialect. Writing conditions do not
enter this grid, though formal register and a rigid template are available to
any author regardless of group.

For our work this is the nearest neighbour. If the machine-likeness score rises
with regimentation, then part of the measured subgroup effect may be explained
not by the group but by the conditions under which that group usually writes.

\subsection{6. Evaluation of style transfer and editing}

The toolkit for evaluating edits is less developed than the editing itself. A
survey of 97 style transfer studies: human evaluation protocols are
under-specified and not standardised, and reproducibility suffers (Briakou et al., 2021). Only a few automatic
metrics have been validated in experiments with humans (Ostheimer et al., 2023). Content preservation
metrics do not abstract away from the change of style, so meta-evaluation on
existing datasets leads to false conclusions (Pauli et al., 2025). Metrics are tested on sentiment
transfer and detoxification in English, Hindi and Bengali through
correlation with human judgements (Mukherjee et al., 2025).

The very separability of style and content is also contested. Studies of adversarial
encoder-decoder systems conclude that style within them does not separate from
content (Jafaritazehjani et al., 2020). Applied studies nevertheless measure style shift as a separate
quantity and obtain meaningful numbers: BERTScore F1 0.924 with 17 times fewer
parameters (Paneru, 2026).

A recurring design limitation: the target style is defined as a class — polite,
formal, non-toxic. Style as a consequence of an external requirement — a
template, a mandatory set of sections, a length limit — is not considered in
these settings.

Hence a methodological foothold for our work and at the same time a caution:
style shift can be measured, but the metric must be chosen with an eye to the
fact that its validity has not been tested everywhere.

\subsection{7. What of this has been tested on Russian}

Beyond English the picture differs. On Portuguese news, a disinformation
detection model gives 93\% accuracy on human texts and 75\% on
LLM outputs, and confidence markers in ChatGPT's factual news are 152\% higher (Rodrigues et al., 2026). On
Japanese public comments, stylometry separates the classes with 99.8\% accuracy
(Zaitsu et al., 2025). Multilingual shared tasks and benchmarks cover from 11 to 22 languages
(Wang et al., 2024c, Macko et al., 2025, Wang et al., 2025, Macko et al., 2023).

The sample contains five primary Russian-language studies. RuATD 2022 assembled 14 generators
and put binary and multi-class detection into a shared task, with 30 systems in the binary
subtask (Shamardina et al., 2022). LLMTrace annotates mixed authorship character by character for
English and Russian and supports locating machine spans within a text
(Tolstykh et al., 2025). AINL-Eval 2025 provided 52 305 Russian scientific abstracts from twelve
fields, five generators, and a requirement of transfer to unseen fields and models
(Batura et al., 2025). Russian is also part of multilingual benchmarks (Macko et al., 2023).

The fourth study stands closest to the subject of the present research. CoAT is
a corpus of 246 thousand human and artificial texts, thirteen generators,
six domains from the Russian National Corpus to digital diaries (Shamardina et al., 2024).
A transformer detector achieves accuracy 0.864 there, humans on the same data — 0.66,
generator attribution among fourteen candidates — macro-F1 0.521. Transfer to
unseen models and domains does not hold.

What matters for us is another dimension of the same study: \textbf{detectability
depends strongly on the type of the generative task.} Machine-translated texts
are recognised with accuracy 0.621, paraphrase — 0.768, free generation — 0.99.
That is, the way a text was produced shifts recognition accuracy from 0.621
to 0.99.

CoAT stops in the same place as the rest. The generator changes together with
the task type, so the effect of production conditions is not separated from the
effect of origin. The human part is collected from the same six domains with no
control of task, regimentation or channel. The authors name the limitations
themselves: no diversity of decoding strategies and a risk of corpus obsolescence.

The fifth study comes from the practice of document checking. Gritsay and
co-authors address the detection of machine fragments within scientific and
educational texts and bring quality to F1 0.968 for Russian at precision 0.970
and recall 0.968; the same approach gives 0.825 for English (Gritsay et al., 2023). The quality rests on multi-domain training and
paraphrase augmentation, and false positives are controlled with the
Benjamini-Hochberg correction. The authors' conclusion matches the line of the
whole section: a universal detector for all domains and languages, in their
estimate, is hardly achievable.

The divergence here is not between results but between the scope of the tasks.
Russian material is covered along the line of text origin and noticeably less
so along the line of the conditions of its creation: none of the five studies
varies task mode, regimentation or channel while origin stays the same.

The recurring design limitation is the same as in section 1, only sharper: the
corpus is taken from available sources, and the writing conditions of the human
part are not documented.

\textbf{Among the included studies, no FPR estimates were found simultaneously for
formal and for translated Russian-language prose.}

\subsection{Summary: what remains unmeasured}

The literature measures two things with confidence. The first is the origin of
a text, and here both the methods and the limits of their transfer are known.
The second is the result of editing, and here it is known that expert editing
moves a text away from detection, while style after it stays closer to the
model than to the author.

\textbf{In the included sample of 73 studies we found no research that separately
assessed how text production conditions — task mode, regimentation level and
generation channel — change a text's machine-likeness indicators while its
origin stays the same.}

Three groups of studies come right up to this point and stop. Prevalence
estimates count the share of text that has passed through a model: from 6.5\% to 16.9\%
in the reviews of four conferences (Liang et al., 2024), at least 13.5\% of biomedical abstracts
in 2024 (Kobak et al., 2025). Studies of hybrid authorship vary the mode of editing but not the
conditions of the original writing (Artemova et al., 2025, Su et al., 2025, Zeng et al., 2024, Baumler et al., 2026). Fairness studies measure
the dependence of the score on the writer's group, not on the conditions under which the writer wrote (Liang et al., 2023,
Stowe et al., 2026, Basu et al., 2026).

The practical stakes are visible from Fredrick and Craven, 2025: the direction of differences in lexical
diversity changes with the composition of the human control group. If a
machine-likeness indicator depends on the conditions under which the human
reference was written, a threshold calibrated on one corpus transfers to
another without guarantees. In accusatory use this means the risk of punishing
regimentation rather than the origin of the text.

Add to this Geng and Trotta, 2025: the frequency of words publicly named as characteristic of the model
falls in arXiv abstracts soon after such lists are published — authors adapt.
Any conclusion that rests on a marker dictionary has a shelf life, and this is
an argument for features not reducible to a word list.

\section{Methods}

\subsection{1. Corpus}

The analytical corpus contains 1882 Russian-language documents: 1079 machine and
803 human. Its composition is fixed by the registry \texttt{04-\allowbreak{}corpus/documents-\allowbreak{}registry.csv};
each document carries a passport of 60 fields, including generation channel, task mode,
genre, publication date, licence status and split groups.

\subsubsection{1.0. Data flow}

Corpus composition by stage and stratum is given in Table S1.

\begin{center}
\small
\begin{tabularx}{\textwidth}{Xlll}
\toprule
Stage & Machine & Human & Total \\
\midrule
Before extraction correction and final exclusions & 1079 & 837 & 1916 \\
Excluded after correction & 0 & 34 & 34 \\
Analytical corpus \texttt{prep-\allowbreak{}v5} & 1079 & 803 & 1882 \\
\bottomrule
\end{tabularx}
\end{center}

All 34 exclusions fell on the human subset and on a single cause: after
correction of an extraction defect, the text dropped below the 700-word threshold.
The threshold rule was registered before the correction; the decisions are recorded
in \texttt{00-\allowbreak{}admin/exclusion-\allowbreak{}log.csv} under the stage \texttt{extraction correction correction-\allowbreak{}v5.0}
dated 2026-07-29.

\begin{center}
\small
\begin{tabular}{llll}
\toprule
Stratum & Before & After & Change \\
\midrule
Journalism, \texttt{news} & 180 & 152 & $-$28 \\
Template-based materials, \texttt{seo} & 237 & 231 & $-$6 \\
Free-form authorial texts, \texttt{prose} & 180 & 180 & unchanged \\
Standardised prose, \texttt{science} & 167 & 167 & unchanged \\
Language-contact texts, \texttt{translation} & 73 & 73 & unchanged \\
\bottomrule
\end{tabular}
\end{center}

Intermediate design documents written before these exclusions contain the numbers
1916 and 837 — these are snapshots as of their date, not a discrepancy. The registry
is the only authoritative source of counts.

\subsubsection{1.1. Machine subset}

The machine subset is built to the design \texttt{45 tasks $\times$ 4 generation channels $\times$ 3 task modes $\times$ 2 replicates}. The forty-five tasks are distributed evenly across three
genres — analytics, commercial text, SEO. The task modes: \textbf{P1} — a neutral
technical brief, \textbf{P2} — a strict editorial brief, \textbf{P3} — a task requiring the
avoidance of typical machine-text features. The full design yields 1080 cells;
the corpus has 1079, because one \texttt{prompt\_condition=P3} cell was missing in the
analytics genre.

\textbf{The unit of variation is the generation channel, not the model family.} A channel
comprises the model, the access wrapper and the wrapper's potentially hidden system
prompt. These cannot be separated: for two channels the wrapper adds its own system
prompt to the request, and its content cannot be reconstructed after the fact.
Between-channel differences therefore cannot be attributed to the model. Channels,
models and access parameters are given in Table S2.

\begin{center}
\small
\begin{tabularx}{\textwidth}{llXll}
\toprule
Channel & Model & Wrapper and access & Open weights & Temperature and top-p \\
\midrule
\texttt{real\_claude} & \texttt{claude-\allowbreak{}opus-\allowbreak{}4-\allowbreak{}8} & nested CLI, clean process per cell & no & channel defaults \\
\texttt{gpt} & \texttt{gpt-\allowbreak{}5.5} & \texttt{codex exec}, reasoning effort high & no & channel defaults \\
\texttt{deepseek\_pro} & \texttt{deepseek-\allowbreak{}v4-\allowbreak{}pro} & HTTPS API without a wrapper & yes & 1.0 / 1.0 \\
\texttt{nemotron} & \texttt{nemotron-\allowbreak{}3-\allowbreak{}ultra-\allowbreak{}free} & HTTPS API via a gateway & yes & 1.0 / 1.0 \\
\bottomrule
\end{tabularx}
\end{center}

The channels were chosen so that the models belong to different families rather
than to versions of one platform. Each channel produced 270 documents, except
\texttt{nemotron} with 269.

Two limitations were recorded before collection. First: published weights for a
family do not prove that a hosted response came from a known checkpoint with known
quantisation, so these channels cannot be called reproducible. Second: the seed
\texttt{random.Random(20260723)} fixes the order of cells, not the model output —
reproducibility means repeatability of the procedure, not byte-level identity of
the texts.

Two channels operate through a CLI and carry a residual wrapper system prompt that
cannot be removed; it is declared part of the channel. The wrapper version is not
constant within a channel: the OpenAI channel has two recorded CLI versions, and
for Anthropic the version was not recorded at all. The content of the system
prompts cannot be reconstructed after the fact, so one cannot claim that the
prompt changed — nor the opposite.

A response is described by two independent variables instead of a single validity
threshold. \texttt{technical\_validity} covers response integrity: below 50 words a
response counts as a technical failure and is regenerated.
\texttt{length\_instruction\_compliance} describes compliance with the instruction on a
length of 1200–1800 words in five grades from \texttt{severe underlength} to \texttt{severe
overlength}. The separation was introduced because the previous single threshold
of 480 words systematically sent P3-mode texts for regeneration and would thereby
have biased the primary outcome. A model refusal is not screened out by the
threshold and remains a result.

\subsubsection{1.2. Human subset}

Human texts were published before 2022-01-01; the date is confirmed by two
independent methods. The regimentation level is assigned by source and its
editorial standard \textbf{before the text is read}, and the grounds are recorded. The
composition of both corpus subsets together with the production conditions is
given in Table 1.

\setcounter{table}{0}
\input{tables/tab-01-corpus-and-conditions}

Six documents in the \texttt{seo} stratum — 0.7\% of the human subset — were drawn from
the author's 2016–2017 archive, which existed before the study began. They went
through the same processing and the same annotation as the remaining documents.
Their separate influence on the reported estimates was not measured.

\begin{quote}
Published wording: Six SEO documents (0.7\% of the human subset) were drawn
from the author's pre-existing 2016–2017 archive. They were processed and
annotated under the same rules as the remaining documents. Their separate
influence on the reported estimates was not evaluated.
\end{quote}

The hard-human subgroups are annotated as flags on top of the corpus, not as a
separate stratum: a document enters the overall false positive calculation once
and is then shown in the subgroup analysis. The corpus carries the annotations
\texttt{HH-\allowbreak{}formal-\allowbreak{}register} (167 documents), \texttt{HH-\allowbreak{}polished} (152) and \texttt{HH-\allowbreak{}translation} (73).

A known overlap: \texttt{HH-\allowbreak{}formal-\allowbreak{}register} coincides with the level-3 stratum, and
\texttt{HH-\allowbreak{}translation} consists entirely of a single source. The latter is a composition
defect recorded in the limitations: a challenge set built on one corpus teaches a
model the source, not the target attribute.

Hybrids with human editing are not part of the main study: they require editors,
and hiring was ruled out.

\subsection{2. Identifiability and outcome status}

Design identifiability was checked on 27 July 2026 against the registry passports,
\textbf{before the dependent variable was computed} (\texttt{design-\allowbreak{}support-\allowbreak{}table.md}).

The check showed that genre in the human subset serves as a proxy for both the
regimentation level and origin. Level 3 is represented by the single genre
\texttt{science}, level 0 by the single genre \texttt{prose}; for the groups of the primary
outcome O3 the genre intersection is empty.

Hence the statuses assigned before computation:

\begin{itemize}
\item \textbf{O1} — a pre-specified contrast. The difference between modes P2 and P1 lives inside the machine subset, where task, genre and channel are fully crossed, and the contrast itself was registered before data collection.
\item \textbf{O2 and O3} — descriptive. The registered estimates for them are not computed, and no p-values are published.
\end{itemize}

\textbf{The study is not a fully confirmatory analysis.} The O1 contrast was specified
in advance, but three of the four operationalisations used to estimate it received
their operationalisation on 27 July, after the results of the first procedure had
been seen (§4). The correct formulation: a pre-specified contrast estimated
predominantly with exploratory operationalisations. Confirmatory status belongs to
the question, not to the whole set of measurement instruments.

The downgrade of O2 and O3 took place before the machine-likeness score was
computed and does not depend on the obtained numbers.

\subsection{3. Preprocessing and extraction correction}

All procedures work with the \texttt{prep-\allowbreak{}v5} profiles. Preprocessing is frozen by the
specification \texttt{06-\allowbreak{}features/preprocessing-\allowbreak{}spec.md} and builds two profiles: \texttt{prose} —
connected prose without headings, lists and tables, and \texttt{full} — the complete text.
A documentary erratum,
\texttt{06-\allowbreak{}features/preprocessing-\allowbreak{}spec-\allowbreak{}erratum-\allowbreak{}2026-\allowbreak{}08-\allowbreak{}01.md}, applies to §2.28 of the
specification: the number of collapsed typographic hyphenations is 6955 across 124
documents under the frozen \texttt{prep-\allowbreak{}v5} manifest. The erratum does not affect the
rule, the code or the profiles. Counters of the markup removed when building the
profiles are stored in the manifest and used by the format features.

Before the freeze, the script \texttt{09-\allowbreak{}tools/correct\_extraction.py}, following rule
\texttt{correction-\allowbreak{}v5.0}, examined 1916 documents and found 113 candidates with a
verbatim sentence duplication share above 0.1. Of these, 103 were corrected
(Table S3):

\begin{center}
\small
\begin{tabularx}{\textwidth}{llX}
\toprule
Verdict & Documents & Action \\
\midrule
\texttt{extraction-\allowbreak{}defect} & 75 & re-extraction with the corrected parser \\
\texttt{intermediary-\allowbreak{}defect} & 28 & deterministic removal of exact cascades \\
\texttt{unresolved} & 9 & not edited; origin not established \\
\texttt{source-\allowbreak{}property} & 1 & not edited; the repetition is present in the source \\
\bottomrule
\end{tabularx}
\end{center}

The rules were registered by amendment before computation; the source files were
not overwritten. The defect is one-sided — duplicates were found only in the human
subset — and this is recorded in the limitations.

\subsection{4. Four operationalisations of the machine-likeness score}

The effect is tested in four ways so that the conclusion does not reduce to an
artefact of a single implementation. No procedure is designated the reference:
sign agreement across all four is an argument, divergence is a result in its own
right.

\begin{enumerate}
\item[1.] \textbf{Interpretable index.} A weighted procedure over countable features, with
   weights fixed before data collection. Scale 0–100.
\item[2.] \textbf{Trained classifier.} Logistic regression on the primary features, evaluated
   only on group holdouts by topic, source, model, task mode, genre, revision
   family and time. Nested cross-validation for regularisation; the diagnostic
   baselines "topic only" and "source only" are mandatory.
\item[3.] \textbf{Zero-shot metric.} Mean NLL under an open Russian-language model.
\item[4.] \textbf{Judge model.} The median of three runs with fixed seeds; the dispersion is
   recorded separately.
\end{enumerate}

Procedures 2–4 carry exploratory status: their operationalisation was written on
27 July, after the results of procedure 1 had been seen, whereas the
preregistration specified each in a single line. Procedures 1 and 2 count as one
family of evidence: both aggregate the same feature matrix over the same documents
and differ in how the weights are obtained.

\subsection{5. Estimation of the primary outcome O1}

O1 is the difference in machine-likeness score between task modes on paired
\texttt{task $\times$ model $\times$ replicate} observations; two contrasts are registered, \texttt{P3 $-$ P1}
and \texttt{P2 $-$ P1}. The unit is points on the 0–100 scale.

The cluster is the task: forty-five tasks, each containing all channels, modes and
replicates. Confidence intervals are computed by cluster bootstrap over tasks
at the 95\% level; the absolute difference with its interval is published,
not only a p-value.

The family of primary outcomes is closed by a Bonferroni correction for the two
registered tests, $\alpha$ = 0.025 per test. Sensitivity contrasts are compared against
the same threshold.

Benjamini–Hochberg was registered for the feature family and receives the status
\textbf{not applicable}: no inferential tests of individual features were run, and the
four differently scaled summary contrasts do not form a homogeneous family. The
wording was approved by the PI before computation.

Individual features in the figures and the error analysis remain descriptive:
effects and intervals without selection by p-value.

\subsection{6. Mixed-effects: three models and the block outcome}

The preregistration specifies three models (Table 8; identifiability diagnostics
in Table S8):

\setcounter{table}{7}
\input{tables/tab-08-mixed-effects-status}

The order of execution was fixed before fitting: first an identifiability check
against the design — crossing of factor levels, number of clusters per random
effect, degeneracy of interactions — then fitting only the identifiable models.
On convergence failure the diagnostics are saved in full: optimiser warnings,
eigenvalues, the simplified specification.

None of the three models yielded interpretable estimates. M1 did not converge
after removal of an exact linear dependency and exhaustion of the standard
BFGS–L-BFGS–CG chain. M2 proved structurally non-identifiable. For M3 there was no
support for the registered cross-genre origin estimand: within the SEO genre a
conditional contrast is technically available, but it does not identify the origin
effect for the whole corpus.

The wording for the paper was approved by the PI: the estimates were not obtained
or lacked the required interpretation. Claiming that the mixed-effects models
"found no effect" is prohibited — that is a different claim.

Switching the optimiser to Powell or Nelder–Mead, simplifying the random effects,
or a separate model for SEO after the block was closed would have constituted a
new exploratory analysis and were not performed.

The wild cluster bootstrap does not replace these models: it assesses the
stability of the mean paired contrast, whereas mixed-effects models capture the
nesting and the heterogeneity across tasks and channels. Substituting one for the
other in the text is not permitted.

\subsection{7. Stability stress test}

The panel comprises 60 documents selected and fixed before the run
(\texttt{07-\allowbreak{}analysis/stress-\allowbreak{}panel-\allowbreak{}v1.csv}). The panel does not change between revisions.

Of the sixteen registered transformations, ten are executed: em-dash replacement,
markup changes, heading removal, paragraph reordering, punctuation correction,
text shortening, text expansion, adding spelling errors, homoglyph substitution,
transfer between formats. Six are not assessed: five require material absent from
the corpus — the A$\to$H branch with editors, machine rewriting, translation and
back-translation, paraphrasing; the sixth, removal of typical clich\'es, was ruled
infeasible after \texttt{prep-\allowbreak{}v5}, because the registered artefact list produces no hits
on the cleaned text.

The panel of 60 documents and 10 transformations yields 600 primary
\texttt{document $\times$ transformation} cells. Each cell passes through all four procedures on
their native scales: the index on 0–100, the classifier as the probability of
class A, NLL in its own units, the judge on 0–100 with the median of three
seeds.

\textbf{The instability threshold is set separately for each procedure, before the run.}
A single threshold cannot be applied: the procedures operate on different scales.
Inputs, scales, directions and thresholds of the four procedures are summarised in
Table 2.

The notation is kept distinct: \textbf{P1, P2 and P3 are task modes} (§1.1); the
procedures are referred to by number.

\setcounter{table}{1}
\input{tables/tab-02-four-operationalisations}

For procedure 2 the two quantities are computed separately, following the
aggregation rule of the amendment: the share of eligible models with
\texttt{|$\Delta$p| > 0.05} is the instability, the share crossing the decision threshold is
the decision flip; both are averaged over models within a document and then over
the 60 documents with equal weight. For procedure 3 the absolute change cannot be
called points, so no share of unstable cases is defined for it at all.

The 5.0 threshold is operational, not empirically validated. Alongside the event
share, continuous shift magnitudes are published — median, IQR, 95th percentile,
maximum — for all four procedures.

The percentile of a transformed text is computed against the unchanged genre pool:
transformed values are not added to the pool, and the document's original value is
not removed from it.

\subsection{8. Versioning and invalidation rules}

Each run is described by a manifest with hashes of code, inputs and outputs. An
erroneous manifest is not deleted: it is marked \texttt{invalidated} with a reason, and
work continues in a new revision. A revision voids the child manifests assembled
under the previous parent and forces their procedures to be recomputed. The final
numbers of the main series were obtained under the sixth revision in a single
chain; the stress test stands at revision r5 after three invalidated ones.

Changes to the computation code require an amendment and a new revision. Hashes of
code and inputs are fixed before launch; the chain verifies them and blocks the
start on any mismatch. Code checking takes place before the run, on synthetic data
of the same type that production supplies.

Parse caches are keyed by \texttt{document\_id $\times$ input\_sha256 $\times$ cache\_version $\times$
model\_revision}. The preprocessing version is recorded as provenance but is not
part of the validity check: a byte-identical input yields a byte-identical parse.
New entries do not overwrite earlier ones, so earlier series remain reproducible.

The procedure completion gate checks completeness mechanically — the number of
cells, feature coverage, denominators, absence of dropped categories — and stops
the chain on a \texttt{blocked} or \texttt{incomplete} status. A triggered gate counts as a
quality control result; loosening the tolerance to let a run pass is prohibited.

A separate invariant checks cells whose input has not changed: when the \texttt{prose}
profile, the \texttt{full} profile and all preprocessing counters match, the score must
equal the baseline identically, with a tolerance of 1e-9.

\section{Results}

\subsection{1. Corpus composition and QA closure}

The corpus comprises 1882 documents: 1079 machine and 803 human (Table 1; the
data flow and per-stratum exclusions are in Table S1). The machine part was
produced by four generation channels — each backed by its own model family — across 45 tasks in three task modes with two
replicates. The human part consists of archival texts published before
2022-01-01. The active preprocessing version is \texttt{prep-\allowbreak{}v5}; the features are
collected in \texttt{feature-\allowbreak{}matrix-\allowbreak{}v5.csv}.

Preprocessing removes collection artefacts, and most of its rules apply to the
human part: comment threads in 67 documents totalling 680 570 characters,
sidebars and navigation, dangling lead-ins, paragraph segmentation in 304
documents, typographic hyphenation of the form \texttt{\foreignlanguage{russian}{за-\allowbreak{} ключение}} — 6955 cases in
124 documents, all human: \texttt{urfu} 3460, \texttt{spbgu} 3390, \texttt{cyberleninka} 100, with
isolated cases in \texttt{taiga\_magazines}, \texttt{taiga\_arzamas} and \texttt{lenta}. For the
machine part, the channel wrapper was removed: 99 \texttt{deepseek\_pro} documents
opened with an address to the client who commissioned the text instead of the article text.
The asymmetry of the rules is acknowledged and recorded: syllable hyphenation is
produced by PDF layout, while machine texts arrive from the API without layout.

Residual verbatim repeats were measured by a symmetric scan of both parts. The
threshold is a share of repeated sentences above 0.10 with a length filter of
40 characters. \textbf{17 human documents exceeded the threshold, and none of the
1079 machine documents did}; in the machine part the share of repeats is zero
for every document, not merely below the threshold. On the previous
preprocessing version the same ratio was sharper: 97 against zero, so the
correction removed about eighty per cent of the cases.

The seventeen candidates were classified manually by a single rater (Table
S9): 10 are confirmed extraction defects, 4 are legitimate properties of the
source, 3 remain unresolved. The defects are block-level repeats of fragments
in the intermediary dataset \texttt{buriy\_2014} and a cookie-plugin line. The
legitimate repeats are a doubled bibliography in three SPbU papers, where the
\texttt{References} section splits the repeats exactly in half, and a block of
interview questions. Separately, a mechanism common to the preprocessing is
recorded: detaching a heading from text fused to it can produce a verbatim
repeat where the source had none.

*Sources: \texttt{04-\allowbreak{}corpus/documents-\allowbreak{}registry.csv}, \texttt{06-\allowbreak{}features/preprocessing-\allowbreak{}spec.md},
\texttt{07-\allowbreak{}analysis/repeat-\allowbreak{}scan-\allowbreak{}v2-\allowbreak{}report.md}, \texttt{repeat-\allowbreak{}scan-\allowbreak{}v2-\allowbreak{}verdicts.md}.*

\subsection{2. Primary outcome O1: four procedures and a sign disagreement}

O1 compares task modes within the machine corpus. Both halves of a pair were
produced by the same model for the same task; only the wording of the brief
differs; human texts are not part of the contrast.

The \texttt{P3 $-$ P1} contrast, estimand full, is given in Table 3, which reports both
contrasts and both estimands:

\setcounter{table}{2}
\input{tables/tab-03-o1-contrasts}

\textbf{The signs disagree, and this is a result in its own right.} Three
procedures give a minus, the judge gives a plus. The disagreement is no reason
to look for an error in one of them: the inputs differ — feature matrix versus
raw text, and fixed weights versus learned ones.

\textbf{The plan for the cross-procedure comparison and the specifications of
procedures 2–4 were fixed after the result of procedure 1 had been obtained,
but before procedures 2–4 were run.} The rule that reads a sign disagreement
as a result in its own right is recorded in the same place — on 27 July, in
§4.4 of the registration. This is a sequentially registered comparison, not a
fully prospective registration of all four results: by the time the plan was
fixed, the result of the first procedure was already known.

Procedures 1 and 2 count as one family of evidence, not two independent ones:
both aggregate the same matrix over the same documents.

Procedure 2 computes the contrast on a different population — only the \texttt{seo}
genre, 120 pairs and 15 clusters against 359 pairs and 45 clusters for the
others. These are not the same pairs.

On the \texttt{P2 $-$ P1} contrast the picture differs: procedures 1, 2 and 4 give a
plus and procedure 3 a minus, but its interval covers zero (p = 0.1988), so it
contributes little to the disagreement. For the judge, p = 0.0392 is above
$\alpha$ = 0.025 after the Bonferroni correction, and this contrast is not called
significant.

There is no single number across the four procedures and there will be none:
no "two out of three" vote is applied, and no pooled meta-analytic effect is
computed. The magnitudes are not comparable across rows — index points, class
probability, mean NLL and the rubric scale are four different scales.

*Source: \texttt{07-\allowbreak{}analysis/synthesis-\allowbreak{}o1-\allowbreak{}v2-\allowbreak{}report.md}.*

\subsection{3. Index decomposition, classifier ablation, and comparison of run schemes}

These are three different things, and they must not be conflated. The first is
an algebraic decomposition of the index contrast, exact by construction. The
second is a comparison of separately trained models, which is not a
decomposition. The third is a comparison of two frozen runs with each other.

\subsubsection{3.1. Index decomposition: algebraic and exact}

The index is assembled from categories with weights fixed before data
collection, so its contrast decomposes into summands exactly; the check of the
decomposition on the v1 series is in Table S14. Both categories are computed
for all documents; the denominator is the same.

\begin{center}
\small
\begin{tabularx}{\textwidth}{Xl}
\toprule
Component of the \texttt{P3 $-$ P1} contrast & Value \\
\midrule
contribution of the style categories (common) & \textbf{+1.6740} \\
contribution of the structural category (format) & \textbf{$-$3.4395} \\
sum of contributions & $-$1.7655 \\
full contrast of the frozen run & $-$1.7655 \\
\bottomrule
\end{tabularx}
\end{center}

The discrepancy between the sum and the full contrast is 1.6$\cdot$10$^{-4}$ — rounding
to four digits when writing to CSV.

The mechanism reads off directly: the common component shifted the index
upward by +1.6740 points, while the format component shifted it downward by
$-$3.4395 and determined the negative overall contrast. The format contribution
exceeds the common contribution roughly twofold in absolute value.

This is a property of this index on this corpus, not a universal ratio: the
category weights are set by the specification, and the feature set fully
covers two categories out of thirteen. The index without the structural
category gives +2.0264 [0.8953; 3.1429], p = 0.0016 — the sign flips.

\subsubsection{3.2. Classifier: ablation, not decomposition}

The \texttt{full} and \texttt{net} models are trained separately, so their difference cannot
be called a contribution of structure. Both contrasts appear in Table 4
alongside the quality of the model variants.

\setcounter{table}{3}
\input{tables/tab-04-classifier-and-ablation}

The correct wording: excluding the structural features changed the direction
of the contrast, which shows that the result depends on the feature set.
Wordings of the form "the structural component contributed so many points" do
not apply to the classifier — they presuppose an additivity that separately
trained models do not have.

Without the structural component the minus disappears in both procedures: for
the index the sign flips, for the classifier the effect ceases to differ from
zero.

\subsubsection{3.3. Comparing the two runs with each other}

A separate question is what each of the two changes between the frozen runs
contributed. Both changed the selected regularisation for three holdouts out
of eighteen.

\textbf{Corpus correction} — 34 documents were excluded, and the text changed in

\begin{enumerate}
\item[68.] The shifts are small in absolute value on average and concentrated on one
holdout: AUROC 0.005 on average with a maximum of +0.034, MCC 0.029 with
+0.093, FPR 0.021 with $-$0.078. The maximum across all six metrics falls on
\texttt{holdout\_genre\_seo}.
\end{enumerate}

\textbf{Nested CV fix} — the corpus is the same; only the inner-split scheme
changed: for seven holdouts the validation became two-class. The shifts are of
the same order but with the opposite sign for SEO: AUROC $-$0.050, MCC $-$0.227,
FPR +0.234, TPR@1\%FPR $-$0.869.

The wording of the last result is fixed and matters: the degenerate fold
happened to select a stronger regularisation, which worked better on this
outer holdout, and this can only be learned from the test metrics. Writing
that it "picked C more successfully" is prohibited — such a wording would
present a defect as an advantage.

The defect predates the correction: seven of the 54 planned inner folds had a
single-class validation already in the first version of the classifier, where
the selection of C skipped them silently — neither the metrics, nor the
manifest, nor the report contained the number of folds used.

*Sources: \texttt{07-\allowbreak{}analysis/score-\allowbreak{}v2-\allowbreak{}index-\allowbreak{}decomposition.md}, \texttt{clf-\allowbreak{}v2-\allowbreak{}report.md},
\texttt{clf-\allowbreak{}v2-\allowbreak{}comparison.md}, \texttt{clf-\allowbreak{}v1-\allowbreak{}split-\allowbreak{}degeneracy.md}.*

\subsection{4. P2a and P2b: quality, SEO instability and formal register}

The classifier is evaluated on 18 group holdouts. For five of them the test
part is single-class, so AUROC is undefined; below are the medians over the
thirteen remaining (Table 4). Per-holdout metrics row by row are in Table S5;
the false positive rate by holdout is in Table S4.

\textbf{The key row here is \texttt{source-\allowbreak{}only}.} The source identifier turned out to be
an almost deterministic proxy for the origin class: the \texttt{source-\allowbreak{}only} baseline
reached a median AUROC of 1.0000 over the thirteen holdouts where AUROC is defined — higher than the
model on twenty-two style features. This points to strong confounding of class
with the corpus assembly mechanism and limits the transferability of the
assessment to sources where human and machine texts coexist.

\textbf{Perfect ranking here does not mean error-free classification.} Next to the
AUROC of 1.0000, the same model has a median FPR of 1.0000: at the fixed
threshold of 0.5, \texttt{source-\allowbreak{}only} classified all human documents of an unseen
source as machine. The source works as a strong ranking proxy, but the
decision threshold does not transfer to a new source. The superiority over the
model on twenty-two style features is therefore claimed only in AUROC and does
not extend to threshold-based quality.

What is compared here are models and input sets, not the informativeness of
individual features: the baseline is built on a single input, the main model
on twenty-two, and no conclusion about the individual contribution of any
feature follows from this comparison.

The quality of the main model is uneven. Median MCC 0.9338 with a minimum of
0.6560, median FPR 0.0380 with a maximum of 0.3098 — both worst values on
\texttt{holdout\_genre\_seo}. The same holdout is the worst in AUROC (0.9433) and in
TPR@5\%FPR (0.6250 against a median of 0.9833).

\textbf{Formal register.} The registered subgroup disparity rule fired for the
\texttt{HH-\allowbreak{}formal-\allowbreak{}register} subgroup in 5 holdouts out of 18: the subgroup's median
FPR is 0.066, with a difference of 0.043 from the rest of the test set. The
aggregate fairness verdict did not change, but the vulnerability ceased to be
a single-source effect.

The bias was observed in the strata annotated \texttt{HH-\allowbreak{}formal-\allowbreak{}register},
\texttt{HH-\allowbreak{}polished} and \texttt{HH-\allowbreak{}translation}, and each of these labels coincides with
the corresponding genre document for document: formal-register — all 167
\texttt{science} texts, polished — all 152 \texttt{news}, translation — all 73
\texttt{translation}. The effects of register and genre are therefore not separately
identifiable on these data, and no number in this subsection can be attributed
to register over and above genre.

Subgroup magnitudes are in Table 5; it prints the median, the range and the
difference with the rest of the human test set instead of the "5 out of 18"
count.

\setcounter{table}{4}
\input{tables/tab-05-formal-register-fairness}

The genre label also hides the spread between sources within a genre. In
the \texttt{news} genre the median FPR is 0.013, while across the sources within it:
\texttt{buriy\_2014} 0.033, \texttt{gazeta} 0.000, \texttt{lenta} 0.000.

*Sources: \texttt{07-\allowbreak{}analysis/clf-\allowbreak{}v2-\allowbreak{}p2a-\allowbreak{}metrics.csv}, \texttt{fairness-\allowbreak{}v2-\allowbreak{}report.md},
\texttt{clf-\allowbreak{}v1-\allowbreak{}split-\allowbreak{}degeneracy.md}.*

\subsection{5. Calibration and risk–coverage}

The primary estimate is a macro average over the 18 holdouts with equal weight
(Table 6; per-holdout values are in Table S6). Equal weight goes to the
registered holdout procedures, not to independent document sets: the test
parts overlap, and the composition of \texttt{holdout\_genre\_news},
\texttt{holdout\_genre\_science} and \texttt{holdout\_genre\_translation} coincides document for
document — all three contain the same 392 texts.

\setcounter{table}{5}
\input{tables/tab-06-calibration-and-risk-coverage}

The overall error of the probabilistic predictions is small. ECE is stable
across the number of bins — 0.0557, 0.0571 and 0.0581 — but \textbf{this is
stability of the estimate, not proof of good calibration}.

Calibration is uneven. \texttt{holdout\_genre\_seo} is simultaneously the worst in
Brier (0.1796), ECE (0.1973), MCE (0.7548) and FPR (0.3098).
\texttt{holdout\_genre\_prose} shows the flip side: ECE 0.1727 with Brier 0.0674 —
acceptable separation does not guarantee good calibration.

The risk–coverage curve is descriptive. The error falls from 0.0468 at full
coverage to 0.0059 at half coverage; the actually achieved coverages are
1.0000, 0.9003, 0.8004, 0.6997, 0.5009. \textbf{No rejection threshold has been
chosen, and the curve does not turn into an operational recommendation.}

The pooled variant of the same curve is a sensitivity to the weighting unit,
not an alternative result: 0.0500 at full coverage and 0.0137 at half. The
difference from macro is explained by the repeated weight of documents scored
by several models, not by a deterioration of the classifier itself: a document
enters the test in one to eighteen splits, and in pooled it is counted as many
times as the number of splits that cover it.

The roll-up to the document level is a sensitivity with a different estimand,
not an improved version of the primary estimate: Brier 0.0499, ECE 0.0438 on
1882 documents.

Two limitations were recorded during the computation. First, for four holdouts
out of eighteen the equal-frequency binning degenerates, leaving 8–9 non-empty
bins instead of ten — the scores saturate at the edges of the scale; in
\texttt{holdout\_prompt\_P2}, 253 values out of 752 lie above 1 $-$ 5$\cdot$10$^{-7}$ even before
rounding. Second, the precision of the source is limited to $\pm$0.5$\cdot$10$^{-6}$, while
the decisions and the composition are reproduced in full — 12 112 out of
12 112.

The score does not translate into a probability of origin: the analysis plan
prohibits such a translation without a representative calibration corpus, a
predefined target population, drift checks and a published error profile.

*Sources: \texttt{07-\allowbreak{}analysis/calibration-\allowbreak{}v1-\allowbreak{}report.md}, \texttt{calibration-\allowbreak{}v1.sha256.md}.*

\subsection{6. Mixed-effects: three different reasons for the absence of estimates}

None of the three registered models produced interpretable estimates, and the
reasons differ.

\textbf{M1} — the machine corpus, yields O1. The matrix was reduced to full rank 22
of 22, but convergence was not reached by any of the three optimisers
permitted in advance: BFGS, L-BFGS and CG, with gradient norms 35.26, 15.62
and 4.47. Coefficients and intervals were not interpreted. The O1 conclusions
rest on paired within-task contrasts and a cluster bootstrap over the 45
tasks.

\textbf{M2} — the human corpus. Structurally unidentifiable: rank 7 of 22, all
twelve \texttt{level $\times$ genre} interactions are degenerate, and the registered random
term "task" has zero clusters in the human part.

\textbf{M3} — the cross-genre estimand of origin. The registered contrast has no
support in the data.

The wording "mixed-effects found no effect" is wrong: the estimates were not
obtained or lacked the required interpretation, which is a different claim.
The bootstrap is the basis of the current estimates but not an equivalent
repair of the non-converged model: it assesses the stability of the mean
paired contrast, whereas mixed-effects models the nesting and the
heterogeneity across tasks and channels.

Powell, Nelder–Mead, simplification of the random effects and a separate model
for SEO would, after the closure of the block, have constituted a new
exploratory analysis and were not run.

*Sources: \texttt{07-\allowbreak{}analysis/mixed-\allowbreak{}m1-\allowbreak{}v2.md}, \texttt{mixed-\allowbreak{}identifiability.md},
\texttt{analysis-\allowbreak{}closure.md} §1.*

\subsection{7. Stress tests across the four procedures}

The panel is 60 documents, ten executable transformations, 600 primary cells
(Table 7; all four procedures on each transformation are given in Table S7).
All four procedures passed the completion gates. Procedure 2 produced 4640 rows on 464
document $\times$ eligible-model pairs, with reproduction determinism max|$\Delta$| = 0.

\setcounter{table}{6}
\input{tables/tab-07-stress-ten-transformations}

The shares are computed under the registered aggregation rule: the mean over
eligible models within a document, then equal weight for each of the 60
documents. The 0.451 for t11 is therefore the mean per-document share of models
that flipped their decision, not a share of the 464 pairs: in total the decision
flipped in 109 pairs of 464 for t11 and in 4 of 464 for t10 — the flips are
concentrated in documents with few eligible models.

\textbf{Transformations 10 and 11 are composite, and this changes how the table reads.}
Both functions split the text into sentences with the regex \texttt{\textbackslash{}s+} and reassemble it
with single spaces, so length, verbatim sentence duplication and paragraph
markup change at once: the median loss is 83.5\% of paragraphs under shortening and
75.6\% under expansion. The effect of volume on its own is not tested by these
transformations.

What is tested directly: shortening by 24\% produces 4 decision flips out of 464,
expansion by 25\% produces 109, and it is shortening that loses more paragraphs.
Neither length nor loss of markup explains this asymmetry. The pair is separated by
feature M01 — in the model it is the standard deviation of cosine similarities of
adjacent sentences, not the mean similarity. Duplication places identical phrases
side by side with a cosine of 1.0 against a typical 0.5 and raises the dispersion —
by a median of 7.86 $\sigma$, with an increase in all sixty documents.

All 109 decision flips go in one direction — towards the "human" class. The mechanism
is measured only for Procedure 2: in the training part, dispersion is systematically
higher in human texts (median 0.0907 versus 0.0746, Cliff's $\delta$ = $-$0.78), and the
M01 coefficient is negative in all eighteen models, with a median rank of 4 out of
22 by absolute value. The mechanism is corpus- and model-specific: it describes this
classifier on this corpus and does not make verbatim repetition a universal
marker of human writing. For Procedure 1 the mechanism requires algebraic
decomposition; for NLL and the judge it remains unknown.

Transformation t14, "removal of typical clich\'es", was reclassified as \texttt{not executable}:
after preprocessing there is nothing to remove — the registered list of artefacts yields
zero hits on the panel. The denominator is 10, and the result is called the
10-transformation stress test.

A separate sensitivity check, "paragraph-preserving truncation without verbatim
duplication", was registered to separate volume from duplication, and was
\textbf{rejected at preflight}: it passed three of the four criteria, but on M01 the median
$\Delta$z was +0.0079 against the requirement "not above zero", and in ten documents the
increase exceeded 0.5 $\sigma$. The criterion was not relaxed after seeing the result. This is
not a null result on the effect of length but the infeasibility of this operationalisation.

*Sources: \texttt{07-\allowbreak{}analysis/stress-\allowbreak{}p2a-\allowbreak{}r11-\allowbreak{}*} attempt \texttt{20260731T220132Z},
\texttt{stress-\allowbreak{}t10-\allowbreak{}t11-\allowbreak{}volume-\allowbreak{}defect.md}, \texttt{m01-\allowbreak{}direction.md},
\texttt{sensitivity-\allowbreak{}v1-\allowbreak{}rejection.json}.*

\subsection{8. Manual analysis of unstable cases}

Thirty cases were selected by a rule fixed before selection: sorting by the share of
models that flipped their decision, with ties broken by the largest probability shift.
The set is 30 cells on 28 documents, with a cutoff share of 0.750.

The analysis was blind; the procedure was structural attribution. Hidden were the document identifier, the origin class,
the split, the transformation name, both probabilities, the direction of the decision
flip and the position in the selection; two texts were presented in random order. The
verdict table was fixed by hash before the key was revealed.

\textbf{The procedure is called blind structural attribution with programmatic text
matching}, not manual reading, and this matters for reading the result. Pairs
were matched programmatically — number of sentences, paragraph separators,
repeated occurrences, the set of unique sentences and a line-by-line diff — and the
verbatim fragments cited as grounds were extracted from the differences found. The
analysis took two minutes forty-four seconds with a single rater. This procedure
establishes the mechanism of change and makes no claim to semantic expertise: shifts
not visible in structure it does not detect.

\subsubsection{Correspondence of verdicts to transformations}

Verdicts for each of the thirty cards are given in Table S10.

\begin{center}
\small
\begin{tabular}{lll}
\toprule
Transformation & Cards & Verdict \\
\midrule
t11 text expansion & 26 & composite impact, 26 of 26 \\
t04 paragraph reordering & 2 & interpretable stress case, 2 of 2 \\
t05 punctuation correction & 1 & interpretable stress case \\
t02 markdown markup change & 1 & indeterminate \\
\bottomrule
\end{tabular}
\end{center}

Blind, the mechanism was identified correctly in all twenty-nine cases where it
appeared at all in the presented profile. Twenty-six t11 cards were marked as
duplication together with paragraph restructuring — the very composite impact
that keeps the volume transformations from being read as a test of length. Both t04
cards were described as movement of blocks without verbatim repeats, and the t05 card
as a difference in spaces before punctuation marks with the sentence and paragraph
composition unchanged.

Distribution by origin class: all three interpretable cases are human documents;
twenty-five of the twenty-six composite ones are machine.

\subsubsection{The case found precisely because of blinding}

The card with transformation t02 received the verdict "indeterminate" on the ground
that "the texts match character for character". This turned out to be a correct
observation: for document \texttt{nemotron\_b016\_P1\_r1}, the \texttt{full} profile before and after
the transformation is identical — 14 485 characters, a byte-for-byte match. Yet
\texttt{flip\_rate} = 1.000 and max|$\Delta$p| = 0.9664: the probability of machine origin falls
from 0.97–1.00 to 0.03–0.23 across all six eligible models.

The result, stated: \textbf{removing emphasis markers did not change the \texttt{full} profile
by a single character, but it changed the distribution of the text between headings
and \texttt{prose}, the values of the format counters and the classifier's input vector. In
all six applicable models this turned a confident machine decision into a human
one.}

The mechanism can be traced through the files. The profiles contain no markdown
at all — preprocessing strips the markup. The transformation changes it in the source
text, and this shifts the boundary between heading and prose: lines like "\foreignlanguage{russian}{Как перевести}
\quotedblbase{}\foreignlanguage{russian}{сеошные}'' \foreignlanguage{russian}{отчеты в деньги}:", previously recognised as bold headings, cease to be such
and move into the prose profile. \texttt{prose} grows from 5540 to
5792 characters, the bold-fragment counter falls from 76 to zero, and the full text
does not change.

The unchanged-input invariant worked correctly: the cell is not flagged
\texttt{input\_unchanged}, because the markup counters changed.

This should be read as sensitivity of the entire measurement pipeline, not as a
direct effect of bold type. The model reacts not only to the words of the text but
also to the preprocessing decisions about which fragments count as prose: the
prose-profile features receive different input, and the format features see zero
emphasis spans instead of seventy-six.

\subsubsection{Limitations of the procedure}

\textbf{The card showed the \texttt{full} profile.} Transformations acting through the
\texttt{prose}/\texttt{full} boundary and through the markup counters are entirely invisible
in it — which is exactly why the t02 case received the verdict "indeterminate". For
this class of effects, blind analysis on a single profile is insensitive.

\textbf{Texts were truncated to 4000 characters.} Truncation equalises volume: in all
thirty cards both texts run to 494–589 words, and the growth in length under expansion
is not visible in the card. Identification of t11 relied on duplication and paragraphs,
not on volume.

\textbf{The composition of the set limits the blinding.} Twenty-six of the thirty cases
fall on a single transformation, so after a few cards the type of impact reconstructs
itself. The limitation was recorded before the analysis began.
A separate test for an order effect was not introduced: it had not been registered.

\textbf{Two cards were relabelled before the table was fixed.} An automatic heuristic
attributed them to a difference in spaces, whereas what is there is movement of blocks;
this was corrected by observation, without opening the key. Both belong to t04, and
after the correction the verdict matched the mechanism.

*Sources: \texttt{07-\allowbreak{}analysis/instability-\allowbreak{}v1-\allowbreak{}blinding-\allowbreak{}spec.md},
\texttt{instability-\allowbreak{}v1-\allowbreak{}blind-\allowbreak{}verdicts.csv}, \texttt{instability-\allowbreak{}v1-\allowbreak{}blind-\allowbreak{}session.json},
\texttt{instability-\allowbreak{}v1-\allowbreak{}blinding-\allowbreak{}key.json}.*

\section{Discussion}

\subsection{1. There is no single measurable "machine-likeness"}

Four procedures assess one corpus on the same documents and disagree in sign.
On the \texttt{P3 $-$ P1} contrast, the style index, the classifier and zero-shot NLL give
a minus, the judge model a plus, and none of the four intervals covers zero. On
the second contrast the signs of procedures 1, 2 and 4 diverge the same way;
there the diverging procedure 3 is indistinguishable from zero (p = 0.1988), so
its contribution to the disagreement is weaker.

This is not an error in one of the procedures that should be found and fixed.
Their inputs differ: two work with the feature matrix, two with raw text; one has
weights fixed before data collection, another has weights trained on the data.
The sign disagreement is a property of the construct, not a measurement failure.

The direct practical consequence: the question "which detector is correct" is
ill-posed. The better question is what exactly each of them operationalises, and
whether their operationalisations coincide in the application of interest.

They cannot be averaged. Index points, class probability, mean token NLL and a
rubric scale are four different scales, and a pooled effect across them is
meaningless. Procedures 1 and 2 are also not independent: both aggregate the same
feature matrix.

\subsection{2. Corpus assembly conditions are stronger than stylistic signals}

The baseline on a single input — the source identifier — reached a median AUROC
of 1.0000 across the thirteen holdouts where AUROC is defined, above the model on twenty-two style features
(0.9973). For comparison: the length baseline gives 0.4364, the negative control
with cluster-level shuffled labels 0.3614.

The source identifier turned out to be an almost deterministic proxy for the
origin class. The corpus itself is built that way: human texts were collected
from platforms, machine texts were produced by generation channels, and no source
supplies both classes.

The practical limit this imposes is hard. The assessment does not transfer to
sources where human and machine texts coexist — and that is exactly the situation
detectors are applied to: one platform, one section, a mixed stream. High metrics
on group holdouts do not justify such transfer.

\subsection{3. The classifier learns corpus mechanisms, not only style}

The stress test produced a chain in which every link is measured.

Verbatim sentence repeats in the corpus occur only in human texts: 17 human
documents exceeded the 0.10 threshold and none of the 1079 machine documents did,
and in the machine part the repeat share is zero for every document. Of the
seventeen candidates, ten turned out to be extraction defects, four were
legitimate properties of the source, and three remained unresolved.

Feature M01 — the standard deviation of cosine similarity between adjacent
sentences — is systematically higher in human texts: median 0.0907 versus 0.0746,
Cliff's $\delta$ = $-$0.78. Its coefficient is negative in all eighteen models, with a
median rank of 4 out of 22 by absolute value.

Sentence duplication raises this quantity by 7.86 $\sigma$ and shifts the decision
towards the "human" class. All 109 decision flips under text expansion go in one
direction.

The mechanism is corpus- and model-specific. It explains the behaviour of this
classifier on this corpus and does not make verbatim repetition a marker of human
writing: the association arose because repeats in the training data came from one
side, mostly as a trace of collection.

\subsection{4. Preprocessing is part of the measurement instrument}

The \texttt{nemotron\_b016\_P1\_r1} case shows this in pure form. Removing emphasis markers
did not change the \texttt{full} profile by a single character — 14 485 characters
byte-for-byte — but changed the distribution of text between headings and
\texttt{prose}, the values of the format counters and the classifier's input vector. In
all six applicable models, a confident machine decision flipped to human: the
probability dropped from 0.97–1.00 to 0.03–0.23.

The mechanism is traceable: lines previously recognised as bold headings stop
being headings and move into the prose profile. \texttt{prose} grows from 5540 to 5792
characters, the emphasis counter falls from 76 to zero.

So the model reacts not only to the words of the text but also to preprocessing
decisions about which fragments count as prose. The data preparation pipeline
enters the measurement instrument on a par with the model, and its version must
be published and pinned the same way as the weights.

The case was found by blind structural attribution: the card showed the \texttt{full}
profile, where there are no differences, and the verdict "the intervention showed
no effect" was a correct observation alongside a decision flip of 0.97.

\subsection{5. Formal register produces a systematic bias, but this is not demographic unfairness}

The registered subgroup disparity rule fired for the \texttt{HH-\allowbreak{}formal-\allowbreak{}register}
subgroup in 5 of 18 holdouts: the subgroup's median FPR is 0.066, with a
difference of 0.043 from the rest of the test set. The aggregate fairness verdict
did not change, but the effect stopped being a property of a single platform.

Calling this demographic unfairness is wrong, and the reason is not caution of
wording. The subgroup coincides with the \texttt{science} genre document for document —
167 texts — so register and genre are not separable on this corpus. The group is
defined by text type, not by a property of people.

This coincidence is not isolated. All three hard-human labels reproduce a genre
in full: \texttt{HH-\allowbreak{}formal-\allowbreak{}register} — 167 \texttt{science} texts, \texttt{HH-\allowbreak{}polished} — 152 \texttt{news},
\texttt{HH-\allowbreak{}translation} — 73 \texttt{translation}. The annotation was conceived as a layer on
top of the corpus, but in fact it repeated the corpus's genre partition, and
separate estimation of register and genre effects on these data is impossible in
principle, not for lack of power. Separating them requires a corpus where the
formal register occurs in several genres — no such sample exists in this study.

The correct name is an error asymmetry by register. It means that the cost of
error is borne by authors of formal scientific texts, but it does not permit
claims about groups of people.

\subsection{6. Good averages hide genre heterogeneity}

Median AUROC of the main model 0.9973, median MCC 0.9338, median FPR 0.0380 — by
average metrics the classifier looks almost error-free.

The spread says otherwise. For \texttt{holdout\_genre\_seo}, AUROC is 0.9433, MCC 0.6560,
FPR 0.3098, TPR@5\%FPR 0.6250 against the median 0.9833. The same holdout is the
worst in calibration: Brier 0.1796, ECE 0.1973, MCE 0.7548.

The coincidence is not accidental and it matters: SEO is problematic in
separation quality and in calibration at once. Publishing one averaged figure for
such a profile means hiding that in one genre holdout out of seven every third
human text receives a machine verdict.

The genre label also hides the spread between platforms within a genre: in the
\texttt{news} genre the median FPR is 0.013, while across the platforms
inside it — 0.033, 0.000 and 0.000.

\subsection{7. Abstaining from low-confidence decisions reduces risk, but no operating point has been chosen}

The risk–coverage curve shows the expected: error falls from 0.0468 at full
coverage to 0.0059 at half coverage. Abstaining from the least confident
decisions works.

The threshold, however, has not been chosen deliberately. There is no justified
number, and setting one after seeing the results would mean fitting it to the
answer. The curve describes the system's behaviour and is not a deployment
recommendation.

Separately: the abstention policy registered for the instrument is formulated in
terms of properties of the input text — length, genre, conflicting signals — not
the predicted probability. The measured coverage does not account for these
abstentions.

Overall calibration is, at that, acceptable — macro Brier 0.0365, ECE robust to
the number of bins in the range 0.0557–0.0581 — but stability of the estimate is
not proof of good calibration, and the heterogeneity from point 6 does not go
away.

\subsection{8. Practical conclusion}

The detector assesses a combination of text, source, formatting and pipeline, not
biological or creative authorship directly.

The study establishes this through four measured facts: the source separates the
classes better than style (point 2), the classifier learned the corpus trace of
repeats (point 3), a change of formatting without a change of text flips the
decision (point 4), quality and calibration depend sharply on genre (point 6).

For application this means the following. A "machine text" verdict is not a claim
about authorship and cannot serve as grounds for accusation. An assessment
obtained on one corpus does not transfer to a stream where both classes come from
one source. A change of publication template or of the text preparation pipeline
changes the verdict without the author's involvement.

None of these conclusions requires assuming that detectors are useless. They
require assuming that a detector measures text production conditions together
with the text itself — which was the study's original hypothesis.

\subsection{9. Limitations and future validation}

The main limitation is named in point 2: \textbf{the classes share no sources.}
Because sources do not overlap between classes, the separate effects of text
origin and platform are not identifiable on these data without additional
untestable assumptions. Future work must include sources that supply both
classes.

\textbf{Transformations must be validated on a separate development panel.} The
attempt to register shortening without duplication showed why: the transformation
was rejected at preflight by a criterion that itself proved ill-matched to the
nature of the quantity, and choosing a threshold after seeing the result is
inadmissible. The panel for validating criteria must be separated from the panel
on which the result is computed.

\textbf{An external corpus is mandatory.} All conclusions were obtained on one corpus
of one language, collected by one procedure. Transfer to another language,
another domain or another generation of models has not been tested.

\textbf{The operationalisation of preprocessing must be stable and published.} Point 4
shows that the decision about the boundary between heading and prose changes the
verdict. Hence the pipeline version is part of the method description, and
comparing detectors without pinning the preprocessing is invalid.

In addition, the known and recorded limitations remain: mixed-effects models did
not yield interpretable estimates for three different reasons, the volume
transformations turned out to be composite, three repeat cases remained
unresolved, and candidate classification was carried out by a single rater.

\section{Limitations}

\subsection{Limitations}

\textbf{PI decision of 29 July:} the main paper gives groups (Table S16); the full list goes to the appendix. The list has 26 items: the one-sided text extraction defect was added on 29 July, items 22–24 followed the stress test on 31 July, and items 25–26 followed the analysis of the instability cards on 1 August.

\begin{center}
\small
\begin{tabularx}{\textwidth}{llX}
\toprule
Group for the main paper & Items it folds in & What it tells the reviewer \\
\midrule
1. Identifiability of the design & 2, 12, 13 & two of three primary outcomes are unidentifiable, the negative control is not feasible, five holdouts are single-class and three of them coincide \\
2. Content validity of the features & 1, 17 & the index fully covers two of thirteen categories, fourteen features were not computed \\
3. Dependence on the procedure & 3, 4, 6, 7, 9, 10 & procedures 2–4 are exploratory, 1 and 2 are one family, net is not available for all, effects are small, Bonferroni correction, different contrast populations \\
4. Source and genre & 11, 14, 15, 19, 21 & the extraction defect is one-sided, missing values carry the source label, four machine clusters, level annotated by source by one person, the human part predates 2022 \\
5. Limited reproducibility of the judge & 8 & ICC(2,1) 0.477, fifteen values of the scale median, 85.9\% at two points \\
6. Absence of a prospective bridge set & 5 & all procedures on one corpus: alternative measurements, not independent samples \\
7. Boundaries of the stress test & 22, 23, 24, 25, 26 & tercile discretisation, format counters normalised to volume, denominator 10 of 16, composite volume transformations, the length sensitivity check rejected at preflight \\
\bottomrule
\end{tabularx}
\end{center}

Items 16, 18 and 20 do not go into the main text and remain in the appendix only.

\textbf{Item 14 is a candidate for a separate subsection, not a footnote.} The defect is one-sided and affects three of the four procedures; the decision on its fate rests with the PI.

The full list follows. The order runs from what limits the main conclusion to the particular.

\subsubsection{Construct and design}

\begin{enumerate}
\item[1.] \textbf{The index covers a small part of the stated construct.} Of thirteen categories, two are fully operationalised, five partially, four are not operationalised, one is reported as a separate output, one is absent. The total weight of the non-operationalised categories in the "Mixed" domain is 0.23 out of 1.00, and it has not been redistributed among the rest. All four share one reason: the source formulates the mechanism as a narrowing of dispersion, while the features measure level. Source: \texttt{construct-\allowbreak{}coverage.md}.
\item[2.] \textbf{Two of the three primary outcomes are unidentifiable.} In the human part \texttt{regulation\_level} coincides with \texttt{genre} one to one; for the O3 groups the genre intersection is empty. O2 and O3 were reclassified as descriptive before the dependent variable was computed; no p-values were calculated for them. Source: \texttt{design-\allowbreak{}support-\allowbreak{}table.md}.
\item[3.] \textbf{Procedures 2–4 have exploratory status.} Their operationalisation was written on 27 July, after the results of procedure 1 had been seen; §7 of the preregistration specified each in a single line. Source: \texttt{procedures-\allowbreak{}2-\allowbreak{}4-\allowbreak{}registration.md} §2.
\item[4.] \textbf{Procedures 1 and 2 are one family of evidence.} Both aggregate the same feature matrix on the same documents and differ in how the weights are obtained. Their agreement counts as one source, not two.
\item[5.] \textbf{All four procedures share one corpus.} These are alternative measurements on the same documents, not independent samples: shared selection biases, genres, time slice and collection defects. Genuine replication would require a bridge set; its requirements are in \texttt{design-\allowbreak{}support-\allowbreak{}table.md} §5.
\item[6.] \textbf{The net estimand cannot be operationalised for procedures 3 and 4.} Removing markup is not the same as subtracting the structure prescribed by the brief: the brief also dictates the order of exposition, the set of sections and the mandatory points. No imitation was created for the sake of table symmetry.
\end{enumerate}

\subsubsection{Magnitude and precision}

\begin{enumerate}
\item[7.] \textbf{All effects are small.} Standardised paired effects of procedure 1: |d\_z| from 0.143 to 0.305. The mean differences $-$1.77 and +3.28 are points of the operationalised index, not percentages of machine-likeness.
\item[8.] \textbf{The judge's resolution is below the effect size.} 15 unique values of the three-run median across 1882 documents, 85.9\% of scores at two points of the scale, ICC(2,1) = 0.477, mean range across three runs 4.78 against an SD of 3.26 for the median machine-part scores.
\item[9.] \textbf{The intervals are unadjusted, 95\%.} Bootstrap replicates were not saved; 97.5\% intervals cannot be extracted from the run. Decisions are made by p against $\alpha$ = 0.025 after Bonferroni. The sensitivity contrast O1-net \texttt{P2 $-$ P1} does not pass the threshold (p = 0.0416) and is not called significant.
\item[10.] \textbf{The 120 classifier pairs are 15 task clusters.} The P2b contrast was computed on a different population than the rest: only the seo genre against 45 clusters for procedures 1, 3 and 4.
\item[11.] \textbf{Four machine source clusters give a weak estimate of cross-channel generalisability} — this limitation is published at every mention of the external P2b folds.
\end{enumerate}

\subsubsection{Diagnostics that did not work}

\begin{enumerate}
\item[12.] \textbf{A cluster-level negative control is not feasible in this design.} Four machine clusters against 124 human ones give a bimodal distribution over twenty permutations (mean AUROC 0.132–0.434 with a spread of 0.001–0.998); the mean is not a null hypothesis. Replacing the cluster permutation with a document permutation would bring the leakage back. The diagnostic is recorded as not feasible.
\item[13.] \textbf{Five of the eighteen holdouts yield a single-class test}, where only the FPR is defined. The test sets \texttt{genre\_news}, \texttt{genre\_science} and \texttt{genre\_translation} coincide in composition — the same 392 documents of series v2, verified by comparing identifiers — because rule §5.2.1 always keeps hard-human out of training. The three table rows read as one observation; the wording "verified on seven genre holdouts" is incorrect. The macro averaging reads the same way: the eighteen holdouts are eighteen registered split procedures with overlapping test parts, not eighteen independent document sets, so equal weight in the macro goes to procedures, not samples.
\end{enumerate}

\subsubsection{Data}

\begin{enumerate}
\item[14.] \textbf{One-sided text extraction defect.} 97 human documents out of 837 contain verbatim repeated sentences; in the machine part, none of the 1079 documents do. Documents with repetition above 10\% are accused by procedure 2 in 49.5\% of cases against 11.6\% for the rest, and for Lenta the separation is complete: 28 of 28 against 0 of 32. The defect biases the class comparison in one direction and affects more than the classifier: \texttt{human\_news\_lenta\_0019} has a mean NLL of 0.886 against a corpus median of 3.09. Found 2026-07-29, scan \texttt{text-\allowbreak{}defects-\allowbreak{}v1.md}. The decision on cleaning the corpus and recomputing rests with the PI. \textbf{Updated 2026-08-01:} the numbers above refer to \texttt{prep-\allowbreak{}v4}, i.e. the state before extraction correction. On the current \texttt{prep-\allowbreak{}v5}, a symmetric scan of both parts of the corpus gives 17 human documents out of 803 against zero machine documents out of 1079 — the correction removed about eighty percent of the cases; the asymmetry remains. Scan v1 did not cover the machine part at all: \texttt{text\_defects\_scan.py:80} selects documents with the condition \texttt{origin\_class == "H"}, so the word "one-sided" in the item's title rested on an unverified comparison; the symmetric computation was performed only on 1 August — \texttt{repeat-\allowbreak{}scan-\allowbreak{}v2-\allowbreak{}report.md}. The classification of the 17 remaining candidates was performed on 1 August by a single rater (10 confirmed defects, 4 legitimate source properties, 3 unresolved — \texttt{repeat-\allowbreak{}scan-\allowbreak{}v2-\allowbreak{}verdicts.md}); the classification has not been independently checked.
\item[15.] \textbf{M02 missing values carry the source label.} In the machine part there are zero; in the human part, 183 of 837, from 100\% for Lenta and Gazeta.ru to 2.1\% for UrFU. The mechanism does not act on O1 — the contrasts are computed within the machine corpus — but for any joint human/AI model this feature works as a source identifier.
\item[16.] \textbf{103 Pilot-1 documents were withdrawn from confirmatory}, of which 100 remain active in the corpus out of 1916 (5.2\%), distributed unevenly: 7.4\% of the human part against 3.5\% of the machine part. The second edition of the set was selected after feature computation, recorded in the amendment of 27 July.
\item[17.] \textbf{Fourteen features were not computed}, all exploratory. D01 and D02 were moved from primary to exploratory for lack of published dictionaries; D01 and D06 were computed from the Grammatika-80 and Vinogradov lists and rejected — 34\% false positives on a sample of 200 occurrences.
\item[18.] \textbf{Raw JSON responses were not saved for all channels.} For \texttt{deepseek\_pro} and \texttt{nemotron} the response body was not saved and cannot be recovered.
\item[19.] \textbf{The regimentation level was annotated by one person at the source level}; there is no independent re-check of the annotation.
\item[20.] \textbf{There is no manual human annotation by design}; inter-rater agreement between humans was not measured.
\item[21.] \textbf{The human part consists of texts written before 2022.} Contemporary human writing is not in the corpus: there is no way to prove its origin.
\end{enumerate}

\subsubsection{Added 31 July 2026 following the stress test}

The rubrics were set by the PI: items 22 and 23 are limitations of measurement and construct validity; item 24 concerns the scope of the robustness check and external validity.

\textbf{Limitations of measurement and construct validity}

\begin{enumerate}
\item[22.] \textbf{Hard tercile discretisation amplifies small numerical changes near the boundaries.} \texttt{tercile\_score} cuts the rank at 1/3 and 2/3, so an input that changed slightly produces a stepwise jump in the index. Confirmed by the computational reproducibility diagnostic: in the 30 July measurement, out of 1853 checks of features M01, M02 and M05 over the panel, 30 values lie within one rank step of a boundary and 9 exactly on it (\texttt{status-\allowbreak{}2026-\allowbreak{}07-\allowbreak{}30.md}). This is not an estimate of the effect of the excluded transformation t14.
\end{enumerate}

\begin{enumerate}
\item[23.] \textbf{Format counters are normalised to volume.} R06, R07 and F01 are computed per 1000 words of the \texttt{full} profile, P01 as the share of the profile outside connected prose. A change in text length can therefore shift the format score without any change in the absolute number of markup elements. The limitation concerns the operationalisation of the features itself and holds regardless of which transformations entered the final set. The specific numbers on which the mechanism was discovered are moved to the reproducibility appendix with the label invalidated diagnostic run: they were obtained on a run with a transformation recognised as defective and are not a stress-test result.
\end{enumerate}

\textbf{Scope of the robustness check and external validity}

\begin{enumerate}
\item[24.] \textbf{The final stress test covers 10 of the 16 registered transformations.} Six intervention-dependent or technically infeasible transformations were not evaluated: five require material the corpus does not contain (the A$\to$H branch, machine rewriting, translation and back-translation, paraphrasing), and the sixth — t14, "removal of typical clich\'es" — was moved to \texttt{not executable} on 31 July because after \texttt{prep-\allowbreak{}v5} preprocessing there is nothing to remove: the registered list of artefacts yields zero hits on the panel. The denominator is 10; there are 600 primary cells. In the paper the result is called a 10-transformation stress test. Amendment: \texttt{amendment-\allowbreak{}stress-\allowbreak{}r5-\allowbreak{}t14-\allowbreak{}not-\allowbreak{}executable.md}.
\end{enumerate}

\subsubsection{Added 1 August 2026 following the review of the instability cards}

\begin{enumerate}
\item[25.] \textbf{Transformations 10 and 11 are composite, and the effect of volume was not tested by them in isolation.} Both functions split the text into sentences by \texttt{\textbackslash{}s+} and reassemble it with spaces, so three things change at once: length, verbatim duplication or deletion of sentences, and paragraph markup — the median loss is 83.5\% of paragraphs for shortening and 75.6\% for expansion. The contributions cannot be separated on these data, and three consequences are recorded directly. First: the effect of volume without duplication has not been tested; the panel contains no transformation that changes length without repeating sentences. Second: the independent effect of paragraph structure is not identified. Third: the direction of the classifier's response is tied to how verbatim repetitions are distributed in the training corpus. A symmetric scan of both parts (\texttt{repeat-\allowbreak{}scan-\allowbreak{}v2-\allowbreak{}report.md}) gives an automatic result: \textbf{17 human and 0 machine documents exceeded the 0.10 threshold on \texttt{prep-\allowbreak{}v5}}. On \texttt{prep-\allowbreak{}v4} the ratio is the same and sharper — 97 against zero out of 1079; the historical number reproduced exactly; in the machine part the repetition share is zero for all documents, not merely below the threshold. Beyond this the automatic criterion does not go, and the candidates were classified manually into three classes (\texttt{repeat-\allowbreak{}scan-\allowbreak{}v2-\allowbreak{}verdicts.md}): 10 documents \texttt{confirmed\_defect}, 4 \texttt{legitimate\_source\_property}, 3 \texttt{unresolved}. The classification was done by a single rater, and this is a limitation of the result. The confirmed defects are block repetitions of fragments in the intermediary dataset \texttt{buriy\_2014} and a cookie-plugin line; the legitimate repetitions are a doubled bibliography in three SPbU articles, where \texttt{References} splits the repetitions exactly in half, and a block of interview questions. A separate mechanism is recorded for \texttt{human\_news\_buriy\_2014\_0048}: the detachment of the heading by preprocessing made a sentence verbatim equal to the next one, and the repetition appeared where the raw file had none. The material for verification is \texttt{repeat-\allowbreak{}scan-\allowbreak{}v2-\allowbreak{}manual.md}, which gives verbatim fragments and structural context for each candidate but no procedure 2 scores, so that the classification would not depend on them. What is established for procedure 2 and for it alone: the decision flip is predominantly associated with an M01 increase of 7.86 $\sigma$, and the shift goes towards the "human" class. Under the name M01 the model receives not the mean similarity of adjacent sentences but its standard deviation: \texttt{extract\_semantic.py} lines 290–292 write the mean to \texttt{raw\_value} and the dispersion to \texttt{normalized\_value}, while \texttt{clf\_run.load\_matrix} reads \texttt{normalized\_value or raw\_value}. Duplication places identical phrases side by side with a cosine of 1.0 against a typical 0.5 and raises precisely the dispersion. Why this pulls towards humans is shown by \texttt{m01-\allowbreak{}direction.md}: in the training part the dispersion is systematically higher for human texts — median 0.0907 against 0.0746, Cliff's $\delta$ = $-$0.78 as the median over 18 holdouts — and the M01 coefficient is negative in all eighteen models, with a median rank of 4 out of 22 by absolute value. The mechanism is stated as follows: the model associates a larger dispersion of cosine similarities of adjacent sentences with the human class; verbatim duplication creates extreme adjacent pairs, raises M01 and shifts the decision in that direction. All 109 decision flips go exactly there. The mechanism is corpus- and model-dependent: it describes this classifier on this corpus and does not make verbatim repetition a universal marker of human text. This is a property of this classifier on this corpus, not of duplication as a marker of authorship; for procedure 1 the mechanism requires algebraic decomposition, and for the NLL and the judge it remains unknown. The earlier interpretation via volume is withdrawn and moved to the reproducibility appendix. The analysis: \texttt{07-\allowbreak{}analysis/stress-\allowbreak{}t10-\allowbreak{}t11-\allowbreak{}volume-\allowbreak{}defect.md}.
\end{enumerate}

\begin{enumerate}
\item[26.] \textbf{The effect of shortening length without a concomitant change in the M01 distribution could not be tested.} A separate transformation was registered for this — "paragraph-preserving truncation without verbatim duplication", shortening by roughly a quarter while preserving paragraph markup and without verbatim duplication (\texttt{amendment-\allowbreak{}sensitivity-\allowbreak{}truncation.md}). It passed three of the four acceptance criteria: markup preserved in all sixty panel documents, no duplication introduced, median volume ratio 0.7904. The fourth criterion, written down before the computation, was violated: the median $\Delta$z for M01 was +0.0079 against the requirement "not above zero", and in ten documents the increase exceeded 0.5 $\sigma$ with a maximum of +2.855. The transformation was rejected at preflight; procedure 2 was not run on it; there are no sensitivity results. The criterion was not relaxed after seeing the result: an alternative formulation via a systematic shift would have passed on these data for certain, i.e. it would have been fitted to an already known answer. \textbf{This is not a null result on the effect of length but the infeasibility of this operationalisation.} A diagnostic obtained after the gate had fired: the shift is symmetric — an increase in thirty documents, a decrease in thirty — whereas for transformation 11 the median $\Delta$z is +7.86 with an increase in all sixty; the largest deviations fall on short documents, because M01 is itself a sample standard deviation over twenty to seventy pairs of adjacent sentences. Returning to the question requires a new transformation whose criteria are checked on a separate development panel. The decision and the numbers: \texttt{07-\allowbreak{}analysis/sensitivity-\allowbreak{}v1-\allowbreak{}rejection.json}.
\end{enumerate}

\section{Declarations}

\subsection{0. Title block}

Fixed by the PI on 2026-08-01, without a country:

\begin{verbatim}
Stanislav Kirichenko
Independent Researcher
Email: staurus86@gmail.com
ORCID: https://orcid.org/0009-0001-2914-9541
Corresponding author
\end{verbatim}

\begin{center}
\small
\begin{tabularx}{\textwidth}{lX}
\toprule
Field & Status \\
\midrule
Name in Latin script & \texttt{Stanislav Kirichenko} — checked against the ORCID profile \\
Affiliation & \texttt{Independent Researcher} — the status in the ORCID profile \\
ORCID & \texttt{0009-\allowbreak{}0001-\allowbreak{}2914-\allowbreak{}9541}, public profile \\
Corresponding author & Stanislav Kirichenko \\
Email & \texttt{staurus86@gmail.com} — not public in the ORCID profile; the PI provides it for submission \\
City and country & not listed in the profile \\
\bottomrule
\end{tabularx}
\end{center}

\textbf{The country is omitted by decision of the PI.} The ORCID profile lists Russia,
Ukraine and Belarus — this is the profile's list of countries, not a place of work.
arXiv does not require an address; if the Cambridge editorial office requests it at
submission, the field is filled in then.

To check before submission: Cambridge Natural Language Processing and arXiv differ
in their requirements for field order and for whether ORCID is mandatory.

\subsection{1. Acknowledgements}

The author thanks the authors of open sources whose publications entered the
sample under the source's licence. Recruitment of participants who would
contribute texts was planned but did not take place: there are no participant
acknowledgements because there were no participants.

There are no acknowledgements by name. If the author decides to name someone, the
name is added here before submission.

\subsection{2. CRediT author statement}

The work was carried out by a single author, so all roles belong to him:

\textbf{Stanislav Kirichenko:} Conceptualization, Methodology, Software, Validation,
Formal analysis, Investigation, Data curation, Visualization, Writing — original
draft, Writing — review and editing.

The roles Funding acquisition and Project administration are not claimed: the work
has no external funding, which is confirmed by the PI and recorded in the Funding
statement.

\subsection{3. Funding statement}

Confirmed by the PI on 2026-08-01: there was no funding; the work was done on the
author's personal time and personal funds, including paid access to two generation
channels.

\textbf{Published text:}

\begin{quote}
This research received no specific grant from any funding agency in the public,
commercial, or not-for-profit sectors. The study was conducted independently using
the author's personal time and resources.
\end{quote}

\subsection{4. Competing interests}

The previous version stated that the author does not own a detector. That was wrong
and has been removed: procedure 1 — the interpretable style index — is built on the
author's package \texttt{ai-\allowbreak{}human-\allowbreak{}style-\allowbreak{}lab-\allowbreak{}v0.2.0}, which sits inside the project and is
referenced by \texttt{statistical-\allowbreak{}analysis-\allowbreak{}plan.md}, \texttt{codebook.md}, \texttt{metrics-\allowbreak{}spec.md} and
the verdict-refusal policy. Owning one's own tool does not discredit the study, but
it is disclosed explicitly.

Confirmed by the PI on 2026-08-01: the detector has not been commercially licensed.
A second fact was disclosed at the same time — the author contributed six documents
to the human part of the corpus, which he himself annotated and analysed.

\textbf{Published text:}

\begin{quote}
The author developed and owns the text-detection system evaluated as one of the
four operationalizations and contributed six pre-existing documents to the human
reference corpus. The detector has not been commercially licensed. These
relationships are disclosed because the author also performed the corpus
annotation and analysis.
\end{quote}

What works against a suspicion of bias: the author's own procedure is not designated
as the reference, and results unfavourable to it are published on equal terms with
the rest. For the style index, the \texttt{P3 $-$ P1} contrast changes sign when moving from
the full estimand to net — from $-$1.7655 to +2.0264 — and this is printed in Table 3,
not hidden. The work proposes no new detector and does not improve existing ones.

\subsection{5. Data and code availability}

\textbf{What is published.} The feature matrix, result files of all runs, manifests with
hashes of inputs, code and splits, procedure specifications, the preregistration and
amendments, the computational code in full, and the scripts that build the figures
and tables.

\textbf{What is published as links rather than copies.} The hard-human texts from open
sources: they are distributed under the licence of their source, and the study's
author has no right to republish them. For each document the URL, the access date
and the hash of the normalised text are published, from which the extraction is
reproduced.

\textbf{What is never published.} Participants' personal data: names, contact details,
authorship declarations with identifying information. They are stored separately
from the corpus and are not part of the repository.

\textbf{The machine part of the corpus} is published where the model provider's policy
permits it; the policy has been checked separately for each generation channel.

Reproducibility does not rest on the texts alone: the manifests record the hashes of
inputs and code, so the computation can be verified on the published features even
where the source text is closed by licence.

\textbf{The platform was chosen by the PI on 2026-08-01: GitHub with a release archived in
Zenodo.} The repository receives a DOI through Zenodo, and the paper prints the DOI
of the release, not a link to a branch: a branch changes, while the DOI points to an
immutable snapshot.

\textbf{Published text:}

\begin{quote}
Code, feature matrices, run manifests with input and code hashes, preregistration
and amendments are available in a public GitHub repository, archived in Zenodo
(concept DOI 10.5281/zenodo.21749574, which resolves to the latest version;
the v1.0.0 release is archived under version DOI 10.5281/zenodo.21750014).
Source texts of the
human subcorpus are distributed as links rather than copies: they remain under the
licences of their sources. For each such document the repository publishes the
URL, the access date and the hash of the normalised text, which is sufficient to
reproduce the feature extraction.
\end{quote}

The repository was published on 2026-08-02:
\url{https://github.com/staurus86/beyond-authorship-russian-ai-detection}, commit
\texttt{1f0da5b6}. Zenodo archival record: the concept DOI \texttt{10.5281/zenodo.21749574} resolves
to the latest version; the first published version is \texttt{10.5281/zenodo.21749575}. The
paper prints the concept DOI.

\subsection{6. Ethics statement}

Ethics committee approval was not required. The study works with published
texts: the human part of the corpus is drawn from open sources, and six
documents come from the author's personal archive that predates the study. No
experiments on humans were conducted, and personal data neither enter the
analysis nor are published.

Recruitment of participants who would contribute texts under written consent
was planned — the consent forms and the registry were prepared and are
published in the repository — but did not take place: no text from external
participants entered the corpus, and no payments were made.

\textbf{The author considers the risk of applying the results to be the main ethical
question of this work.} Machine-text detectors are used for accusations, and a
type I error costs a person their degree or their payment, while they have no
systematic way to prove their authorship. The work therefore publishes FPR
separately for vulnerable subgroups rather than average accuracy alone; describes
refusal to issue a verdict as a normal operating mode; and states directly that none
of the four procedures is fit to serve as the sole basis for an accusation.

\textbf{Dual use.} The features by which a text is recognised as machine-generated also
help to disguise it. The work does not optimise detection evasion and does not
publish a ready evasion pipeline; the text transformations are set up as robustness
stress tests and described as such.

The work draws no conclusions about the language competence, education or origin of
the authors.

\subsection{7. AI-use disclosure}

The author used language models as a tool throughout the work and discloses this to
the extent required by arXiv and Cambridge.

The previous version stated that the models took no part in design decisions. That
is stronger than what actually happened: the models proposed operationalisation
options, critiqued specifications and found defects. The final decisions were made
by the author.

\textbf{Published text:}

\begin{quote}
Generative AI tools were used to propose and critique methodological alternatives,
assist with code development and debugging, perform secondary consistency checks,
and support language editing. All methodological choices were reviewed and
approved by the author. No numerical result was accepted without programmatic
verification against the frozen source artifacts. The author assumes full
responsibility for the study design, analyses, interpretations, and manuscript.
\end{quote}

\textbf{The tools and what is recorded about them.} The work ran from 24 July to
2 August 2026; session dates are reconstructed from the \texttt{00-\allowbreak{}admin/status-\allowbreak{}*.md} logs.

\begin{center}
\small
\begin{tabularx}{\textwidth}{lX}
\toprule
Tool & What the logs record \\
\midrule
Claude in the Claude Code interface & the version is named in the entry of 2026-08-01: Claude Opus 5; the session of 2026-08-02 — Claude Fable 5. For the sessions of 24–31 July the model version was not entered in the log \\
Codex & the name and role are recorded — independent read-only verification of numbers, 31 July and 1 August. The version is not recorded: the interface did not display it \\
\bottomrule
\end{tabularx}
\end{center}

Exactly this is published, without reconstruction from memory: there is no way to
name the model version for the early sessions, and substituting today's version
would be a guess.

\textbf{Translation of the manuscript.} The English version of the manuscript was
obtained by machine translation of the frozen Russian text (Claude Fable 5,
2 August 2026) with automatic verification gates: numbers, identifiers,
structure and code spans were checked against the original byte for byte, and
a semantic fidelity review of key sections was performed as a separate pass.
The Russian text remains the source of record.

\textbf{Published wording on the tools:}

\begin{quote}
Work was carried out between 24 July and 2 August 2026 using Claude in the Claude
Code interface (model version recorded in the project log for the final sessions:
Claude Opus 5 on 1 August, Claude Fable 5 on 2 August) and Codex for independent
read-only verification of reported numbers (version not recorded; the interface
did not display it). The English version of the manuscript was machine-translated
from the frozen Russian text (Claude Fable 5, 2 August 2026) under scripted
verification gates: numbers, identifiers, structure and code spans were checked
against the source byte for byte, and key sections underwent a separate semantic
fidelity review. The Russian text remains the source of record.
\end{quote}

The specific episodes — what discrepancy the independent verification found and
when — remain in the reproducibility appendix, not in this declaration.

\textbf{A separate note on the subject of the work.} The study examines the assessment of
machine-likeness of text, and part of its own text was written with a model's
involvement. There is no contradiction here: the work does not claim that machine
origin makes a text worse — it shows that machine-likeness assessment procedures
react to text production conditions, not only to origin.

\section{References}

The list contains 68 sources — all that are cited in the text. A further 5 records of the systematic sample are not cited in the manuscript and are not included in the list: the completeness of the search matrix is documented separately, in \texttt{01-\allowbreak{}literature/evidence-\allowbreak{}matrix.csv}.

\noindent\hangindent=1.5em Abassy et al., 2024. Abassy, M., Elozeiri, K., Aziz, A., Ta, M. N., Tomar, R. V., Adhikari, B., Ahmed, S. E. D., Wang, Y., Afzal, O. M., Xie, Z., Mansurov, J., Artemova, E., Mikhailov, V. et al. (2024). LLM-DetectAIve: a Tool for Fine-Grained Machine-Generated Text Detection. EMNLP 2024: System Demonstrations. \url{https://aclanthology.org/2024.emnlp-demo.35/}\par\smallskip

\noindent\hangindent=1.5em Artemova et al., 2025. Artemova, E., Lucas, J. S., Venkatraman, S., Lee, J., Tilga, S., Uchendu, A., Mikhailov, V (2025). Beemo: Benchmark of Expert-edited Machine-generated Outputs. NAACL 2025. \url{https://aclanthology.org/2025.naacl-long.357/}\par\smallskip

\noindent\hangindent=1.5em Bakhshi, 2026. Bakhshi, A. D (2026). Saying More Than They Know: A Framework for Quantifying Epistemic-Rhetorical Miscalibration in Large Language Models. arXiv:2604.19768, submitted 2026-03-27; \foreignlanguage{russian}{подана в} Assessing Writing (Elsevier). \url{https://arxiv.org/abs/2604.19768}\par\smallskip

\noindent\hangindent=1.5em Bao et al., 2024. Bao, G., Zhao, Y., Teng, Z., Yang, L., Zhang, Y (2024). Fast-DetectGPT: Efficient Zero-Shot Detection of Machine-Generated Text via Conditional Probability Curvature. ICLR 2024; arXiv:2310.05130, v1 2023-10-08, v3 2024-12-16. \url{https://openreview.net/forum?id=Bpcgcr8E8Z}\par\smallskip

\noindent\hangindent=1.5em Bao et al., 2025. Bao, G., Rong, L., Zhao, Y., Zhou, Q., Zhang, Y (2025). Decoupling Content and Expression: Two-Dimensional Detection of AI-Generated Text. arXiv:2503.00258, submitted 2025-03-01. \url{https://arxiv.org/abs/2503.00258}\par\smallskip

\noindent\hangindent=1.5em Basani et al., 2025. Basani, A. R., Chen, P.-Y (2025). Diversity Boosts AI-Generated Text Detection. TMLR '26; arXiv:2509.18880, submitted 2025-09-23. \url{https://arxiv.org/abs/2509.18880}\par\smallskip

\noindent\hangindent=1.5em Basu et al., 2026. Basu, P., Zhang, Y., Raheja, V (2026). BAID: A Benchmark for Bias Assessment of AI Detectors. Proceedings of the Second Workshop on Customizable NLP (CustomNLP4U), \foreignlanguage{russian}{июль} 2026, San Diego. \url{https://aclanthology.org/2026.customnlp4u-1.1/}\par\smallskip

\noindent\hangindent=1.5em Batura et al., 2025. Batura, T., Bruches, E., Shvenk, M., Malykh, V (2025). AINL-Eval 2025 Shared Task: Detection of AI-Generated Scientific Abstracts in Russian. AINL 2025; arXiv:2508.09622, submitted 2025-08-13. \url{https://arxiv.org/abs/2508.09622}\par\smallskip

\noindent\hangindent=1.5em Baumler et al., 2026. Baumler, C., Bao, C., Nghiem, H., Yang, X., Carpuat, M., Daum\'e III, H (2026). Can You Make It Sound Like You? Post-Editing LLM-Generated Text for Personal Style. ACL 2026; arXiv:2604.24444, submitted 2026-04-27. \url{https://arxiv.org/abs/2604.24444}; \foreignlanguage{russian}{предрегистрация} aspredicted.org/ck6xk4.pdf\par\smallskip

\noindent\hangindent=1.5em Borile and Abrate, 2025. Borile, C., Abrate, C (2025). How to Generalize the Detection of AI-Generated Text: Confounding Neurons. Findings of the ACL: EMNLP 2025. \url{https://aclanthology.org/2025.findings-emnlp.1388/}\par\smallskip

\noindent\hangindent=1.5em Briakou et al., 2021. Briakou, E., Agrawal, S., Zhang, K., Tetreault, J., Carpuat, M (2021). A Review of Human Evaluation for Style Transfer. First Workshop on Natural Language Generation, Evaluation, and Metrics (GEM), 2021. \url{https://aclanthology.org/2021.gem-1.6/}\par\smallskip

\noindent\hangindent=1.5em Chen et al., 2025. Chen, X., Wu, J., Yang, S., Zhan, R., Wu, Z., Luo, Z., Wang, D., Yang, M., Chao, L. S., Wong, D. F (2025). RepreGuard: Detecting LLM-Generated Text by Revealing Hidden Representation Patterns. Transactions of the ACL (TACL), Vol. 13, pp. 1812–1831. \url{https://aclanthology.org/2025.tacl-1.81/}\par\smallskip

\noindent\hangindent=1.5em Covington and McFall, 2010. Covington, M. A., McFall, J. D (2010). Cutting the Gordian Knot: The Moving-Average Type–Token Ratio (MATTR). Journal of Quantitative Linguistics, 2010, 17(2), 94–100, DOI 10.1080/09296171003643098.\par\smallskip

\noindent\hangindent=1.5em Creo and Pudasaini, 2025. Creo, A., Pudasaini, S (2025). SilverSpeak: Evading AI-Generated Text Detectors using Homoglyphs. Proceedings of the 1st Workshop on GenAI Content Detection (GenAIDetect), \foreignlanguage{russian}{январь} 2025. \url{https://aclanthology.org/2025.genaidetect-1.1/}\par\smallskip

\noindent\hangindent=1.5em Desaire et al., 2023. Desaire, H., Chua, A. E., Isom, M., Jarosova, R., Hua, D (2023). Distinguishing academic science writing from humans or ChatGPT with over 99\% accuracy using off-the-shelf machine learning tools. Cell Reports Physical Science, 2023, 4(6), 101426, DOI 10.1016/j.xcrp.2023.101426.\par\smallskip

\noindent\hangindent=1.5em Devatine and Abraham, 2024. Devatine, N., Abraham, L (2024). Assessing Human Editing Effort on LLM-Generated Texts via Compression-Based Edit Distance. arXiv:2412.17321, submitted 2024-12-23. \url{https://arxiv.org/abs/2412.17321}\par\smallskip

\noindent\hangindent=1.5em Dugan et al., 2024. Dugan, L., Hwang, A., Trhl\'ik, F., Zhu, A., Ludan, J. M., Xu, H., Ippolito, D., Callison-Burch, C (2024). RAID: A Shared Benchmark for Robust Evaluation of Machine-Generated Text Detectors. ACL 2024 (62nd Annual Meeting). \url{https://aclanthology.org/2024.acl-long.674/}\par\smallskip

\noindent\hangindent=1.5em El Attar et al., 2026. El Attar, Y., D\"onmez, E., Maurer, M., Falenska, A (2026). A Systematic Analysis of Linguistic Features in AI-Generated Text Detection Across Domains and Models. arXiv:2606.04177, submitted 2026-06-02. \url{https://arxiv.org/abs/2606.04177}\par\smallskip

\noindent\hangindent=1.5em Fredrick and Craven, 2025. Fredrick, D. R., Craven, L (2025). Lexical diversity, syntactic complexity, and readability: a corpus-based analysis of ChatGPT and L2 student essays. Frontiers in Education, 2025, 10:1616935. \url{https://www.frontiersin.org/journals/education/articles/10.3389/feduc.2025.1616935/full}\par\smallskip

\noindent\hangindent=1.5em Geng and Trotta, 2025. Geng, M., Trotta, R (2025). Human-LLM Coevolution: Evidence from Academic Writing. arXiv:2502.09606, v1 2025-02-13, v2 2025-02-17. \url{https://arxiv.org/abs/2502.09606}\par\smallskip

\noindent\hangindent=1.5em Georgiou, 2025. Georgiou, G. P (2025). Differentiating Between Human-Written and AI-Generated Texts Using Automatically Extracted Linguistic Features. Information, 2025; arXiv:2407.03646, submitted 2024-07-04. \url{https://arxiv.org/abs/2407.03646}\par\smallskip

\noindent\hangindent=1.5em Gu et al., 2026. Gu, Y., Li, S., Hu, X (2026). MASH: Evading Black-Box AI-Generated Text Detectors via Style Humanization. Findings of the ACL: ACL 2026. \url{https://aclanthology.org/2026.findings-acl.1487/}\par\smallskip

\noindent\hangindent=1.5em Hans et al., 2024. Hans, A., Schwarzschild, A., Cherepanova, V., Kazemi, H., Saha, A., Goldblum, M., Geiping, J., Goldstein, T (2024). Spotting LLMs With Binoculars: Zero-Shot Detection of Machine-Generated Text. Proceedings of the 41st International Conference on Machine Learning (ICML 2024). \url{https://proceedings.mlr.press/v235/hans24a.html}\par\smallskip

\noindent\hangindent=1.5em Huang et al., 2024. Huang, G., Zhang, Y., Li, Z., You, Y., Wang, M., Yang, Z (2024). Are AI-Generated Text Detectors Robust to Adversarial Perturbations?. ACL 2024, main conference; arXiv:2406.01179, v1 2024-06-03, \foreignlanguage{russian}{ревизия} 2024-06-26. \url{https://github.com/CarlanLark/Robust-AIGC-Detector}\par\smallskip

\noindent\hangindent=1.5em Jabarian and Imas, 2025. Jabarian, B., Imas, A (2025). Artificial Writing and Automated Detection. BFI Working Paper No. 2025-116, \foreignlanguage{russian}{октябрь} 2025; NBER Working Paper No. 34223.\par\smallskip

\noindent\hangindent=1.5em Jafaritazehjani et al., 2020. Jafaritazehjani, S., Lecorv\'e, G., Lolive, D., Kelleher, J (2020). Style versus Content: A distinction without a (learnable) difference?. COLING 2020, pp. 2169–2180. \url{https://aclanthology.org/2020.coling-main.197/}\par\smallskip

\noindent\hangindent=1.5em Kobak et al., 2025. Kobak, D., Gonz\'alez-M\'arquez, R., Horv\'at, E.-\'A., Lause, J (2025). Delving into LLM-assisted writing in biomedical publications through excess vocabulary. Science Advances, 2025-07-02, Vol. 11, No. 27; arXiv:2406.07016, submitted 2024-06-11. \url{https://arxiv.org/abs/2406.07016}\par\smallskip

\noindent\hangindent=1.5em Krishna et al., 2023. Krishna, K., Song, Y., Karpinska, M., Wieting, J., Iyyer, M (2023). Paraphrasing evades detectors of AI-generated text, but retrieval is an effective defense. NeurIPS 2023; arXiv:2303.13408, v1 2023-03-23, v2 2023-10-18. \url{https://arxiv.org/abs/2303.13408}\par\smallskip

\noindent\hangindent=1.5em Lee et al., 2022. Lee, M., Liang, P., Yang, Q (2022). CoAuthor: Designing a Human-AI Collaborative Writing Dataset for Exploring Language Model Capabilities. CHI 2022; arXiv:2201.06796, v1 2022-01-18, v2 2022-01-25. \url{https://arxiv.org/abs/2201.06796}\par\smallskip

\noindent\hangindent=1.5em Liang et al., 2023. Liang, W., Yuksekgonul, M., Mao, Y., Wu, E., Zou, J (2023). GPT detectors are biased against non-native English writers. Patterns; arXiv:2304.02819, v1 2023-04-06, v3 2023-07-10. \url{https://doi.org/10.1016/j.patter.2023.100779}\par\smallskip

\noindent\hangindent=1.5em Liang et al., 2024. Liang, W., Izzo, Z., Zhang, Y., Lepp, H., Cao, H., Zhao, X., Chen, L., Ye, H., Liu, S., Huang, Z., McFarland, D. A., Zou, J. Y (2024). Monitoring AI-Modified Content at Scale: A Case Study on the Impact of ChatGPT on AI Conference Peer Reviews. ICML 2024; arXiv:2403.07183, v1 2024-03-11, v4 2026-05-19. \url{https://arxiv.org/abs/2403.07183}\par\smallskip

\noindent\hangindent=1.5em Macko et al., 2023. Macko, D., Moro, R., Uchendu, A., Lucas, J. S., Yamashita, M., Pikuliak, M., Srba, I., Le, T., Lee, D., Simko, J., Bielikova, M (2023). MULTITuDE: Large-Scale Multilingual Machine-Generated Text Detection Benchmark. EMNLP 2023; arXiv:2310.13606, submitted 2023-10-20. \url{https://arxiv.org/abs/2310.13606}\par\smallskip

\noindent\hangindent=1.5em Macko et al., 2025. Macko, D., Kop\'al, J., Moro, R., Srba, I (2025). MultiSocial: Multilingual Benchmark of Machine-Generated Text Detection of Social-Media Texts. ACL 2025 (63rd Annual Meeting). \url{https://aclanthology.org/2025.acl-long.36/}\par\smallskip

\noindent\hangindent=1.5em Mady et al., 2026. Mady, M., Reschke, J., Schuller, B (2026). Feature-Augmented Transformers for Robust AI-Text Detection Across Domains and Generators. ICML 2026; arXiv:2605.03969, submitted 2026-05-05. \url{https://arxiv.org/abs/2605.03969}\par\smallskip

\noindent\hangindent=1.5em Marchenko Breneur et al., 2026. Marchenko Breneur, O., Danilov, A., Nourbakhsh, A., Lamsiyah, S (2026). NOTAI.AI: Explainable Detection of Machine-Generated Text via Curvature and Feature Attribution. arXiv:2603.05617, submitted 2026-03-05. \url{https://arxiv.org/abs/2603.05617}\par\smallskip

\noindent\hangindent=1.5em McCarthy and Jarvis, 2010. McCarthy, P. M., Jarvis, S (2010). MTLD, vocd-D, and HD-D: A validation study of sophisticated approaches to lexical diversity assessment. Behavior Research Methods, 2010, 42(2), 381–392, DOI 10.3758/brm.42.2.381.\par\smallskip

\noindent\hangindent=1.5em Mitchell et al., 2023. Mitchell, E., Lee, Y., Khazatsky, A., Manning, C. D., Finn, C (2023). DetectGPT: Zero-Shot Machine-Generated Text Detection using Probability Curvature. Proceedings of the 40th International Conference on Machine Learning (ICML 2023). \url{https://proceedings.mlr.press/v202/mitchell23a.html}\par\smallskip

\noindent\hangindent=1.5em Mukherjee et al., 2025. Mukherjee, S., Ojha, A. K., McCrae, J. P., Du\v{s}ek, O (2025). Evaluating Text Style Transfer Evaluation: Are There Any Reliable Metrics?. NAACL 2025, Student Research Workshop, pp. 418–434. \url{https://aclanthology.org/2025.naacl-srw.41/}\par\smallskip

\noindent\hangindent=1.5em Ostheimer et al., 2023. Ostheimer, P., Nagda, M., Kloft, M., Fellenz, S (2023). A Call for Standardization and Validation of Text Style Transfer Evaluation. Findings of the ACL 2023, pp. 10791–10815. \url{https://aclanthology.org/2023.findings-acl.687/}\par\smallskip

\noindent\hangindent=1.5em Paneru, 2026. Paneru, U (2026). Please Make it Sound like Human: Encoder-Decoder vs. Decoder-Only Transformers for AI-to-Human Text Style Transfer. arXiv:2604.11687, submitted 2026-04-13. \url{https://arxiv.org/abs/2604.11687}\par\smallskip

\noindent\hangindent=1.5em Pauli et al., 2025. Pauli, A. B., Augenstein, I., Assent, I (2025). Mind the Style Gap: Meta-Evaluation of Style and Attribute Transfer Metrics. Findings of the ACL: EMNLP 2025, pp. 21550–21564. \url{https://aclanthology.org/2025.findings-emnlp.1175/}\par\smallskip

\noindent\hangindent=1.5em Przystalski et al., 2026. Przystalski, K., Argasi\'nski, J. K., Grabska-Gradzi\'nska, I., Ochab, J. K (2026). Stylometry recognizes human and LLM-generated texts in short samples. Expert Systems with Applications 296, 129001 (2026); arXiv:2507.00838, v1 2025-07-01, v2 2025-07-15. \url{https://arxiv.org/abs/2507.00838}\par\smallskip

\noindent\hangindent=1.5em Reinhart et al., 2025. Reinhart, A., Markey, B., Laudenbach, M., Pantusen, K., Yurko, R., Weinberg, G., Brown, D. W (2025). Do LLMs write like humans? Variation in grammatical and rhetorical styles. PNAS 122 (2025), e2422455122; arXiv:2410.16107, v1 2024-10-21, v2 2025-08-21. \url{https://arxiv.org/abs/2410.16107}\par\smallskip

\noindent\hangindent=1.5em Ren et al., 2026. Ren, K., Raghavan, M., Garg, N (2026). Hitting a Moving Target: Test-Time Adaptation for AI Text Detection under Continual Distribution Shift. arXiv:2606.25152, submitted 2026-06-23. \url{https://arxiv.org/abs/2606.25152}\par\smallskip

\noindent\hangindent=1.5em Richburg et al., 2024. Richburg, A., Bao, C., Carpuat, M (2024). Automatic Authorship Analysis in Human-AI Collaborative Writing. LREC-COLING 2024, Torino, pp. 1845–1855. \url{https://aclanthology.org/2024.lrec-main.165/}\par\smallskip

\noindent\hangindent=1.5em Rodrigues et al., 2026. Rodrigues, F. A., Sturm, N. F., Pinheiro, F. L (2026). A linguistic comparison between human- and AI-generated content. iScience, 2026 Feb 12; 29(3):114976. \url{https://pmc.ncbi.nlm.nih.gov/articles/PMC12969083/}\par\smallskip

\noindent\hangindent=1.5em Russell et al., 2025. Russell, J., Karpinska, M., Iyyer, M (2025). People who frequently use ChatGPT for writing tasks are accurate and robust detectors of AI-generated text. ACL 2025. \url{https://aclanthology.org/2025.acl-long.267/}\par\smallskip

\noindent\hangindent=1.5em Sadasivan et al., 2023. Sadasivan, V. S., Kumar, A., Balasubramanian, S., Wang, W., Feizi, S (2023). Can AI-Generated Text be Reliably Detected?. Transactions on Machine Learning Research (TMLR); arXiv:2303.11156, v1 2023-03-17, v4 2025-01-17. \url{https://arxiv.org/abs/2303.11156}\par\smallskip

\noindent\hangindent=1.5em Shamardina et al., 2022. Shamardina, T., Mikhailov, V., Chernianskii, D., Fenogenova, A., Saidov, M., Valeeva, A., Shavrina, T., Smurov, I., Tutubalina, E., Artemova, E (2022). Findings of the The RuATD Shared Task 2022 on Artificial Text Detection in Russian. Dialogue-22; arXiv:2206.01583, submitted 2022-06-03. \url{https://arxiv.org/abs/2206.01583}\par\smallskip

\noindent\hangindent=1.5em Shamardina et al., 2024. Shamardina, T., Saidov, M., Fenogenova, A., Tumanov, A., Zemlyakova, A., Lebedeva, A., Gryaznova, E., Shavrina, T., Mikhailov, V., Artemova, E (2024). CoAT: Corpus of artificial texts. Natural Language Processing (Cambridge University Press), vol. 31, no. 1, January 2025, pp. 150-175; \foreignlanguage{russian}{опубликовано онлайн} 2024-09-06. \url{https://www.cambridge.org/core/journals/natural-language-processing/article/coat-corpus-of-artificial-texts/7E2CA97E21663CC031FB6BAFE56E0046}\par\smallskip

\noindent\hangindent=1.5em Stowe et al., 2026. Stowe, K., Afanaseva, S., Raimundo, R. C., Sun, Y., Patil, K (2026). Identifying Bias in Machine-generated Text Detection. ACL 2026. \url{https://aclanthology.org/2026.acl-long.109/}\par\smallskip

\noindent\hangindent=1.5em Su et al., 2025. Su, Z., Wang, Y., Wan, H., Zhang, Z., Luo, M (2025). HACo-Det: A Study Towards Fine-Grained Machine-Generated Text Detection under Human-AI Coauthoring. ACL 2025 (63rd Annual Meeting), Vienna. \url{https://aclanthology.org/2025.acl-long.1069/}\par\smallskip

\noindent\hangindent=1.5em Ter\v{c}on and Dobrovoljc, 2025. Ter\v{c}on, L., Dobrovoljc, K (2025). Linguistic Characteristics of AI-Generated Text: A Survey. arXiv:2510.05136, submitted 2025-10-01. \url{https://arxiv.org/abs/2510.05136}\par\smallskip

\noindent\hangindent=1.5em Tolstykh et al., 2025. Tolstykh, I., Tsybina, A., Yakubson, S., Kuprashevich, M (2025). LLMTrace: A Corpus for Classification and Fine-Grained Localization of AI-Written Text. arXiv:2509.21269, submitted 2025-09-25. \url{https://arxiv.org/abs/2509.21269}\par\smallskip

\noindent\hangindent=1.5em Tufts et al., 2025. Tufts, B., Zhao, X., Li, L (2025). A Practical Examination of AI-Generated Text Detectors for Large Language Models. Findings of the ACL: NAACL 2025. \url{https://aclanthology.org/2025.findings-naacl.271/}\par\smallskip

\noindent\hangindent=1.5em Tulchinskii et al., 2023. Tulchinskii, E., Kuznetsov, K., Kushnareva, L., Cherniavskii, D., Nikolenko, S., Burnaev, E., Barannikov, S., Piontkovskaya, I (2023). Intrinsic Dimension Estimation for Robust Detection of AI-Generated Texts. NeurIPS 2023, Advances in Neural Information Processing Systems 36.\par\smallskip

\noindent\hangindent=1.5em Venkatraman et al., 2024. Venkatraman, S., Uchendu, A., Lee, D (2024). GPT-who: An Information Density-based Machine-Generated Text Detector. Findings of the ACL: NAACL 2024, pp. 103–115. \url{https://aclanthology.org/2024.findings-naacl.8/}\par\smallskip

\noindent\hangindent=1.5em Vera-Jimenez et al., 2026. Vera-Jimenez, A., Jaeger, S., Ibenye, C., Ghosh, D (2026). Cross-Prompt Generalization in Detecting AI-Generated Fake News Using Interpretable Linguistic Features. arXiv:2606.04199, submitted 2026-06-02. \url{https://arxiv.org/abs/2606.04199}\par\smallskip

\noindent\hangindent=1.5em Verma et al., 2024. Verma, V., Fleisig, E., Tomlin, N., Klein, D (2024). Ghostbuster: Detecting Text Ghostwritten by Large Language Models. NAACL 2024, Volume 1: Long Papers. \url{https://aclanthology.org/2024.naacl-long.95/}\par\smallskip

\noindent\hangindent=1.5em Wang et al., 2024a. Wang, Y., Mansurov, J., Ivanov, P., Su, J., Shelmanov, A., Tsvigun, A., Whitehouse, C., Afzal, O. M., Mahmoud, T., Sasaki, T., Arnold, T., Aji, A. F., Habash, N., Gurevych, I., Nakov, P (2024). M4: Multi-generator, Multi-domain, and Multi-lingual Black-Box Machine-Generated Text Detection. EACL 2024 (18th Conference of the European Chapter of the ACL). \url{https://aclanthology.org/2024.eacl-long.83/}\par\smallskip

\noindent\hangindent=1.5em Wang et al., 2024b. Wang, Y., Feng, S., Hou, A. B., Pu, X., Shen, C., Liu, X., Tsvetkov, Y., He, T (2024). Stumbling Blocks: Stress Testing the Robustness of Machine-Generated Text Detectors Under Attacks. arXiv:2402.11638, submitted 2024-02-18. \url{https://arxiv.org/abs/2402.11638}\par\smallskip

\noindent\hangindent=1.5em Wang et al., 2024c. Wang, Y., Mansurov, J., Ivanov, P., Su, J., Shelmanov, A., Tsvigun, A., Afzal, O. M., Mahmoud, T., Puccetti, G., Arnold, T. et al. (2024). SemEval-2024 Task 8: Multidomain, Multimodel and Multilingual Machine-Generated Text Detection. Proceedings of the 18th International Workshop on Semantic Evaluation (SemEval-2024). \url{https://aclanthology.org/2024.semeval-1.279/}\par\smallskip

\noindent\hangindent=1.5em Wang et al., 2025. Wang, Y., Shelmanov, A., Mansurov, J., Tsvigun, A., Mikhailov, V., Xing, R., Xie, Z., Geng, J., Puccetti, G., Artemova, E. et al. (2025). GenAI Content Detection Task 1: English and Multilingual Machine-Generated Text Detection: AI vs. Human. 1st Workshop on GenAI Content Detection (GenAIDetect), Abu Dhabi, January 2025. \url{https://aclanthology.org/2025.genaidetect-1.27/}\par\smallskip

\noindent\hangindent=1.5em Zaitsu et al., 2025. Zaitsu, W., Jin, M., Ishihara, S., Tsuge, S., Inaba, M (2025). Stylometry can reveal artificial intelligence authorship, but humans struggle: A comparison of human and seven large language models in Japanese. PLOS One, 2025, e0335369. \url{https://journals.plos.org/plosone/article?id=10.1371/journal.pone.0335369}\par\smallskip

\noindent\hangindent=1.5em Zamaraeva et al., 2025. Zamaraeva, O., Flickinger, D., Bond, F., G\'omez-Rodr\'iguez, C (2025). Comparing LLM-generated and human-authored news text using formal syntactic theory. ACL 2025, main conference; arXiv:2506.01407, submitted 2025-06-02. \url{https://arxiv.org/abs/2506.01407}\par\smallskip

\noindent\hangindent=1.5em Zeng et al., 2024. Zeng, Z., Liu, S., Sha, L., Li, Z., Yang, K., Liu, S., Ga\v{s}evi\'c, D., Chen, G (2024). Detecting AI-Generated Sentences in Human-AI Collaborative Hybrid Texts: Challenges, Strategies, and Insights. IJCAI 2024, Special Track AI and Social Good; arXiv:2403.03506, v1 2024-03-06, v4 2024-05-23. \url{https://arxiv.org/abs/2403.03506}\par\smallskip

\noindent\hangindent=1.5em Zhu et al., 2024. Zhu, T., Weissburg, I., Zhang, K., Wang, W. Y (2024). Human Bias in the Face of AI: Examining Human Judgment Against Text Labeled as AI Generated. arXiv:2410.03723, v2. \url{https://arxiv.org/abs/2410.03723}\par\smallskip

\noindent\hangindent=1.5em Gritsay et al., 2023. \foreignlanguage{russian}{Грицай}, \foreignlanguage{russian}{Г}. \foreignlanguage{russian}{М}., \foreignlanguage{russian}{Грабовой}, \foreignlanguage{russian}{А}. \foreignlanguage{russian}{В}., \foreignlanguage{russian}{Кильдяков}, \foreignlanguage{russian}{А}. \foreignlanguage{russian}{С}., \foreignlanguage{russian}{Чехович}, \foreignlanguage{russian}{Ю}. \foreignlanguage{russian}{В} (2023). \foreignlanguage{russian}{Поиск искусственно сгенерированных текстовых фрагментов в научных документах}. \foreignlanguage{russian}{Доклады Российской академии наук}. \foreignlanguage{russian}{Математика}, \foreignlanguage{russian}{информатика}, \foreignlanguage{russian}{процессы управления}, 2023, \foreignlanguage{russian}{том} 514, № 2, \foreignlanguage{russian}{с}. 308-317. \url{https://sciencejournals.ru/view-article/?j=danmiup&y=2023&v=514&n=2&a=DANMIUp2360167Gritsay}\par\smallskip

\section{Appendix A. Reproducibility and data quality}

\subsection{1. The prep-v4 $\to$ prep-v5 transition}

The corpus shrank from 1916 documents to 1882. Excluded were 34 human documents left with fewer than 700 words after verbatim duplication was removed: they passed the collection threshold only through an extraction defect. The text changed in 68 documents; the remaining 1814 matched prep-v4 bitwise.

The rules are registered in \texttt{02-\allowbreak{}preregistration/amendment-\allowbreak{}prep-\allowbreak{}v5-\allowbreak{}data-\allowbreak{}quality.md}; the correction log is \texttt{04-\allowbreak{}corpus/prep-\allowbreak{}v5-\allowbreak{}corrections.csv}.

\subsection{2. Clearance to compute: six invariants}

\texttt{09-\allowbreak{}tools/preflight\_v2\_run.py} assembles a manifest with hashes of inputs, code, splits, and models, and sets \texttt{preflight\_passed} only when all six invariants pass. The outcome under revision 6 — Table S11:

\begin{center}
\small
\begin{tabularx}{\textwidth}{lXX}
\toprule
No. & Invariant & Result \\
\midrule
1 & the registry contains 1882 documents & passed \\
2 & none of the 34 excluded documents entered the matrix & passed \\
3 & raw and normalized values of the 1814 unchanged documents match v4 & passed, 114 282 values compared, zero discrepancies \\
4 & the percentile implementation reproduces v4 on the v4 composition & passed, 88 114 values \\
5 & the v5 percentiles are computed on the new reference sample & passed \\
6 & series v1 results are not overwritten & passed \\
\bottomrule
\end{tabularx}
\end{center}

The v2-series computation scripts call \texttt{require(stage, procedure)}: it re-reads the manifest and \textbf{re-verifies the hashes} of its procedure's inputs and code, so swapping a file after clearance has been granted stops the computation.

\subsection{3. Six manifest revisions}

A faulty manifest is not deleted: it is marked \texttt{invalidated} with a reason, and work continues under a new revision. On 29 July the mechanism fired five times (Table S12).

\begin{center}
\small
\begin{tabularx}{\textwidth}{lX}
\toprule
Revision & What forced the move to the next one \\
\midrule
1 & edit to \texttt{clf\_run.py} under the amendment on two inner CV schemes \\
2 & procedure 1 did not create a child manifest — the chain was broken \\
3 & the \texttt{pick\_c} call in P2b kept the old number of return values and stopped the run \\
4 & the fairness and error-analysis reports labelled themselves series v1 and referenced the v1 verification file although they were verified against v2 \\
5 & error analysis retrained the model with its own \texttt{GroupKFold} instead of the registered scheme \\
6 & in effect \\
\bottomrule
\end{tabularx}
\end{center}

Each revision invalidates the child manifests assembled under the previous parent and forces their procedures to be recomputed. The final numbers were obtained under revision 6 in a single chain.

A separate finding: before 29 July, \texttt{require} read a single file name, and the first \texttt{invalidated} mark would have blocked the series instead of moving to a new revision. The revision mechanism existed in the specification but had never been executed.

\subsection{4. The error-analysis defect}

\texttt{error\_run.fit\_for\_split} reproduces the model of the run under analysis in order to compute feature contributions for each card. Before the fix, it selected regularisation with its own \texttt{GroupKFold}. In series v1 this matched the run, because there the split really was built in place; in series v2 the run uses the registered scheme, and the cards would have explained the errors of a model that never existed in the run.

After the fix, reproduction was verified on three holdouts: 2750 scores compared, maximum discrepancy 5$\cdot$10$^{-7}$ — half of the last digit recorded in the CSV.

\subsection{5. The parse cache key}

The Stanza, NER, and embedding caches were addressed by \texttt{document\_id} alone. During the prep-v5 correction, the text of 68 documents changed while their identifiers stayed the same, and the cache would have returned a stale parse for them.

The key was switched to \texttt{document\_id $\times$ input\_sha256 $\times$ cache\_version $\times$ model\_revision}. The preprocessing version is recorded as provenance but does not enter the validity check: a byte-identical input yields a byte-identical parse, and requiring the version to match would have forced recomputation of 1814 documents with no change in the result. The premise was verified — for all 1916 documents, the \texttt{prose\_sha256} of the version under which the parse was built matched the \texttt{prose\_sha256} in prep-v4.

New entries do not overwrite old ones: the name contains the input hash, so the prep-v4 parse stays on disk and series v1 remains reproducible.

\subsection{6. The reuse gate}

\texttt{09-\allowbreak{}tools/feature\_reuse\_gate.py} builds the parse from scratch on a sentinel sample — three documents of each class from among the unchanged ones, selected by identifier sort order before the run. Verification results — Table S13.

\begin{center}
\small
\begin{tabularx}{\textwidth}{lX}
\toprule
Stage & Result \\
\midrule
Stanza & bitwise match on six documents \\
NER & bitwise match on six documents \\
embeddings & vector discrepancy up to 3.5$\cdot$10$^{-7}$, sem-v1 features up to 2.5$\cdot$10$^{-6}$, against a tolerance of 1$\cdot$10$^{-4}$ \\
\bottomrule
\end{tabularx}
\end{center}

Bitwise agreement never happens for embeddings: the cache was computed in batches across document boundaries, and the reduction order in the sum differs. The tolerance is taken from the NLL gate, not set to fit the observed number. The practical consequence is the opposite of the concern — cache reuse removes this source of noise.

Partial recomputation of procedures 3 and 4 was confirmed by the same means. NLL: 1814 documents inherited, 68 recomputed, the repeatability block fell entirely on the inherited ones, \texttt{max|$\Delta$| = 4.46$\cdot$10$^{-7}$}. Judge: the same 1814 and 68, nine sentinel calls on three documents, zero discrepancies.

\subsection{7. Index decomposition: a stability check on v1}

The decomposition of the contrast into style and structural components was computed on the main v2 series and reported in the main text. The same script, run on the frozen v1 outputs, reproduced the numbers of the 27 July addendum — which is the check that the formula did not change along with the series.

\begin{center}
\small
\begin{tabularx}{\textwidth}{llXllll}
\toprule
Contrast & Series & Common contribution & Format contribution & Sum & Frozen $\Delta$full & Discrepancy \\
\midrule
P3 $-$ P1 & v2 & +1.6740 & $-$3.4395 & $-$1.7655 & $-$1.7655 & 1.6$\cdot$10$^{-4}$ \\
P3 $-$ P1 & v1 & +1.5458 & $-$3.4577 & $-$1.9118 & $-$1.9118 & 1.5$\cdot$10$^{-4}$ \\
P2 $-$ P1 & v2 & $-$1.3927 & +4.6739 & +3.2812 & +3.2812 & 1.5$\cdot$10$^{-4}$ \\
P2 $-$ P1 & v1 & $-$1.4879 & +4.6800 & +3.1921 & +3.1921 & 1.5$\cdot$10$^{-4}$ \\
\bottomrule
\end{tabularx}
\end{center}

The format-sensitive score as a separate output on \texttt{P3 $-$ P1}: $-$19.78 in v2 against $-$19.88 in v1, 359 pairs in both series.

The discrepancy between the sum of contributions and the recorded contrast is rounding to four digits when writing to the CSV. The script stops if even one document has a category left without features: the denominator then differs and the decomposition stops being exact.

Input and code hashes — \texttt{07-\allowbreak{}analysis/score-\allowbreak{}v2-\allowbreak{}index-\allowbreak{}decomposition.md} and \texttt{score-\allowbreak{}v1-\allowbreak{}index-\allowbreak{}decomposition.md}; the source run — \texttt{manifests-\allowbreak{}v2/proc1-\allowbreak{}v2-\allowbreak{}manifest.json}.

\subsection{9. Explaining the effect of transformations 10 and 11 through volume — a withdrawn interpretation}

\textbf{Status: withdrawn interpretation.} The wording is given in full because for a week
it served as the basis for reading the result, and the entries of 31 July and
1 August referred to it. It has no force in Results or in the main conclusion.

The withdrawn wording: "the classifier is robust to cosmetic edits and collapses
under volume change; expanding the text flips the decision in nearly half of the
cases, shortening — in one and a half percent, with high probability instability".

What refutes it. Shortening reduces volume by 24\%, expansion increases it by
25\% — symmetric changes. Yet the decision flips number 4 of 464 versus 109 of 464,
a 27-fold difference. It is shortening that loses more paragraphs: a median of 83.5\%
against 75.6\%. Neither length nor loss of markup explains such asymmetry.

A single feature separates the pair. Median $\Delta$z over the 60 panel documents,
standardised on the 1882 documents of \texttt{feature-\allowbreak{}matrix-\allowbreak{}v5.csv} — Table S15:

\begin{center}
\small
\begin{tabularx}{\textwidth}{Xll}
\toprule
Feature & t10 shortening & t11 expansion \\
\midrule
M01 dispersion of similarity of adjacent sentences & +0.13 & +7.86 \\
S08 & $-$0.27 & +4.53 \\
L02 & +0.11 & $-$3.02 \\
L01 & +0.15 & $-$1.45 \\
R06 & $-$1.30 & $-$1.39 \\
R07 & $-$1.00 & $-$1.00 \\
P01 & $-$0.37 & $-$0.34 \\
\bottomrule
\end{tabularx}
\end{center}

The four format features shift identically under both — that is the contribution
of the lost markup, and it does not flip decisions. The transformations diverge
on M01.

What enters the model under the name M01 is not the mean similarity of adjacent
sentences but its standard deviation: lines 290–292 of \texttt{extract\_semantic.py} put
the mean into \texttt{raw\_value} and the dispersion into \texttt{normalized\_value}, while
\texttt{clf\_run.load\_matrix} reads \texttt{normalized\_value or raw\_value}. Duplicating every
fourth sentence places two identical sentences next to each other with a cosine
of 1.0 against typical values of $\sim$0.5, so the dispersion moves 7.86 $\sigma$ upward in
all sixty documents. Shortening creates no identical neighbours and shows no
such shift.

All 109 decision flips go towards the "human" class, not one in the opposite direction.

What holds instead of the withdrawn wording is limitation 25 of the main text.
The analysis with reproduction — \texttt{07-\allowbreak{}analysis/stress-\allowbreak{}t10-\allowbreak{}t11-\allowbreak{}volume-\allowbreak{}defect.md};
the PI decision — \texttt{00-\allowbreak{}admin/status-\allowbreak{}2026-\allowbreak{}08-\allowbreak{}01.md}.

\subsection{10. What remains open}

\begin{itemize}
\item \texttt{holdout\_source} under scheme B has two inner folds instead of three: two machine channels remain in the training part, and a third two-class fold is impossible under any arrangement. The reason was recorded before the metrics were computed;
\item for two documents — \texttt{human\_seo\_drmax\_0013} and \texttt{human\_seo\_madcats\_0005} — the duplication mechanism differs from the ones analysed, and no rule was devised for them after inspection.
\end{itemize}

\section{Appendix B. Figures}

\subsection{Six figures}

\begin{center}
\small
\begin{tabularx}{\textwidth}{llXX}
\toprule
No. & File & Data source & What it shows \\
\midrule
1 & \texttt{fig-\allowbreak{}01-\allowbreak{}design-\allowbreak{}and-\allowbreak{}corpus} & \texttt{documents-\allowbreak{}registry.csv} & how the corpus was assembled: two production branches, shared preprocessing, final composition \\
2 & \texttt{fig-\allowbreak{}02-\allowbreak{}o1-\allowbreak{}four-\allowbreak{}procedures} & \texttt{synthesis-\allowbreak{}o1-\allowbreak{}v2.csv} & four contrasts \texttt{P3 $-$ P1}, each on its own scale; after the convention the sign is minus for three procedures and plus for the judge \\
3 & \texttt{fig-\allowbreak{}03-\allowbreak{}classifier-\allowbreak{}heterogeneity} & \texttt{clf-\allowbreak{}v2-\allowbreak{}p2a-\allowbreak{}metrics.csv} & FPR across 18 group holdouts: median 0.0380 versus 0.310 on \texttt{genre\_seo} \\
4 & \texttt{fig-\allowbreak{}04-\allowbreak{}calibration-\allowbreak{}and-\allowbreak{}risk-\allowbreak{}coverage} & \texttt{calibration-\allowbreak{}v1-\allowbreak{}by-\allowbreak{}holdout.csv}, \texttt{calibration-\allowbreak{}v1-\allowbreak{}manifest.json} & calibration and trading coverage for risk; no threshold is selected \\
5 & \texttt{fig-\allowbreak{}05-\allowbreak{}stress-\allowbreak{}ten-\allowbreak{}transformations} & \texttt{stress-\allowbreak{}p2a-\allowbreak{}r11-\allowbreak{}cells.csv} & decision flips and instability across ten transformations \\
6 & \texttt{fig-\allowbreak{}06-\allowbreak{}t02-\allowbreak{}pipeline-\allowbreak{}mechanism} & \texttt{stress-\allowbreak{}p2a-\allowbreak{}r11-\allowbreak{}scores.csv}, the \texttt{prep-\allowbreak{}v5} and \texttt{stress-\allowbreak{}v3/t02} profiles & a case where the decision flips while the full profile is unchanged \\
\bottomrule
\end{tabularx}
\end{center}

\subsection{Captions}

Each caption reads without the main text. The numbers below are reproduced from the
same files the figure was built from.

\begin{figure}[p]
\centering
\includegraphics[width=\linewidth]{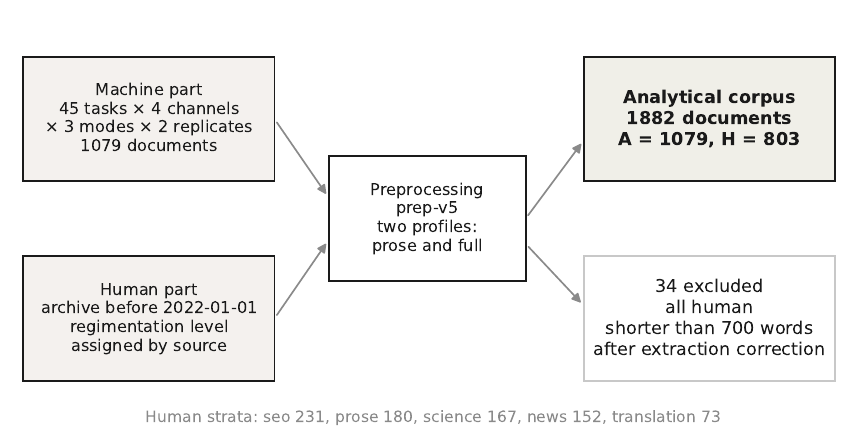}
\par\vspace{6pt}
{\small \textbf{Figure 1. Corpus design.} The machine branch: 45 tasks $\times$ 4 channels $\times$ 3 task modes $\times$
2 replicates, 1079 documents; the human branch: an archive published before 2022-01-01,
with annotation of the source regimentation level, 803 documents. Both branches pass
through the shared \texttt{prep-\allowbreak{}v5} preprocessing in two profiles, \texttt{prose} and \texttt{full}.
The analytical corpus is 1882 documents. 34 documents were excluded, all human,
shorter than 700 words after extraction correction. Human strata: seo 231, prose 180,
science 167, news 152, translation 73.\par}
\end{figure}

\begin{figure}[p]
\centering
\includegraphics[width=\linewidth]{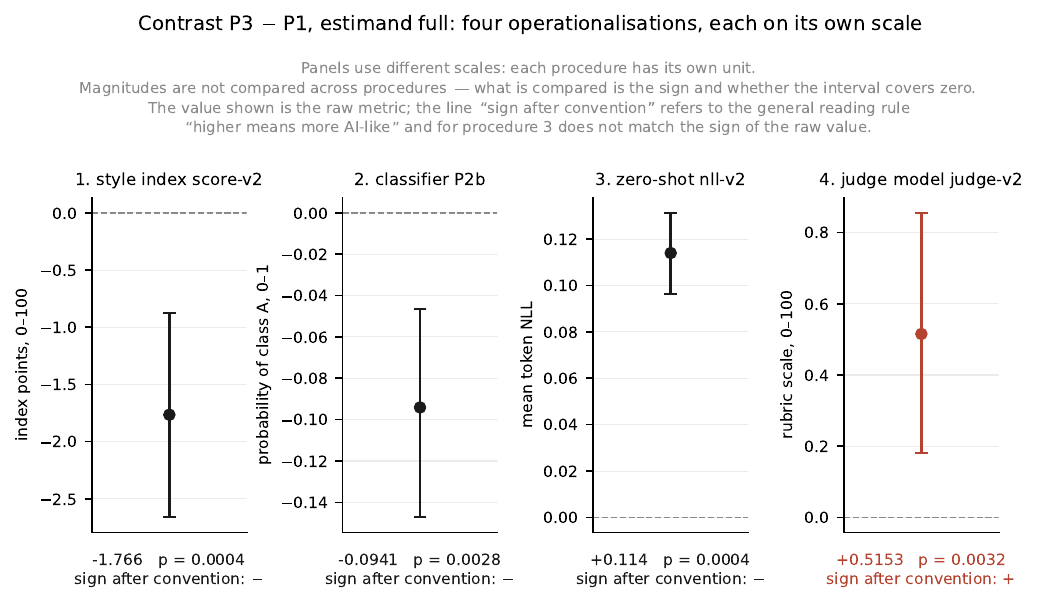}
\par\vspace{6pt}
{\small \textbf{Figure 2. Contrast \texttt{P3 $-$ P1}, estimand full, four operationalisations.} The dot is
the estimate, the segment the 95\% interval. The panels are drawn on different scales,
each procedure with its own unit; magnitudes are not compared across procedures — what
is compared is the sign and whether the interval covers zero. The value shown is the
raw metric, while the line "sign after convention" refers to the shared interpretation
rule "higher means more AI-like" and, for procedure 3, does not match the sign of the raw value. The raw values: style index $-$1.766 (p = 0.0004), classifier
$-$0.0941 (p = 0.0028), zero-shot NLL +0.114 (p = 0.0004), judge model +0.5153
(p = 0.0032). After conversion to the "higher means more AI-like" convention, the sign
of the first three procedures is negative and the judge's is positive; the diverging
procedure is marked by colour and a label.\par}
\end{figure}

\begin{figure}[p]
\centering
\includegraphics[width=\linewidth]{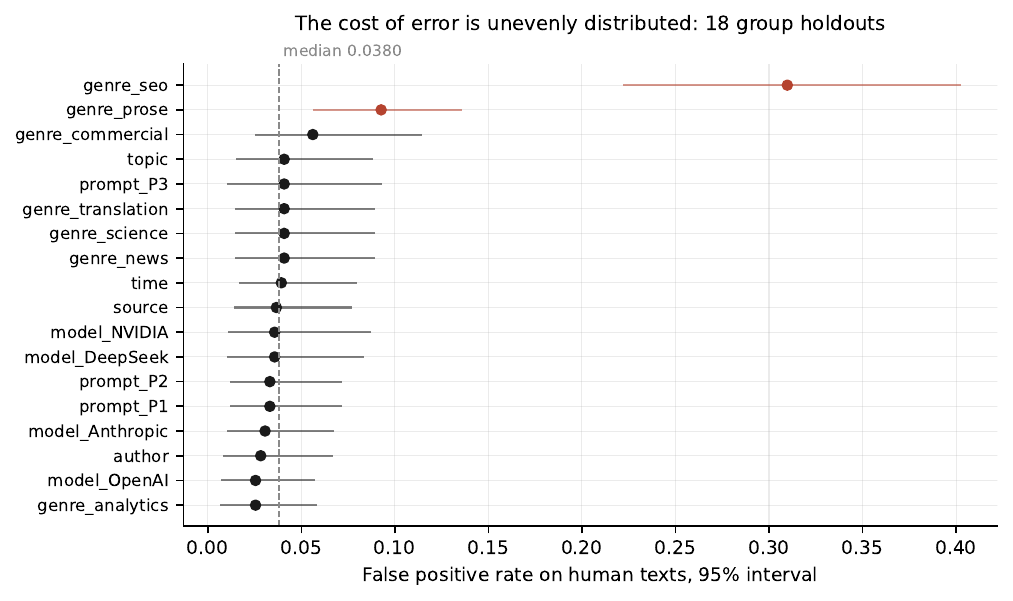}
\par\vspace{6pt}
{\small \textbf{Figure 3. False positive rate across 18 group holdouts.} The dot is the FPR of the
main model on estimand full, the segment the 95\% interval. Median 0.0380. The extremes:
\texttt{genre\_seo} 0.310 [0.223; 0.402] and \texttt{model\_OpenAI} 0.0255 [0.0076; 0.0566] —
a difference of more than an order of magnitude with the same model and the same
threshold. Colour marks the two strata where the cost of error is highest.\par}
\end{figure}

\begin{figure}[p]
\centering
\includegraphics[width=\linewidth]{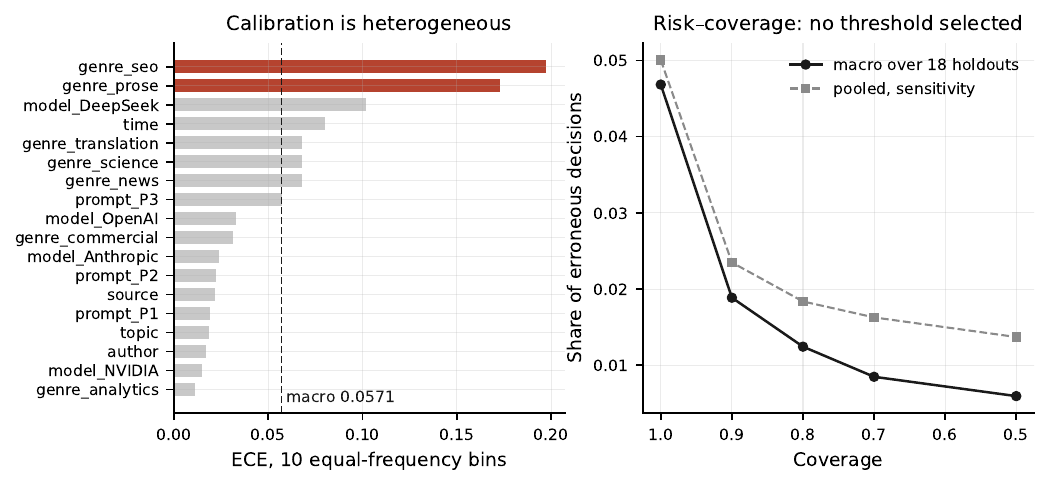}
\par\vspace{6pt}
{\small \textbf{Figure 4. Calibration and trading coverage for risk.} Left, reliability across 18
holdouts; right, the share of erroneous decisions as coverage shrinks. Macro-average
Brier 0.0365 with a spread from 0.0109 to 0.1796; the worst holdout is \texttt{genre\_seo}.
Macro-average ECE over ten equal-frequency bins 0.0571, maximum 0.1973. Risk
falls from 0.0468 at full coverage to 0.0059 at 50\% coverage. The ECE, MCE and
risk–coverage values are descriptive: the metric name was registered, the
operationalisation was refined post hoc before computation. No threshold is selected
from this figure.\par}
\end{figure}

\begin{figure}[p]
\centering
\includegraphics[width=\linewidth]{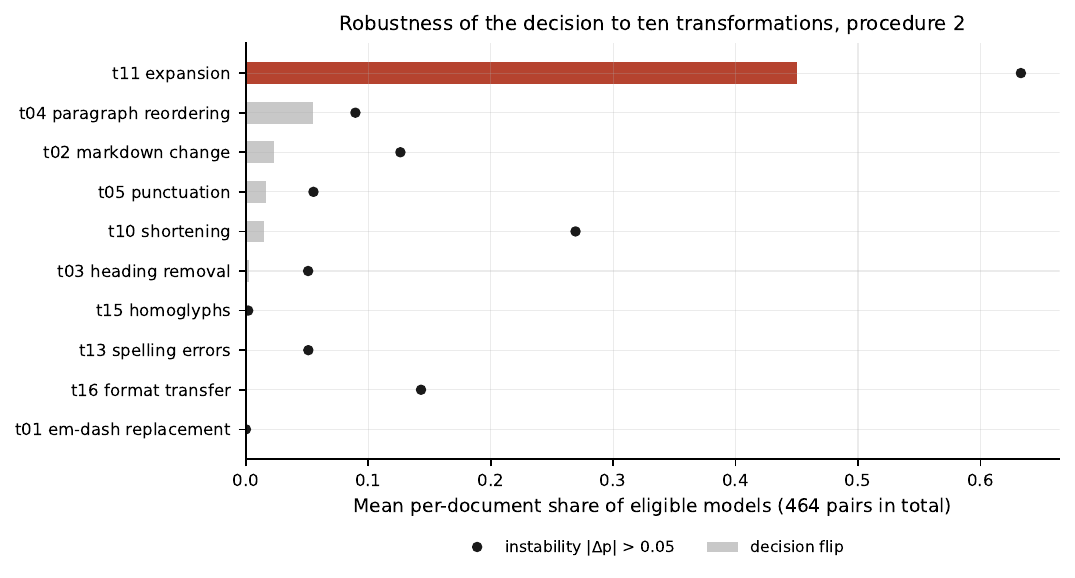}
\par\vspace{6pt}
{\small \textbf{Figure 5. Robustness of the decision to ten transformations, procedure 2.} The bar
is the mean over the 60 documents of the per-document share of eligible models
that flipped their decision; the
dot is the same aggregation for instability |$\Delta$p| > 0.05. The aggregation rule is
registered: the mean over models within a document, then equal weight for the
documents; 464 document $\times$ eligible-model pairs in total, 600 cells = 60 documents $\times$
10 transformations. The extremes: t11 (expansion) gives a mean decision-flip share
of 45.1\% (109 pairs of 464 in total) and instability of 63.3\%; t01 (dash
replacement) and t16 (formats) flip no decisions, yet t16 shifts the probability
for 14.3\% of a document's models on average.\par}
\end{figure}

\begin{figure}[p]
\centering
\includegraphics[width=\linewidth]{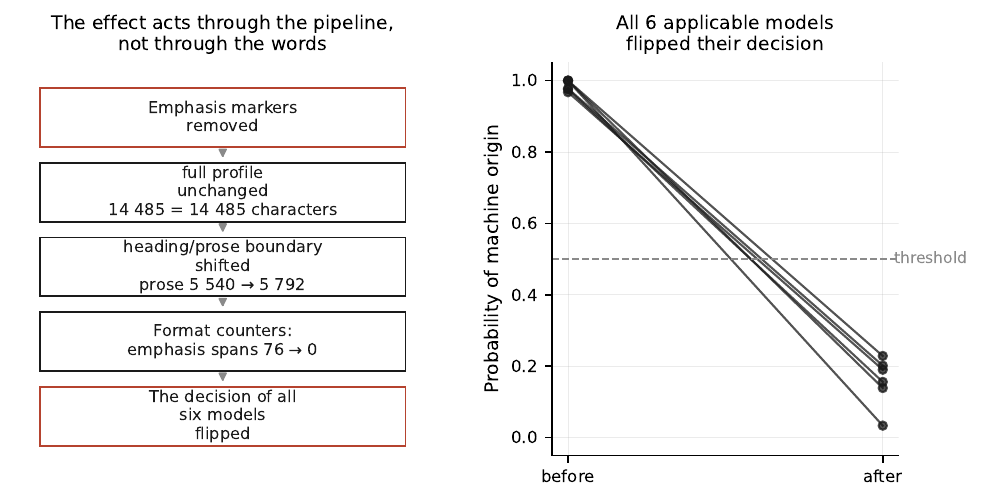}
\par\vspace{6pt}
{\small \textbf{Figure 6. The t02 mechanism: the effect goes through the pipeline, not through the words.}
Document \texttt{nemotron\_b016\_P1\_r1}. Removing the emphasis markers did not change the full
profile — 14 485 characters before and after — but shifted the heading/prose boundary:
the prose profile grew from 5 540 to 5 792 characters, the emphasis counter fell from
76 to 0. The decision of all six applicable models flipped: the probability of class A
fell from the range 0.968–1.000 to 0.034–0.229 at a threshold of 0.5.\par}
\end{figure}

\subsection{Presentation rules observed in construction}

\begin{itemize}
\item a figure is readable without the caption in the text: title, axes and units are in place;
\item confidence intervals are shown wherever there is an estimate;
\item the scales of the four procedures are not brought onto a single axis — they are incommensurable by construction (§4.5 of the registration);
\item colour is not the only carrier of distinction: the highlighted cases are labelled in words;
\item each figure is saved as PNG and PDF; inputs and outputs are fixed by hashes.
\end{itemize}

\subsection{What was dropped compared with the plan of nine}

\begin{center}
\small
\begin{tabularx}{\textwidth}{lX}
\toprule
Dropped & Reason \\
\midrule
full versus net for procedures 1 and 2 & the material remains in the table; a separate figure would duplicate it \\
decomposition into common and format & the same; the numbers are given in Results §3 \\
structural response by channel and genre & twelve cells read better as a table than as a plot \\
resolving power of the judge & moved to the Supplementary together with the full distributions \\
format-only versus the full model & folded into the table of metrics by holdout \\
examples of errors and disagreements & text fragments, not suitable for a figure \\
\bottomrule
\end{tabularx}
\end{center}

The calibration plot, removed in the previous revision as an unclosed item, returned
as Figure 4: calibration was computed on 2026-08-01 under the amendment
\texttt{02-\allowbreak{}preregistration/amendment-\allowbreak{}calibration-\allowbreak{}spec.md}.

\clearpage
\renewcommand{\thetable}{S\arabic{table}}
\setcounter{table}{0}
\section{Supplementary Material}

\subsection{S1. Corpus composition}

\input{tables/tab-S01-corpus-flow}

Data flow from 1916 documents to 1882 and the distribution of exclusions across
strata. Referenced from Methods §1.0 and Results §1.

\input{tables/tab-S02-machine-channels}

Four generation channels: models, access wrappers, weight openness, sampling
parameters. Referenced from Methods §1.1.

\input{tables/tab-S03-preprocessing-verdicts}

Extraction correction verdicts for the 113 candidates. Referenced from Methods §3.

\subsection{S4–S6. Classifier and calibration for each holdout}

\input{tables/tab-S04-fpr-by-holdout}

False positive rate of the main model on each of the 18 group holdouts, with
intervals and the breakdown by hard-human subgroups. Full version of Figure 3.

\input{tables/tab-S05-classifier-quality-by-holdout}

AUROC, balanced accuracy, MCC and TPR at FPR 1\%, row by row, with the selected
regularisation parameter. Five holdouts yield a single-class test, and AUROC is
undefined there.

\input{tables/tab-S06-calibration-by-holdout}

Brier, ECE and MCE on ten bins, sensitivity to five and twenty bins, risk at
full and half coverage. Full version of Figure 4.

\subsection{S7. Stress test row by row}

\input{tables/tab-S07-stress-by-transform-and-procedure}

Ten transformations $\times$ four procedures: median and maximum of the absolute shift,
share of instability events, each procedure's own event rule. Full version of
Figure 5.

\subsection{S8. Mixed-effects diagnostics}

\input{tables/tab-S08-mixed-identifiability}

Design matrix rank, merged levels, cluster sizes and the verdict for each of the
three preregistered models. Referenced from Methods §6.

\subsection{S9–S10. Manual analyses}

\input{tables/tab-S09-repeat-audit}

Seventeen documents that passed the threshold of the symmetric repeat scan, with
mechanism and verdict. Referenced from Results §1.

\input{tables/tab-S10-blind-instability-cards}

Thirty cards of the blind instability analysis: presence of the expected change,
unexpected changes, type and verdict. Referenced from Results §8.

\subsection{S11–S13. Gates and revisions}

\input{tables/tab-S11-preflight-invariants}

Six invariants for admission to the v2 series computation. Referenced from
Appendix A §2.

\input{tables/tab-S12-manifest-revisions}

Six revisions of the admission manifest and the reason for moving to the next
one. Referenced from Appendix A §3.

\input{tables/tab-S13-reuse-gate}

Reconciliation of parse reuse on the sentinel sample. Referenced from
Appendix A §6.

\subsection{S14–S16. Checks and the composition of the limitations}

\input{tables/tab-S14-index-decomposition}

Decomposition of the style index into stylistic and structural parts in the v2
and v1 series. Referenced from Results §3.1 and Appendix A §7.

\input{tables/tab-S15-t10-t11-feature-shift}

Feature shift under shortening and expansion of the text — the interpretation
via volume, which has been ruled out. Referenced from Appendix A §9.

\input{tables/tab-S16-limitations-groups}

Twenty-six limitation items collapsed into the seven groups of the main text.
Referenced from the Limitations section.

\end{document}

%% file: tables/tab-01-corpus-and-conditions.tex
\begin{table}[htbp]
\centering
\small
\caption{Corpus composition and text production conditions}
\label{tab:1}
\begin{tabular}{llll}
\toprule
Corpus part & Production condition & Category & Documents \\
\midrule
machine & generation channel & real\_claude & 270 \\
machine & generation channel & deepseek\_pro & 270 \\
machine & generation channel & gpt & 270 \\
machine & generation channel & nemotron & 269 \\
machine & task mode & P1 & 360 \\
machine & task mode & P2 & 360 \\
machine & task mode & P3 & 359 \\
machine & replicate & 1 & 540 \\
machine & replicate & 2 & 539 \\
human & level 2 & Template-based materials, seo & 231 \\
human & level 0 & Free-form authorial texts, prose & 180 \\
human & level 3 & Standardised prose, science & 167 \\
human & level 1 & Journalism, news & 152 \\
human & no level assigned & Language-contact texts, translation & 73 \\
both & total & analytical corpus & 1882 \\
\bottomrule
\end{tabular}
\begin{minipage}{\linewidth}\footnotesize\vspace{2pt}
\textit{Note.} The unit of analysis is the document. Denominator: 1882 documents of the analytical corpus, 1079 machine and 803 human. Data: corpus registry, preprocessing profile prep-v5. The table reports document counts; it contains no estimates or intervals. Scale: none; the table describes composition. Abbreviations: P1, P2 and P3 are task modes, P1 — a neutral technical brief, P2 — a strict editorial brief, P3 — a task requiring the avoidance of typical machine-text features; level is the source regimentation level, assigned from the source before reading the text. Missing values: the translation stratum receives no level: for translations the regimentation is set by the source text, not by the editorial process. No correction for multiple comparisons was applied.
\end{minipage}
\end{table}

%% file: tables/tab-08-mixed-effects-status.tex
\begin{table*}[htbp]
\centering
\small
\caption{Three preregistered mixed-effects models: identifiability and status}
\label{tab:8}
\begin{tabular}{>{\RaggedRight\hspace{0pt}\arraybackslash}p{0.120\linewidth}>{\RaggedRight\hspace{0pt}\arraybackslash}p{0.120\linewidth}>{\RaggedRight\hspace{0pt}\arraybackslash}p{0.120\linewidth}>{\RaggedRight\hspace{0pt}\arraybackslash}p{0.120\linewidth}>{\RaggedRight\hspace{0pt}\arraybackslash}p{0.120\linewidth}>{\RaggedRight\hspace{0pt}\arraybackslash}p{0.120\linewidth}>{\RaggedRight\hspace{0pt}\arraybackslash}p{0.120\linewidth}}
\toprule
Model & Corpus & Primary outcome & Observations & Design matrix rank & Aliased levels & Status \\
\midrule
M1 & machine & O1 & 1079 & 22 of 23 & 1 & not identifiable; fitting was run and did not converge \\
M2 & human & O2 & 803 & 7 of 22 & 15 & not identifiable \\
M3 & combined & O3 & 1882 & 11 of 11 & 0 & identifiable \\
\bottomrule
\end{tabular}
\begin{minipage}{\linewidth}\footnotesize\vspace{2pt}
\textit{Note.} Unit of analysis: the observation is a document. Denominator: 1079 machine, 803 human and 1882 combined documents. Data: series v2, identifiability diagnostics of 2026-07-29. Design matrix rank against the number of candidate columns; aliased levels are factors indistinguishable by design. The table reports no effect estimates: no model produced the primary outcome. Scale: none; the table describes the status of the procedure. Abbreviations: REML — restricted maximum likelihood; aliased level — a factor level linearly dependent on the remaining design columns. Missing entries: the estimates column is omitted deliberately: M1 did not converge, M2 is not identifiable by design, and M3 has no support for the registered estimand, so there are no numbers that could be published. No multiple-comparison correction was applied.
\end{minipage}
\end{table*}

%% file: tables/tab-02-four-operationalisations.tex
\begin{table*}[htbp]
\centering
\small
\caption{Four operationalisations of machine-likeness and the direction of their scales}
\label{tab:2}
\begin{tabular}{>{\RaggedRight\hspace{0pt}\arraybackslash}p{0.120\linewidth}>{\RaggedRight\hspace{0pt}\arraybackslash}p{0.120\linewidth}>{\RaggedRight\hspace{0pt}\arraybackslash}p{0.120\linewidth}>{\RaggedRight\hspace{0pt}\arraybackslash}p{0.120\linewidth}>{\RaggedRight\hspace{0pt}\arraybackslash}p{0.120\linewidth}>{\RaggedRight\hspace{0pt}\arraybackslash}p{0.120\linewidth}>{\RaggedRight\hspace{0pt}\arraybackslash}p{0.120\linewidth}}
\toprule
Procedure & Input & Origin of weights & Scale & Direction under the convention & Registration status & Instability event rule \\
\midrule
1 — style index & feature matrix & weights fixed before data collection & 0–100 points & higher values mean higher machine-likeness & confirmatory & $|$Δ$|$ $>$ 5.0 points; exactly 5.0 does not count as an event \\
2 — classifier & the same feature matrix & weights learned from the data & probability of class A, 0–1 & higher values mean higher machine-likeness & exploratory & $|$Δp$|$ $>$ 0.05 and a decision flip relative to the threshold \\
3 — zero-shot NLL & raw text, prose profile & no weights, a model metric & mean token NLL & lower values mean higher machine-likeness & exploratory & no binary threshold is applied; the continuous ΔNLL is published \\
4 — judge model & raw text, prose profile & rubric fixed before the run & 0–100 rubric scale & higher values mean higher machine-likeness & exploratory & $|$Δ$|$ $>$ 5.0 points on the median of three seeds \\
\bottomrule
\end{tabular}
\begin{minipage}{\linewidth}\footnotesize\vspace{2pt}
\textit{Note.} The unit of analysis is the scoring procedure. Denominator: four procedures applied to the same corpus. Data: series v2. The table describes procedures; it contains no numerical estimates. Scale: interpretation convention: higher means more AI-like. For procedure 3 it inverts the sign of the raw metric, so magnitudes are not comparable across procedures. Abbreviations: NLL — negative log-likelihood, averaged over tokens; confirmatory — a preregistered analysis; exploratory — an analysis whose operationalisation was written after viewing the results of procedure 1. Missing values: none. No correction for multiple comparisons was applied.
\end{minipage}
\end{table*}

%% file: tables/tab-03-o1-contrasts.tex
\begin{sidewaystable}[p]
\centering
\scriptsize
\caption{Primary outcome O1 contrasts across the four procedures, both estimands and both contrasts}
\label{tab:3}
\begin{tabular}{lllllllll}
\toprule
Contrast & Estimand & Procedure & Contrast population & Pairs & Effect, raw scale & 95\% CI & p & Sign under the convention \\
\midrule
P3 − P1 & full & 1 — style index score-v2 & whole corpus, 45 tasks & 359 & −1.7655 & [−2.6639; −0.8773] & 0.0004 & − \\
P3 − P1 & full & 2 — classifier P2b & seo genre only, 15 tasks & 120 & −0.0941 & [−0.1473; −0.0463] & 0.0028 & − \\
P3 − P1 & full & 3 — zero-shot nll-v2 & whole corpus, 45 tasks & 359 & +0.1140 & [0.0964; 0.1313] & 0.0004 & − \\
P3 − P1 & full & 4 — judge model judge-v2 & whole corpus, 45 tasks & 359 & +0.5153 & [0.1806; 0.8552] & 0.0032 & + \\
P3 − P1 & net & 1 — style index score-v2 & whole corpus, 45 tasks & 359 & +2.0264 & [0.8953; 3.1429] & 0.0016 & + \\
P3 − P1 & net & 2 — classifier P2b & seo genre only, 15 tasks & 120 & +0.0169 & [−0.0134; 0.0496] & 0.3168 & + \\
P3 − P1 & net & 3 — zero-shot nll-v2 & whole corpus, 45 tasks & — & not defined & — & — & — \\
P3 − P1 & net & 4 — judge model judge-v2 & whole corpus, 45 tasks & — & not defined & — & — & — \\
P2 − P1 & full & 1 — style index score-v2 & whole corpus, 45 tasks & 360 & +3.2812 & [2.0186; 4.6068] & 0.0004 & + \\
P2 − P1 & full & 2 — classifier P2b & seo genre only, 15 tasks & 120 & +0.0523 & [0.0325; 0.0738] & 0.0002 & + \\
P2 − P1 & full & 3 — zero-shot nll-v2 & whole corpus, 45 tasks & 360 & +0.0125 & [−0.0065; 0.0298] & 0.1988 & − \\
P2 − P1 & full & 4 — judge model judge-v2 & whole corpus, 45 tasks & 360 & +0.5125 & [0.0181; 1.0472] & 0.0392 & + \\
P2 − P1 & net & 1 — style index score-v2 & whole corpus, 45 tasks & 360 & −1.6860 & [−3.2913; −0.0343] & 0.0416 & − \\
P2 − P1 & net & 2 — classifier P2b & seo genre only, 15 tasks & 120 & +0.0722 & [0.0395; 0.1036] & 0.0014 & + \\
P2 − P1 & net & 3 — zero-shot nll-v2 & whole corpus, 45 tasks & — & not defined & — & — & — \\
P2 − P1 & net & 4 — judge model judge-v2 & whole corpus, 45 tasks & — & not defined & — & — & — \\
\bottomrule
\end{tabular}
\begin{minipage}{\linewidth}\footnotesize\vspace{2pt}
\textit{Note.} The unit of analysis is a pair of texts produced by the same model for the same task in two modes. Denominator: 359 or 360 pairs for procedures 1, 3 and 4 with 45 clusters; 120 pairs for procedure 2 with 15 clusters, since it computes the contrast on the seo genre only. Data: series v2, status current. Intervals are 95\%, cluster bootstrap over tasks; the cluster is the task, 45 tasks in total. The absolute difference is published with its own interval; there is no single number across the four procedures, and no pooled effect was computed. Scale: interpretation convention: higher means more AI-like. For procedure 3 it inverts the sign of the raw metric, so magnitudes are not comparable across procedures. Abbreviations: estimand full — the full score; net — the score without structural components; P1, P2 and P3 are task modes. Missing values: the net estimand is defined only for procedures 1 and 2; zero-shot NLL and the judge have no decomposition into components by construction, and the cell means ``quantity not defined'', not ``not computed''. The primary outcome family is closed by a Bonferroni correction for the two registered tests, α = 0.025 per test; the intervals are unadjusted.
\end{minipage}
\end{sidewaystable}

%% file: tables/tab-04-classifier-and-ablation.tex
\begin{table*}[htbp]
\centering
\scriptsize
\caption{Classifier: performance across model variants and ablation of structural features}
\label{tab:4}
\begin{tabular}{>{\RaggedRight\hspace{0pt}\arraybackslash}p{0.102\linewidth}>{\RaggedRight\hspace{0pt}\arraybackslash}p{0.102\linewidth}>{\RaggedRight\hspace{0pt}\arraybackslash}p{0.102\linewidth}>{\RaggedRight\hspace{0pt}\arraybackslash}p{0.102\linewidth}>{\RaggedRight\hspace{0pt}\arraybackslash}p{0.102\linewidth}>{\RaggedRight\hspace{0pt}\arraybackslash}p{0.102\linewidth}>{\RaggedRight\hspace{0pt}\arraybackslash}p{0.102\linewidth}>{\RaggedRight\hspace{0pt}\arraybackslash}p{0.102\linewidth}}
\toprule
Model variant & Feature set & Holdouts with defined AUROC & Median AUROC & AUROC range & Median FPR & P3 − P1 contrast, full matrix & P3 − P1 contrast, without structural features \\
\midrule
source-only & source platform identifier only & 13 of 18 & 1.0000 & 0.5000–1.0000 & 1.0000 & — & — \\
main+M02 & plus the paragraph similarity feature & 13 of 18 & 0.9979 & 0.9606–1.0000 & 0.0332 & — & — \\
main & 22 style features & 13 of 18 & 0.9973 & 0.9433–1.0000 & 0.0380 & −0.0941 [−0.1473; −0.0463], p = 0.0028 & +0.0169 [−0.0134; 0.0496], p = 0.3168 \\
format-only & four structural features & 13 of 18 & 0.9775 & 0.9154–1.0000 & 0.0118 & — & — \\
genre-only & genre only & 13 of 18 & 0.6667 & 0.5000–0.7259 & 1.0000 & — & — \\
length-only & text length only & 13 of 18 & 0.4364 & 0.2957–0.6138 & 0.6276 & — & — \\
negative-control & shuffled class labels & 13 of 13 & 0.3614 & 0.1279–0.4440 & — & — & — \\
\bottomrule
\end{tabular}
\begin{minipage}{\linewidth}\footnotesize\vspace{2pt}
\textit{Note.} The unit of analysis for the performance metrics is the group holdout; for the contrasts, a pair of texts produced by the same model for the same task. Denominator: 18 group holdouts; the contrast is computed on 120 pairs with 15 clusters, on the seo genre only. Data: series v2, status current. Median and range across holdouts; the contrast is the absolute difference in probabilities with a 95\% interval and a wild cluster bootstrap p. Intervals are 95\%, cluster bootstrap over author and document; the cluster is the task, 45 tasks in total. Scale: AUROC: higher means better separation; FPR: lower means fewer false accusations of a human; the contrast in class-A probability: higher means more AI-like. Abbreviations: AUROC — area under the ROC curve; FPR — false positive rate on human texts; ablation — a comparison of the model with and without the features, not a decomposition of the score into summands. Missing values: AUROC is not defined on five single-class holdouts, so the column states on how many it was computed; the contrast was computed for the main model only, the other variants show a dash; for the negative-control model FPR was not computed and there are thirteen holdouts: it was run through a series of cluster permutations and holds post hoc status. For the source-only model an AUROC of 1.0000 coexists with an FPR of 1.0000: ranking by platform is error-free, but at the 0.5 threshold the model declares the entire human test set of an unseen platform machine-generated. The primary outcome family is closed by a Bonferroni correction for the two registered tests, α = 0.025 per test; the intervals are unadjusted.
\end{minipage}
\end{table*}

%% file: tables/tab-05-formal-register-fairness.tex
\begin{table*}[htbp]
\centering
\small
\caption{Cost of error across subgroups of the human corpus: formal register, translation and genres}
\label{tab:5}
\begin{tabular}{>{\RaggedRight\hspace{0pt}\arraybackslash}p{0.120\linewidth}>{\RaggedRight\hspace{0pt}\arraybackslash}p{0.120\linewidth}>{\RaggedRight\hspace{0pt}\arraybackslash}p{0.120\linewidth}>{\RaggedRight\hspace{0pt}\arraybackslash}p{0.120\linewidth}>{\RaggedRight\hspace{0pt}\arraybackslash}p{0.120\linewidth}>{\RaggedRight\hspace{0pt}\arraybackslash}p{0.120\linewidth}>{\RaggedRight\hspace{0pt}\arraybackslash}p{0.120\linewidth}}
\toprule
Subgroup & Hard-human flag & Holdouts with the subgroup & Median FPR & FPR range & Median difference from the rest of the test set & Worst holdout \\
\midrule
entire human test set & — & 18 & 0.0380 & 0.0255–0.3098 & — & genre\_seo \\
genre=\allowbreak{}science & HH-formal-register & 18 & 0.0659 & 0.0419–0.5210 & 0.0429 & genre\_seo \\
genre=\allowbreak{}translation & HH-translation & 18 & 0.0342 & 0.0137–0.1233 & −0.0087 & genre\_seo \\
genre=\allowbreak{}news & HH-polished & 18 & 0.0132 & 0.0000–0.0658 & −0.0380 & genre\_prose \\
genre=\allowbreak{}seo & — & 4 & 0.0261 & 0.0000–0.3766 & −0.0095 & genre\_seo \\
genre=\allowbreak{}prose & — & 2 & 0.0639 & 0.0000–0.1278 & 0.0089 & genre\_prose \\
\bottomrule
\end{tabular}
\begin{minipage}{\linewidth}\footnotesize\vspace{2pt}
\textit{Note.} The unit of analysis is a human document within a group holdout. Denominator: 18 group holdouts; in each, the subgroup is computed on the human part of its test set. Data: series v2, status current. Median and range of FPR across holdouts where the subgroup is present; the difference is the subgroup FPR minus the FPR of the rest of the human test set in the same holdout. Intervals are 95\%, cluster bootstrap over author and document; the cluster is the task, 45 tasks in total. Scale: lower means fewer false accusations of a human. Abbreviations: FPR — false positive rate; HH — hard-human, flags applied on top of the corpus: formal-register — formal scientific register, translation — translated prose, polished — editorially polished text. Missing values: each hard-human flag coincides with a genre one to one: formal-register — all 167 science documents, polished — all 152 news documents, translation — all 73 translation documents. Rows are therefore not duplicated, and the flag is shown as a column; this design cannot separate the effect of the flag from the effect of the genre. A dash in the difference column means the subgroup coincided with the entire human test set. No correction for multiple comparisons was applied. The subgroup analysis is descriptive.
\end{minipage}
\end{table*}

%% file: tables/tab-06-calibration-and-risk-coverage.tex
\begin{table}[htbp]
\centering
\small
\caption{Probability calibration and the coverage–risk trade-off}
\label{tab:6}
\begin{tabular}{l>{\hspace{0pt}}p{2.4cm}>{\hspace{0pt}}p{2.4cm}>{\hspace{0pt}}p{1.9cm}>{\hspace{0pt}}p{2.6cm}}
\toprule
Quantity & Macro-average over 18 holdouts & Range across holdouts & Worst holdout & Quantity status \\
\midrule
Brier & 0.0365 & 0.0109–0.1796 & genre\_seo & registered metric \\
ECE, 10 bins & 0.0571 & 0.0113–0.1973 & genre\_seo & descriptive \\
MCE, 10 bins & 0.2647 & 0.0523–0.7548 & genre\_seo & descriptive \\
Risk at 100\% coverage & 0.0468 & 0.0173–0.2014 & genre\_seo & descriptive \\
Risk at 90\% coverage & 0.0189 & 0.0000–0.1774 & genre\_seo & descriptive \\
Risk at 80\% coverage & 0.0124 & 0.0000–0.1425 & genre\_seo & descriptive \\
Risk at 70\% coverage & 0.0085 & 0.0000–0.1076 & genre\_seo & descriptive \\
Risk at 50\% coverage & 0.0059 & 0.0000–0.0854 & genre\_seo & descriptive \\
\bottomrule
\end{tabular}
\begin{minipage}{\linewidth}\footnotesize\vspace{2pt}
\textit{Note.} The unit of analysis is a document within a group holdout. Denominator: 18 group holdouts; the macro-average is taken over holdouts, not over documents, so large holdouts do not outweigh small ones. Data: series calibration-v1, computed read-only from the fairness-v2 predictions. Brier is the mean squared deviation of the probability from the outcome; ECE and MCE are the mean and maximum gap between probability and frequency on equal-frequency bins; risk is the share of erroneous decisions among those retained as coverage is reduced. No intervals were computed. Scale: lower is better for all quantities in the table. Abbreviations: ECE — expected calibration error; MCE — maximum calibration error; coverage — the share of documents on which a decision is made, the rest go to abstention. Missing values: none; no threshold is selected from this table. No correction for multiple comparisons was applied.
\end{minipage}
\end{table}

%% file: tables/tab-07-stress-ten-transformations.tex
\begin{table*}[htbp]
\centering
\small
\caption{Stability of the decision under ten text transformations across four procedures}
\label{tab:7}
\begin{tabular}{>{\RaggedRight\hspace{0pt}\arraybackslash}p{0.120\linewidth}>{\RaggedRight\hspace{0pt}\arraybackslash}p{0.120\linewidth}>{\RaggedRight\hspace{0pt}\arraybackslash}p{0.120\linewidth}>{\RaggedRight\hspace{0pt}\arraybackslash}p{0.120\linewidth}>{\RaggedRight\hspace{0pt}\arraybackslash}p{0.120\linewidth}>{\RaggedRight\hspace{0pt}\arraybackslash}p{0.120\linewidth}>{\RaggedRight\hspace{0pt}\arraybackslash}p{0.120\linewidth}}
\toprule
Transformation & Feature input unchanged & Procedure 2: instability & Procedure 2: decision flip & Procedure 1: share of $|$Δ$|$ $>$ 5 points & Procedure 3: median $|$ΔNLL$|$ & Procedure 4: share of $|$Δ$|$ $>$ 5 points \\
\midrule
t11 text expansion & 0.000 & 0.633 & 0.451 & 0.567 & 0.4366 & 0.233 \\
t04 paragraph reordering & 0.250 & 0.089 & 0.055 & 0.267 & 0.0279 & 0.133 \\
t02 markdown markup change & 0.533 & 0.126 & 0.023 & 0.067 & 0.0000 & 0.033 \\
t05 punctuation correction & 0.017 & 0.055 & 0.017 & 0.100 & 0.0066 & 0.050 \\
t10 text shortening & 0.000 & 0.269 & 0.015 & 0.567 & 0.1584 & 0.117 \\
t03 heading removal & 0.600 & 0.051 & 0.003 & 0.000 & 0.0000 & 0.000 \\
t13 spelling errors & 0.000 & 0.051 & 0.001 & 0.050 & 0.1207 & 0.000 \\
t15 homoglyphs & 0.000 & 0.002 & 0.001 & 0.000 & 0.0216 & 0.033 \\
t01 em-dash replacement & 0.083 & 0.000 & 0.000 & 0.000 & 0.0019 & 0.033 \\
t16 transfer between formats & 0.450 & 0.143 & 0.000 & 0.217 & 0.0000 & 0.050 \\
\bottomrule
\end{tabular}
\begin{minipage}{\linewidth}\footnotesize\vspace{2pt}
\textit{Note.} Unit of analysis: for procedure 2, the document × eligible-model pair; for procedures 1, 3 and 4, the document. Denominator: 60 panel documents; for procedure 2 the shares follow the registered aggregation rule — the mean over eligible models within a document, then equal weight for the 60 documents (464 pairs per transformation in total); for the remaining procedures, 60 documents per transformation. Data: procedure 2 — stress revision r5, procedure 4 — r5, procedures 1 and 3 — r4 within the validity domain of the ten transformations. Shares and medians by cell; no intervals were computed. The instability event is defined by each procedure's own rule, so the columns are not additive across procedures. Scale: lower means a more stable decision. Abbreviations: Δ — shift of the score after the transformation; NLL — negative log-likelihood; decision flip — crossing the 0.5 threshold; ``feature input unchanged'' — the share of documents whose prose profile, full profile and all markup counters matched after the transformation, that is, the procedure received the same input as before. Missing entries: transformation t14 is excluded: the revision r5 amendment moved it to not executable status, so the denominator is ten rather than eleven. No multiple-comparison correction was applied. The stress test is descriptive.
\end{minipage}
\end{table*}

%% file: tables/tab-S01-corpus-flow.tex
\begin{table}[htbp]
\centering
\small
\caption{Data flow and exclusions by stratum}
\label{tab:s1}
\begin{tabular}{>{\RaggedRight\hspace{0pt}\arraybackslash}p{0.222\linewidth}>{\RaggedRight\hspace{0pt}\arraybackslash}p{0.222\linewidth}>{\RaggedRight\hspace{0pt}\arraybackslash}p{0.222\linewidth}>{\RaggedRight\hspace{0pt}\arraybackslash}p{0.222\linewidth}}
\toprule
Slice & Before & After & Change \\
\midrule
machine part & 1079 & 1079 & unchanged \\
human part & 837 & 803 & −34 \\
full corpus & 1916 & 1882 & −34 \\
stratum: Journalism, news & 180 & 152 & −28 \\
stratum: Templated materials, seo & 237 & 231 & −6 \\
stratum: Free authorial texts, prose & 180 & 180 & unchanged \\
stratum: Standardised prose, science & 167 & 167 & unchanged \\
stratum: Language-contact texts, translation & 73 & 73 & unchanged \\
\bottomrule
\end{tabular}
\begin{minipage}{\linewidth}\footnotesize\vspace{2pt}
\textit{Note.} Unit of analysis: the document. Denominator: 1916 documents before extraction correction, 1882 after. Data: corpus registry, correction correction-v5.0 of 2026-07-29. Document counts; no estimates or intervals. Scale: none. Abbreviations: prep-v5 — the active preprocessing profile; each human-part stratum is named together with its genre code. Missing entries: 34 documents excluded, all human: after correction of the extraction defect their text fell below the registered threshold of 700 words. No multiple-comparison correction was applied.
\end{minipage}
\end{table}

%% file: tables/tab-S02-machine-channels.tex
\begin{table}[htbp]
\centering
\small
\caption{Generation channels of the machine part: models, access and parameters}
\label{tab:s2}
\begin{tabular}{ll>{\hspace{0pt}}p{3.3cm}>{\hspace{0pt}}p{1.4cm}>{\hspace{0pt}}p{2.4cm}}
\toprule
Channel & Model & Wrapper and access & Open weights & Temperature and top-p \\
\midrule
real\_claude & claude-opus-4-8 & nested CLI, clean process per cell & no & channel default \\
gpt & gpt-5.5 & codex exec, reasoning effort high & no & channel default \\
deepseek\_pro & deepseek-v4-pro & HTTPS API without a wrapper & yes & 1.0 / 1.0 \\
nemotron & nemotron-3-ultra-free & HTTPS API via a gateway & yes & 1.0 / 1.0 \\
\bottomrule
\end{tabular}
\begin{minipage}{\linewidth}\footnotesize\vspace{2pt}
\textit{Note.} Unit of analysis: the generation channel. Denominator: four channels, 1079 machine documents; each channel produced 270 documents, except nemotron with 269. Data: corpus registry. The table describes generation conditions; it contains no run numbers. Scale: none. Abbreviations: top-p — the share of probability mass in sampling; ``channel default'' means the wrapper does not disclose the parameters. Missing entries: temperature and top-p are unavailable for two channels: the wrapper does not report them, so between-channel differences cannot be attributed to the model. No multiple-comparison correction was applied.
\end{minipage}
\end{table}

%% file: tables/tab-S03-preprocessing-verdicts.tex
\begin{table}[htbp]
\centering
\small
\caption{Extraction correction verdicts for the 113 candidates}
\label{tab:s3}
\begin{tabular}{>{\RaggedRight\hspace{0pt}\arraybackslash}p{0.302\linewidth}>{\RaggedRight\hspace{0pt}\arraybackslash}p{0.302\linewidth}>{\RaggedRight\hspace{0pt}\arraybackslash}p{0.302\linewidth}}
\toprule
Verdict & Documents & Action \\
\midrule
extraction-defect & 75 & re-extraction with the fixed parser \\
intermediary-defect & 28 & deterministic removal of exact cascades \\
unresolved & 9 & not edited; origin of the repetition not established \\
source-property & 1 & not edited; the repetition is present in the source \\
\bottomrule
\end{tabular}
\begin{minipage}{\linewidth}\footnotesize\vspace{2pt}
\textit{Note.} Unit of analysis: the document. Denominator: 113 candidates with a verbatim sentence duplication share above 0.1, found in a review of 1916 documents; 103 were corrected. Data: preprocessing prep-v5, correction correction-v5.0. Document counts by verdict. Scale: none. Abbreviations: extraction-defect — a defect in text extraction; intermediary-defect — a defect in an intermediary file; unresolved — the origin of the repetition is not established; source-property — the repetition is present in the source. Missing entries: ten of the 113 documents were left unedited: nine unresolved and one source-property. No multiple-comparison correction was applied.
\end{minipage}
\end{table}

%% file: tables/tab-S04-fpr-by-holdout.tex
\begin{sidewaystable}[p]
\centering
\scriptsize
\caption{False positive rate of the main model for each of the 18 group holdouts}
\label{tab:s4}
\begin{tabular}{>{\RaggedRight\hspace{0pt}\arraybackslash}p{0.102\linewidth}>{\RaggedRight\hspace{0pt}\arraybackslash}p{0.102\linewidth}>{\RaggedRight\hspace{0pt}\arraybackslash}p{0.102\linewidth}>{\RaggedRight\hspace{0pt}\arraybackslash}p{0.102\linewidth}>{\RaggedRight\hspace{0pt}\arraybackslash}p{0.102\linewidth}>{\RaggedRight\hspace{0pt}\arraybackslash}p{0.102\linewidth}>{\RaggedRight\hspace{0pt}\arraybackslash}p{0.102\linewidth}>{\RaggedRight\hspace{0pt}\arraybackslash}p{0.102\linewidth}}
\toprule
Holdout & Test documents & Human & FPR & 95\% CI & Formal-register FPR & Translation FPR & Cleaned-text FPR \\
\midrule
genre\_seo & 983 & 623 & 0.3098 & [0.2226; 0.4023] & 0.5210 & 0.1233 & 0.0658 \\
genre\_prose & 572 & 572 & 0.0927 & [0.0565; 0.1351] & 0.0838 & 0.0822 & 0.0658 \\
genre\_commercial & 752 & 392 & 0.0561 & [0.0260; 0.1141] & 0.0958 & 0.0411 & 0.0197 \\
genre\_news & 392 & 392 & 0.0408 & [0.0153; 0.0890] & 0.0659 & 0.0411 & 0.0132 \\
genre\_science & 392 & 392 & 0.0408 & [0.0153; 0.0890] & 0.0659 & 0.0411 & 0.0132 \\
genre\_translation & 392 & 392 & 0.0408 & [0.0153; 0.0890] & 0.0659 & 0.0411 & 0.0132 \\
prompt\_P3 & 751 & 392 & 0.0408 & [0.0108; 0.0925] & 0.0838 & 0.0274 & 0.0000 \\
topic & 751 & 392 & 0.0408 & [0.0158; 0.0877] & 0.0778 & 0.0137 & 0.0132 \\
time & 458 & 458 & 0.0393 & [0.0170; 0.0792] & 0.0778 & 0.0274 & 0.0132 \\
source & 1003 & 464 & 0.0366 & [0.0148; 0.0767] & 0.0719 & 0.0411 & 0.0132 \\
model\_DeepSeek & 662 & 392 & 0.0357 & [0.0108; 0.0830] & 0.0599 & 0.0274 & 0.0132 \\
model\_NVIDIA & 661 & 392 & 0.0357 & [0.0113; 0.0865] & 0.0599 & 0.0411 & 0.0066 \\
prompt\_P1 & 752 & 392 & 0.0332 & [0.0126; 0.0714] & 0.0539 & 0.0274 & 0.0132 \\
prompt\_P2 & 752 & 392 & 0.0332 & [0.0126; 0.0714] & 0.0539 & 0.0274 & 0.0132 \\
model\_Anthropic & 662 & 392 & 0.0306 & [0.0106; 0.0671] & 0.0479 & 0.0274 & 0.0132 \\
author & 764 & 495 & 0.0283 & [0.0086; 0.0663] & 0.0539 & 0.0411 & 0.0066 \\
genre\_analytics & 751 & 392 & 0.0255 & [0.0068; 0.0576] & 0.0419 & 0.0137 & 0.0132 \\
model\_OpenAI & 662 & 392 & 0.0255 & [0.0076; 0.0566] & 0.0419 & 0.0137 & 0.0132 \\
\bottomrule
\end{tabular}
\begin{minipage}{\linewidth}\footnotesize\vspace{2pt}
\textit{Note.} Unit of analysis: the human document within a holdout. Denominator: 18 group holdouts; each has its own human test part. Data: series v2, status current. Intervals are 95\%, cluster bootstrap over authors and documents; the cluster is the task, with 45 tasks. Scale: lower means fewer false accusations of a human. Abbreviations: FPR — false positive rate; HH subgroups — flags overlaid on the corpus, each document enters the overall calculation once. Missing entries: a dash means the subgroup is not represented in that holdout. Three holdouts — genre\_news, genre\_science and genre\_translation — give identical numbers: their test parts coincide in composition, which is a design limitation, not a typo. No multiple-comparison correction was applied.
\end{minipage}
\end{sidewaystable}

%% file: tables/tab-S05-classifier-quality-by-holdout.tex
\begin{sidewaystable}[p]
\centering
\scriptsize
\caption{Quality of the main model on each of the 18 group holdouts}
\label{tab:s5}
\begin{tabular}{>{\RaggedRight\hspace{0pt}\arraybackslash}p{0.078\linewidth}>{\RaggedRight\hspace{0pt}\arraybackslash}p{0.078\linewidth}>{\RaggedRight\hspace{0pt}\arraybackslash}p{0.078\linewidth}>{\RaggedRight\hspace{0pt}\arraybackslash}p{0.078\linewidth}>{\RaggedRight\hspace{0pt}\arraybackslash}p{0.078\linewidth}>{\RaggedRight\hspace{0pt}\arraybackslash}p{0.078\linewidth}>{\RaggedRight\hspace{0pt}\arraybackslash}p{0.078\linewidth}>{\RaggedRight\hspace{0pt}\arraybackslash}p{0.078\linewidth}>{\RaggedRight\hspace{0pt}\arraybackslash}p{0.078\linewidth}>{\RaggedRight\hspace{0pt}\arraybackslash}p{0.078\linewidth}}
\toprule
Holdout & Documents & Machine & Human & AUROC & Balanced accuracy & MCC & TPR at 1\% FPR & Selected C & Features \\
\midrule
author & 764 & 269 & 495 & 0.9910 & 0.9524 & 0.9079 & 0.8848 & 0.1 & 22 \\
genre\_analytics & 751 & 359 & 392 & 0.9987 & 0.9817 & 0.9628 & 0.9861 & 1.0 & 22 \\
genre\_commercial & 752 & 360 & 392 & 0.9974 & 0.9636 & 0.9264 & 0.9750 & 1.0 & 22 \\
genre\_news & 392 & 0 & 392 & not defined &  &  & — & 0.1 & 22 \\
genre\_prose & 572 & 0 & 572 & not defined &  &  & — & 0.01 & 22 \\
genre\_science & 392 & 0 & 392 & not defined &  &  & — & 0.1 & 22 \\
genre\_seo & 983 & 360 & 623 & 0.9433 & 0.8382 & 0.6560 & 0.0306 & 10.0 & 22 \\
genre\_translation & 392 & 0 & 392 & not defined &  &  & — & 0.1 & 22 \\
model\_Anthropic & 662 & 270 & 392 & 1.0000 & 0.9847 & 0.9634 & 1.0000 & 10.0 & 22 \\
model\_DeepSeek & 662 & 270 & 392 & 0.9958 & 0.9655 & 0.9284 & 0.9222 & 0.01 & 22 \\
model\_NVIDIA & 661 & 269 & 392 & 0.9902 & 0.9580 & 0.9154 & 0.8848 & 0.1 & 22 \\
model\_OpenAI & 662 & 270 & 392 & 0.9973 & 0.9761 & 0.9502 & 0.9519 & 0.1 & 22 \\
prompt\_P1 & 752 & 360 & 392 & 0.9994 & 0.9820 & 0.9632 & 0.9861 & 10.0 & 22 \\
prompt\_P2 & 752 & 360 & 392 & 1.0000 & 0.9834 & 0.9660 & 1.0000 & 10.0 & 22 \\
prompt\_P3 & 751 & 359 & 392 & 0.9777 & 0.9225 & 0.8494 & 0.6518 & 10.0 & 22 \\
source & 1003 & 539 & 464 & 0.9950 & 0.9668 & 0.9338 & 0.9351 & 0.1 & 22 \\
time & 458 & 0 & 458 & not defined &  &  & — & 0.1 & 22 \\
topic & 751 & 359 & 392 & 0.9983 & 0.9740 & 0.9472 & 0.9694 & 1.0 & 22 \\
\bottomrule
\end{tabular}
\begin{minipage}{\linewidth}\footnotesize\vspace{2pt}
\textit{Note.} Unit of analysis: group holdout. Denominator: 18 holdouts; test composition is set by the grouping variable. Data: series v2, status current. Metrics are computed on the test set of each holdout; C is selected by nested cross-validation on the training part. Scale: AUROC, balanced accuracy, MCC and TPR: higher is better. Abbreviations: MCC — Matthews correlation coefficient; TPR — proportion of machine texts correctly identified; C — regularisation parameter. Missing values: AUROC is not defined on five single-class holdouts: the test set lacks one of the two classes, so there is no ranking to compare. No correction for multiple comparisons was applied.
\end{minipage}
\end{sidewaystable}

%% file: tables/tab-S06-calibration-by-holdout.tex
\begin{sidewaystable}[p]
\centering
\scriptsize
\caption{Probability calibration on each of the 18 group holdouts}
\label{tab:s6}
\begin{tabular}{lllllllllll}
\toprule
Holdout & Documents & Machine & Brier & ECE-10 & MCE-10 & Non-empty bins & ECE-5 & ECE-20 & Risk at 100\% coverage & Risk at 50\% coverage \\
\midrule
genre\_seo & 983 & 360 & 0.1796 & 0.1973 & 0.7548 & 10 & 0.1973 & 0.1973 & 0.2014 & 0.0854 \\
genre\_prose & 572 & 0 & 0.0674 & 0.1727 & 0.6253 & 10 & 0.1727 & 0.1727 & 0.0927 & 0.0035 \\
prompt\_P3 & 751 & 359 & 0.0645 & 0.0572 & 0.1973 & 9 & 0.0572 & 0.0573 & 0.0759 & 0.0106 \\
model\_DeepSeek & 662 & 270 & 0.0382 & 0.1021 & 0.2046 & 10 & 0.1021 & 0.1021 & 0.0347 & 0.0000 \\
model\_NVIDIA & 661 & 269 & 0.0335 & 0.0148 & 0.0757 & 10 & 0.0148 & 0.0196 & 0.0408 & 0.0030 \\
author & 764 & 269 & 0.0305 & 0.0171 & 0.0523 & 10 & 0.0063 & 0.0200 & 0.0419 & 0.0026 \\
time & 458 & 0 & 0.0289 & 0.0803 & 0.4668 & 10 & 0.0803 & 0.0803 & 0.0393 & 0.0000 \\
genre\_commercial & 752 & 360 & 0.0258 & 0.0313 & 0.2040 & 10 & 0.0313 & 0.0338 & 0.0372 & 0.0000 \\
genre\_news & 392 & 0 & 0.0254 & 0.0682 & 0.4415 & 10 & 0.0682 & 0.0682 & 0.0408 & 0.0000 \\
genre\_science & 392 & 0 & 0.0254 & 0.0682 & 0.4415 & 10 & 0.0682 & 0.0682 & 0.0408 & 0.0000 \\
genre\_translation & 392 & 0 & 0.0254 & 0.0682 & 0.4415 & 10 & 0.0682 & 0.0682 & 0.0408 & 0.0000 \\
source & 1003 & 539 & 0.0249 & 0.0216 & 0.0666 & 10 & 0.0071 & 0.0216 & 0.0329 & 0.0020 \\
model\_OpenAI & 662 & 270 & 0.0209 & 0.0332 & 0.1280 & 10 & 0.0332 & 0.0332 & 0.0242 & 0.0000 \\
topic & 751 & 359 & 0.0172 & 0.0187 & 0.1035 & 10 & 0.0185 & 0.0227 & 0.0266 & 0.0000 \\
prompt\_P1 & 752 & 360 & 0.0134 & 0.0189 & 0.1223 & 9 & 0.0189 & 0.0191 & 0.0186 & 0.0000 \\
model\_Anthropic & 662 & 270 & 0.0127 & 0.0242 & 0.2337 & 9 & 0.0242 & 0.0242 & 0.0181 & 0.0000 \\
prompt\_P2 & 752 & 360 & 0.0125 & 0.0225 & 0.1390 & 8 & 0.0225 & 0.0226 & 0.0173 & 0.0000 \\
genre\_analytics & 751 & 359 & 0.0109 & 0.0113 & 0.0664 & 10 & 0.0111 & 0.0156 & 0.0186 & 0.0000 \\
\bottomrule
\end{tabular}
\begin{minipage}{\linewidth}\footnotesize\vspace{2pt}
\textit{Note.} Unit of analysis: document within a group holdout. Denominator: 18 group holdouts. Data: series calibration-v1, read-only over fairness-v2 predictions. Bins are equal-frequency; ties in probability are assigned to bins by the mid-rank rule. ECE-5 and ECE-20 give the sensitivity to the number of bins. Intervals were not computed. Scale: lower is better. Abbreviations: ECE — expected calibration error; MCE — maximum calibration error; the number after the hyphen is the bin count. Missing values: the non-empty bins column shows how many of the ten bins received at least one document: on small holdouts some bins are empty, and ECE is computed over the filled ones. No correction for multiple comparisons was applied.
\end{minipage}
\end{sidewaystable}

%% file: tables/tab-S07-stress-by-transform-and-procedure.tex
\begingroup\scriptsize
\begin{longtable}{>{\RaggedRight\hspace{0pt}\arraybackslash}p{0.102\linewidth}>{\RaggedRight\hspace{0pt}\arraybackslash}p{0.102\linewidth}>{\RaggedRight\hspace{0pt}\arraybackslash}p{0.102\linewidth}>{\RaggedRight\hspace{0pt}\arraybackslash}p{0.102\linewidth}>{\RaggedRight\hspace{0pt}\arraybackslash}p{0.102\linewidth}>{\RaggedRight\hspace{0pt}\arraybackslash}p{0.102\linewidth}>{\RaggedRight\hspace{0pt}\arraybackslash}p{0.102\linewidth}>{\RaggedRight\hspace{0pt}\arraybackslash}p{0.102\linewidth}}
\caption{Stress test row by row: ten transformations × four procedures}\label{tab:s7} \\
\toprule
Transformation & Procedure & Cells & Unit & Median $|$Δ$|$ & Maximum $|$Δ$|$ & Instability event rate & Event rule \\
\midrule
\endfirsthead
\toprule
Transformation & Procedure & Cells & Unit & Median $|$Δ$|$ & Maximum $|$Δ$|$ & Instability event rate & Event rule \\
\midrule
\endhead
\bottomrule
\endfoot
t01 em-dash replacement & 1 — style index & 60 & document & 0.000 & 4.348 & 0.000 & $|$Δ$|$ $>$ 5.0 points \\
t01 em-dash replacement & 2 — classifier & 464 & document × model pair & 0.000 & 0.029 & 0.000 & $|$Δp$|$ $>$ 0.05 and decision flip \\
t01 em-dash replacement & 3 — zero-shot NLL & 60 & document & 0.0019 & 0.0095 & — & threshold not applied \\
t01 em-dash replacement & 4 — judge model & 60 & document & 0.000 & 20.000 & 0.033 & $|$Δ$|$ $>$ 5.0 points on the median of three seeds \\
t02 markdown markup change & 1 — style index & 60 & document & 0.000 & 9.179 & 0.067 & $|$Δ$|$ $>$ 5.0 points \\
t02 markdown markup change & 2 — classifier & 464 & document × model pair & 0.000 & 0.966 & 0.023 & $|$Δp$|$ $>$ 0.05 and decision flip \\
t02 markdown markup change & 3 — zero-shot NLL & 60 & document & 0.0000 & 0.0491 & — & threshold not applied \\
t02 markdown markup change & 4 — judge model & 60 & document & 0.000 & 10.000 & 0.033 & $|$Δ$|$ $>$ 5.0 points on the median of three seeds \\
t03 heading removal & 1 — style index & 60 & document & 0.000 & 4.348 & 0.000 & $|$Δ$|$ $>$ 5.0 points \\
t03 heading removal & 2 — classifier & 464 & document × model pair & 0.000 & 0.493 & 0.003 & $|$Δp$|$ $>$ 0.05 and decision flip \\
t03 heading removal & 3 — zero-shot NLL & 60 & document & 0.0000 & 0.0000 & — & threshold not applied \\
t03 heading removal & 4 — judge model & 60 & document & 0.000 & 0.000 & 0.000 & $|$Δ$|$ $>$ 5.0 points on the median of three seeds \\
t04 paragraph reordering & 1 — style index & 60 & document & 1.449 & 17.391 & 0.267 & $|$Δ$|$ $>$ 5.0 points \\
t04 paragraph reordering & 2 — classifier & 464 & document × model pair & 0.000 & 0.825 & 0.055 & $|$Δp$|$ $>$ 0.05 and decision flip \\
t04 paragraph reordering & 3 — zero-shot NLL & 60 & document & 0.0279 & 0.3430 & — & threshold not applied \\
t04 paragraph reordering & 4 — judge model & 60 & document & 0.000 & 43.000 & 0.133 & $|$Δ$|$ $>$ 5.0 points on the median of three seeds \\
t05 punctuation correction & 1 — style index & 60 & document & 0.000 & 15.217 & 0.100 & $|$Δ$|$ $>$ 5.0 points \\
t05 punctuation correction & 2 — classifier & 464 & document × model pair & 0.000 & 0.320 & 0.017 & $|$Δp$|$ $>$ 0.05 and decision flip \\
t05 punctuation correction & 3 — zero-shot NLL & 60 & document & 0.0066 & 0.0944 & — & threshold not applied \\
t05 punctuation correction & 4 — judge model & 60 & document & 0.000 & 7.000 & 0.050 & $|$Δ$|$ $>$ 5.0 points on the median of three seeds \\
t10 text shortening & 1 — style index & 60 & document & 5.556 & 21.618 & 0.567 & $|$Δ$|$ $>$ 5.0 points \\
t10 text shortening & 2 — classifier & 464 & document × model pair & 0.005 & 0.825 & 0.015 & $|$Δp$|$ $>$ 0.05 and decision flip \\
t10 text shortening & 3 — zero-shot NLL & 60 & document & 0.1584 & 0.5387 & — & threshold not applied \\
t10 text shortening & 4 — judge model & 60 & document & 0.000 & 10.000 & 0.117 & $|$Δ$|$ $>$ 5.0 points on the median of three seeds \\
t11 text expansion & 1 — style index & 60 & document & 5.616 & 24.275 & 0.567 & $|$Δ$|$ $>$ 5.0 points \\
t11 text expansion & 2 — classifier & 464 & document × model pair & 0.025 & 1.000 & 0.451 & $|$Δp$|$ $>$ 0.05 and decision flip \\
t11 text expansion & 3 — zero-shot NLL & 60 & document & 0.4366 & 0.5900 & — & threshold not applied \\
t11 text expansion & 4 — judge model & 60 & document & 3.000 & 10.000 & 0.233 & $|$Δ$|$ $>$ 5.0 points on the median of three seeds \\
t13 spelling errors & 1 — style index & 60 & document & 0.000 & 8.696 & 0.050 & $|$Δ$|$ $>$ 5.0 points \\
t13 spelling errors & 2 — classifier & 464 & document × model pair & 0.000 & 0.339 & 0.001 & $|$Δp$|$ $>$ 0.05 and decision flip \\
t13 spelling errors & 3 — zero-shot NLL & 60 & document & 0.1207 & 0.1748 & — & threshold not applied \\
t13 spelling errors & 4 — judge model & 60 & document & 0.000 & 4.000 & 0.000 & $|$Δ$|$ $>$ 5.0 points on the median of three seeds \\
t15 homoglyphs & 1 — style index & 60 & document & 0.000 & 4.348 & 0.000 & $|$Δ$|$ $>$ 5.0 points \\
t15 homoglyphs & 2 — classifier & 464 & document × model pair & 0.000 & 0.333 & 0.001 & $|$Δp$|$ $>$ 0.05 and decision flip \\
t15 homoglyphs & 3 — zero-shot NLL & 60 & document & 0.0216 & 0.0604 & — & threshold not applied \\
t15 homoglyphs & 4 — judge model & 60 & document & 0.000 & 10.000 & 0.033 & $|$Δ$|$ $>$ 5.0 points on the median of three seeds \\
t16 transfer between formats & 1 — style index & 60 & document & 0.000 & 15.700 & 0.217 & $|$Δ$|$ $>$ 5.0 points \\
t16 transfer between formats & 2 — classifier & 464 & document × model pair & 0.000 & 0.461 & 0.000 & $|$Δp$|$ $>$ 0.05 and decision flip \\
t16 transfer between formats & 3 — zero-shot NLL & 60 & document & 0.0000 & 0.1482 & — & threshold not applied \\
t16 transfer between formats & 4 — judge model & 60 & document & 0.000 & 13.000 & 0.050 & $|$Δ$|$ $>$ 5.0 points on the median of three seeds \\
\end{longtable}
\begin{minipage}{\linewidth}\footnotesize\vspace{2pt}
\textit{Note.} Unit of analysis: given in its own column: document for procedures 1, 3 and 4, document × eligible-model pair for procedure 2. Denominator: 60 panel documents; for procedure 2 the shares are the mean over eligible models within a document, then equal weight for the 60 documents (464 pairs per transformation in total). Data: procedures 2 and 4 — revision r5, procedures 1 and 3 — r4 within the validity domain of the ten transformations. Medians and maxima of the absolute shift over the cells of each procedure: for procedures 1, 3 and 4 — over 60 documents, for procedure 2 — over 464 pairs. The event rate is computed by each procedure's own rule. Intervals were not computed. Scale: lower means a more stable decision. Abbreviations: Δ — score shift after the transformation; Δp — shift in the probability of class A. Missing values: for procedure 3 the event rate is not defined: a binary threshold is not applied to continuous NLL; transformation t14 is excluded as not executable under the revision r5 amendment. No correction for multiple comparisons was applied.
\end{minipage}
\endgroup

%% file: tables/tab-S08-mixed-identifiability.tex
\begin{sidewaystable}[p]
\centering
\scriptsize
\caption{Identifiability diagnostics for the three mixed-effects models}
\label{tab:s8}
\begin{tabular}{>{\RaggedRight\hspace{0pt}\arraybackslash}p{0.078\linewidth}>{\RaggedRight\hspace{0pt}\arraybackslash}p{0.078\linewidth}>{\RaggedRight\hspace{0pt}\arraybackslash}p{0.078\linewidth}>{\RaggedRight\hspace{0pt}\arraybackslash}p{0.078\linewidth}>{\RaggedRight\hspace{0pt}\arraybackslash}p{0.078\linewidth}>{\RaggedRight\hspace{0pt}\arraybackslash}p{0.078\linewidth}>{\RaggedRight\hspace{0pt}\arraybackslash}p{0.078\linewidth}>{\RaggedRight\hspace{0pt}\arraybackslash}p{0.078\linewidth}>{\RaggedRight\hspace{0pt}\arraybackslash}p{0.078\linewidth}>{\RaggedRight\hspace{0pt}\arraybackslash}p{0.078\linewidth}}
\toprule
Model & Observations & Design rank & Random effect & Clusters & Minimum size & Median size & Single-observation clusters & Aliased levels & Verdict \\
\midrule
M1 & 1079 & 22 of 23 & task (brief\_id) & 45 & 23 & 24 & 0 & wrapper=none & not identifiable \\
M1 &  &  & replicate (repeat\_index) & 2 & 539 & 540 & 0 &  &  \\
M2 & 803 & 7 of 22 & source & 124 & 1 & 1 & 85 & genre=\allowbreak{}science; genre=\allowbreak{}seo; genre=\allowbreak{}translation; level=1 × genre=\allowbreak{}prose; level=1 × genre=\allowbreak{}science; level=1 × genre=\allowbreak{}seo; level=1 × genre=\allowbreak{}translation; level=2 × genre=\allowbreak{}prose; level=2 × genre=\allowbreak{}science; level=2 × genre=\allowbreak{}seo; level=2 × genre=\allowbreak{}translation; level=3 × genre=\allowbreak{}prose; level=3 × genre=\allowbreak{}science; level=3 × genre=\allowbreak{}seo; level=3 × genre=\allowbreak{}translation & not identifiable \\
M2 &  &  & author & 302 & 1 & 1 & 244 &  &  \\
M2 &  &  & task (brief\_id) & 0 & 0 & 0 & 0 &  &  \\
M3 & 1882 & 11 of 11 & task (brief\_id) & 45 & 23 & 24 & 0 & none & identifiable \\
M3 &  &  & source & 128 & 1 & 1 & 85 &  &  \\
M3 &  &  & author & 306 & 1 & 1 & 244 &  &  \\
M3 &  &  & channel & 4 & 269 & 270 & 0 &  &  \\
\bottomrule
\end{tabular}
\begin{minipage}{\linewidth}\footnotesize\vspace{2pt}
\textit{Note.} Unit of analysis: each row describes one random effect of one model. Denominator: three preregistered models; 1079, 803 and 1882 observations. Data: series v2, diagnostics of 2026-07-29. Design matrix rank against the number of candidate columns; aliased levels were identified before fitting. Scale: none. Abbreviations: aliased level — a factor level linearly dependent on the remaining design columns; a single-observation cluster does not allow the random-effect variance to be estimated. Missing values: observations, rank, aliased levels and verdict refer to the model as a whole and appear in its first row; for M2 the random effect ``task'' has zero clusters because human texts receive no tasks. The table contains no effect estimates: the identifiability check preceded fitting, and some models never reached fitting. No correction for multiple comparisons was applied.
\end{minipage}
\end{sidewaystable}

%% file: tables/tab-S09-repeat-audit.tex
\begin{sidewaystable}[p]
\centering
\scriptsize
\caption{Audit of verbatim repeats: documents that passed the symmetric scan threshold}
\label{tab:s9}
\begin{tabular}{>{\RaggedRight\hspace{0pt}\arraybackslash}p{0.170\linewidth}>{\RaggedRight\hspace{0pt}\arraybackslash}p{0.079\linewidth}>{\RaggedRight\hspace{0pt}\arraybackslash}p{0.079\linewidth}>{\RaggedRight\hspace{0pt}\arraybackslash}p{0.079\linewidth}>{\RaggedRight\hspace{0pt}\arraybackslash}p{0.079\linewidth}>{\RaggedRight\hspace{0pt}\arraybackslash}p{0.079\linewidth}>{\RaggedRight\hspace{0pt}\arraybackslash}p{0.079\linewidth}>{\RaggedRight\hspace{0pt}\arraybackslash}p{0.079\linewidth}>{\RaggedRight\hspace{0pt}\arraybackslash}p{0.079\linewidth}}
\toprule
Document & Source & Genre & Repeat share & Maximum multiplicity & Longest repeated block & Modal offset & Mechanism & Verdict \\
\midrule
human\_news\_buriy\_2014\_0014 & buriy\_2014 & news & 0.320 & 2 & 12 & 44 & block repeat with constant offset & confirmed\_defect \\
human\_news\_buriy\_2014\_0045 & buriy\_2014 & news & 0.317 & 2 & 10 & 23 & block repeat with constant offset & confirmed\_defect \\
human\_news\_buriy\_2014\_0040 & buriy\_2014 & news & 0.262 & 2 & 8 & 33 & block repeat with constant offset & confirmed\_defect \\
human\_news\_buriy\_2014\_0024 & buriy\_2014 & news & 0.235 & 2 & 10 & 11 & block repeat with constant offset & confirmed\_defect \\
human\_seo\_madcats\_0005 & madcats & seo & 0.206 & 2 & 18 & 106 & scattered repeats & legitimate\_source\_property \\
human\_news\_buriy\_2014\_0042 & buriy\_2014 & news & 0.161 & 2 & 5 & 50 & block repeat with constant offset & confirmed\_defect \\
human\_seo\_drmax\_0013 & drmax & seo & 0.157 & 2 & 7 & 52 & block repeat with constant offset & confirmed\_defect \\
human\_seo\_devaka\_ru\_0019 & devaka\_ru & seo & 0.148 & 2 & 1 & 2 & scattered repeats & unresolved \\
human\_seo\_convertmonster\_0001 & convertmonster & seo & 0.138 & 4 & 1 &  & multiplicity above two & confirmed\_defect \\
human\_science\_spbgu\_0042 & spbgu & science & 0.135 & 8 & 9 & 40 & repeats on both sides of a structural heading & legitimate\_source\_property \\
human\_seo\_devaka\_ru\_0027 & devaka\_ru & seo & 0.132 & 2 & 10 & 5 & block repeat with constant offset & confirmed\_defect \\
human\_science\_spbgu\_0062 & spbgu & science & 0.129 & 5 & 5 & 33 & multiplicity above two & legitimate\_source\_property \\
human\_science\_spbgu\_0044 & spbgu & science & 0.118 & 2 & 11 & 36 & repeats on both sides of a structural heading & legitimate\_source\_property \\
human\_science\_urfu\_0039 & urfu & science & 0.117 & 2 & 3 & 39 & block repeat with constant offset & unresolved \\
human\_seo\_drmax\_0024 & drmax & seo & 0.106 & 3 & 4 & 5 & block repeat with constant offset & unresolved \\
human\_news\_buriy\_2014\_0048 & buriy\_2014 & news & 0.105 & 2 & 4 & 2 & scattered repeats & confirmed\_defect \\
human\_seo\_devaka\_ru\_0021 & devaka\_ru & seo & 0.103 & 2 & 6 & 7 & block repeat with constant offset & confirmed\_defect \\
\bottomrule
\end{tabular}
\begin{minipage}{\linewidth}\footnotesize\vspace{2pt}
\textit{Note.} Unit of analysis: document. Denominator: both parts of the corpus went through the scan, 1882 documents; selection threshold — share of verbatim repeated sentences above 0.1. Data: scan repeat-scan-v2, profiles prep-v4 and prep-v5. Repeat share is the sum of multiplicities of repeated sentences divided by the number of sentences in the document. Scale: higher means more heavily contaminated with repeats. Abbreviations: modal offset — the most frequent distance, in positions, between repeated sentences. Missing values: the table lists only documents above the threshold; the rest did not enter the scan and have no rows. No correction for multiple comparisons was applied.
\end{minipage}
\end{sidewaystable}

%% file: tables/tab-S10-blind-instability-cards.tex
\begingroup\scriptsize
\begin{longtable}{lp{2.5cm}p{2.5cm}p{3.3cm}p{3.1cm}}
\caption{Blind analysis of thirty instability cases: rater verdicts}\label{tab:s10} \\
\toprule
Card & Expected change present & Unexpected changes present & Change type & Verdict \\
\midrule
\endfirsthead
\toprule
Card & Expected change present & Unexpected changes present & Change type & Verdict \\
\midrule
\endhead
\bottomrule
\endfoot
BLIND-01 & yes & yes & duplication; paragraph structure & compound effect \\
BLIND-02 & yes & yes & duplication; paragraph structure & compound effect \\
BLIND-03 & yes & yes & duplication; paragraph structure & compound effect \\
BLIND-04 & yes & yes & duplication; paragraph structure & compound effect \\
BLIND-05 & not determinable & no & other & indeterminate \\
BLIND-06 & yes & yes & duplication; paragraph structure & compound effect \\
BLIND-07 & yes & yes & duplication; paragraph structure & compound effect \\
BLIND-08 & yes & yes & duplication; paragraph structure & compound effect \\
BLIND-09 & yes & yes & duplication; paragraph structure & compound effect \\
BLIND-10 & yes & yes & duplication; paragraph structure & compound effect \\
BLIND-11 & yes & yes & duplication; paragraph structure & compound effect \\
BLIND-12 & yes & yes & duplication; paragraph structure & compound effect \\
BLIND-13 & yes & yes & duplication; paragraph structure & compound effect \\
BLIND-14 & yes & yes & duplication; paragraph structure & compound effect \\
BLIND-15 & yes & no & paragraph structure; content loss & interpretable stress case \\
BLIND-16 & yes & yes & duplication; paragraph structure & compound effect \\
BLIND-17 & yes & yes & duplication; paragraph structure & compound effect \\
BLIND-18 & yes & yes & duplication; paragraph structure & compound effect \\
BLIND-19 & yes & yes & duplication; paragraph structure & compound effect \\
BLIND-20 & yes & yes & duplication; paragraph structure & compound effect \\
BLIND-21 & yes & yes & duplication; paragraph structure & compound effect \\
BLIND-22 & yes & yes & duplication; paragraph structure & compound effect \\
BLIND-23 & yes & yes & duplication; paragraph structure & compound effect \\
BLIND-24 & yes & yes & duplication; paragraph structure & compound effect \\
BLIND-25 & yes & yes & duplication; paragraph structure & compound effect \\
BLIND-26 & yes & yes & duplication; paragraph structure & compound effect \\
BLIND-27 & yes & yes & duplication; paragraph structure & compound effect \\
BLIND-28 & yes & no & paragraph structure; content loss & interpretable stress case \\
BLIND-29 & yes & no & markup & interpretable stress case \\
BLIND-30 & yes & yes & duplication; paragraph structure & compound effect \\
\end{longtable}
\begin{minipage}{\linewidth}\footnotesize\vspace{2pt}
\textit{Note.} Unit of analysis: card — a pair of texts before and after a transformation. Denominator: 30 cards selected by the rule ``sort by decision-flip rate, then by maximum probability shift''. Data: analysis instability-v1, blind presentation with seed 20260801. Verdicts are qualitative; the table contains no quantities. Scale: none. Abbreviations: compound effect — the transformation changed the text in more ways than declared; interpretable stress case — the change matches what was declared. Missing values: there was a single rater and the annotation was not duplicated, so inter-rater agreement was not computed. No correction for multiple comparisons was applied.
\end{minipage}
\endgroup

%% file: tables/tab-S11-preflight-invariants.tex
\begin{table}[htbp]
\centering
\small
\caption{Six invariants for admission to the v2 series computation}
\label{tab:s11}
\begin{tabular}{>{\RaggedRight\hspace{0pt}\arraybackslash}p{0.302\linewidth}>{\RaggedRight\hspace{0pt}\arraybackslash}p{0.302\linewidth}>{\RaggedRight\hspace{0pt}\arraybackslash}p{0.302\linewidth}}
\toprule
No. & Invariant & Result \\
\midrule
1 & the registry contains 1882 documents & passed \\
2 & none of the 34 excluded documents entered the matrix & passed \\
3 & raw and normalised values of the 1814 unchanged documents match v4 & passed, 114 282 values compared, zero discrepancies \\
4 & the percentile implementation reproduces v4 on the v4 composition & passed, 88 114 values \\
5 & the v5 percentiles are computed on the new reference sample & passed \\
6 & series v1 results are not overwritten & passed \\
\bottomrule
\end{tabular}
\begin{minipage}{\linewidth}\footnotesize\vspace{2pt}
\textit{Note.} The unit of analysis is the invariant. Denominator: six invariants; admission is granted only when all six pass. Data: admission manifest, revision 6. The table describes a gate; it contains no scores. Scale: none. Abbreviations: raw and normalized — the raw and normalised feature value; percentile implementation — computation of percentiles on the reference sample. Missing values: none. No correction for multiple comparisons was applied.
\end{minipage}
\end{table}

%% file: tables/tab-S12-manifest-revisions.tex
\begin{table}[htbp]
\centering
\small
\caption{Six revisions of the admission manifest and the reasons for moving to the next one}
\label{tab:s12}
\begin{tabular}{>{\RaggedRight\hspace{0pt}\arraybackslash}p{0.460\linewidth}>{\RaggedRight\hspace{0pt}\arraybackslash}p{0.460\linewidth}}
\toprule
Revision & What forced the move to the next one \\
\midrule
1 & edit to clf\_run.py under the amendment on two inner CV schemes \\
2 & procedure 1 did not create a child manifest — the chain was broken \\
3 & the pick\_c call in P2b kept the old number of return values and stopped the run \\
4 & the fairness and error-analysis reports labelled themselves series v1 and referenced the v1 verification file although they were verified against v2 \\
5 & error analysis retrained the model with its own GroupKFold instead of the registered scheme \\
6 & in effect \\
\bottomrule
\end{tabular}
\begin{minipage}{\linewidth}\footnotesize\vspace{2pt}
\textit{Note.} The unit of analysis is the manifest revision. Denominator: six revisions on 29 July 2026. Data: admission manifests of the v2 series. The table describes the history of the runs; it contains no numbers. Scale: none. Abbreviations: inner CV — nested cross-validation; GroupKFold — a split with non-overlapping groups. Missing values: a faulty manifest is not deleted but marked invalidated with a reason, so there are exactly as many rows as there were revisions. No correction for multiple comparisons was applied.
\end{minipage}
\end{table}

%% file: tables/tab-S13-reuse-gate.tex
\begin{table}[htbp]
\centering
\small
\caption{The parse reuse gate: reconciliation on the sentinel sample}
\label{tab:s13}
\begin{tabular}{>{\RaggedRight\hspace{0pt}\arraybackslash}p{0.460\linewidth}>{\RaggedRight\hspace{0pt}\arraybackslash}p{0.460\linewidth}}
\toprule
Parse stage & Reconciliation result on the sentinel sample \\
\midrule
Stanza & bitwise match on six documents \\
NER & bitwise match on six documents \\
embeddings & vector discrepancy up to 3.5·10⁻⁷, sem-v1 features up to 2.5·10⁻⁶, against a tolerance of 1·10⁻⁴ \\
\bottomrule
\end{tabular}
\begin{minipage}{\linewidth}\footnotesize\vspace{2pt}
\textit{Note.} The unit of analysis is the text parse stage. Denominator: six documents of the sentinel sample, three of each class. Data: the feature\_reuse\_gate gate, series v2. The reconciliation is built on recomputation from scratch and comparison against the cache. Scale: the smaller the discrepancy, the more reliable the reuse. Abbreviations: NER — named entity recognition; sem-v1 — the family of semantic features. Missing values: bitwise agreement never happens for embeddings: the cache was computed in batches across document boundaries, and the summation order differs. No correction for multiple comparisons was applied.
\end{minipage}
\end{table}

%% file: tables/tab-S14-index-decomposition.tex
\begin{table*}[htbp]
\centering
\scriptsize
\caption{Decomposition of the style index into stylistic and structural parts, two series}
\label{tab:s14}
\begin{tabular}{>{\RaggedRight\hspace{0pt}\arraybackslash}p{0.102\linewidth}>{\RaggedRight\hspace{0pt}\arraybackslash}p{0.102\linewidth}>{\RaggedRight\hspace{0pt}\arraybackslash}p{0.102\linewidth}>{\RaggedRight\hspace{0pt}\arraybackslash}p{0.102\linewidth}>{\RaggedRight\hspace{0pt}\arraybackslash}p{0.102\linewidth}>{\RaggedRight\hspace{0pt}\arraybackslash}p{0.102\linewidth}>{\RaggedRight\hspace{0pt}\arraybackslash}p{0.102\linewidth}>{\RaggedRight\hspace{0pt}\arraybackslash}p{0.102\linewidth}}
\toprule
Contrast & Series & Pairs & Common contribution & Format contribution & Sum & Frozen Δfull & Maximum discrepancy \\
\midrule
P3-P1 & v2 & 359 & +1.6740 & −3.4395 & −1.7655 & −1.7655 & 1.6e−04 \\
P2-P1 & v2 & 360 & −1.3927 & +4.6739 & +3.2812 & 3.2812 & 1.5e−04 \\
P3-P1 & v1 & 359 & +1.5458 & −3.4577 & −1.9118 & −1.9118 & 1.5e−04 \\
P2-P1 & v1 & 360 & −1.4879 & +4.6800 & +3.1921 & 3.1921 & 1.5e−04 \\
\bottomrule
\end{tabular}
\begin{minipage}{\linewidth}\footnotesize\vspace{2pt}
\textit{Note.} The unit of analysis is a text pair of the contrast. Denominator: 359 or 360 pairs depending on the contrast. Data: series v2 and v1; v1 is given as a stability check of the decomposition, not as the operative result. The decomposition is algebraic and exact: the sum of the contributions matches the frozen contrast to within 1.6·10⁻⁴, which is what the discrepancy column shows. Scale: the sign of a contribution is read under the convention ``higher means more AI-like''. Abbreviations: common — the stylistic feature categories; format — the structural category; Δfull — the full-score contrast from the frozen run. Missing values: none. No correction for multiple comparisons was applied.
\end{minipage}
\end{table*}

%% file: tables/tab-S15-t10-t11-feature-shift.tex
\begin{table}[htbp]
\centering
\small
\caption{Feature shift under shortening and expansion of the text: the withdrawn interpretation via volume}
\label{tab:s15}
\begin{tabular}{>{\RaggedRight\hspace{0pt}\arraybackslash}p{0.302\linewidth}>{\RaggedRight\hspace{0pt}\arraybackslash}p{0.302\linewidth}>{\RaggedRight\hspace{0pt}\arraybackslash}p{0.302\linewidth}}
\toprule
Feature & t10 shortening: median Δz (documents) & t11 expansion: median Δz (documents) \\
\midrule
M01 dispersion of similarity of adjacent sentences & +0.13 (51) & +7.86 (60) \\
S08 & −0.27 (46) & +4.53 (60) \\
L02 & +0.11 (42) & −3.02 (60) \\
L01 & +0.15 (44) & −1.45 (60) \\
R06 & −1.30 (23) & −1.39 (23) \\
R07 & −1.00 (22) & −1.00 (22) \\
P01 & −0.37 (24) & −0.34 (27) \\
F01 & −0.17 (15) & +0.10 (6) \\
\bottomrule
\end{tabular}
\begin{minipage}{\linewidth}\footnotesize\vspace{2pt}
\textit{Note.} The unit of analysis is a panel document. Denominator: 60 panel documents; in parentheses, the number of documents in which the feature shifted. Data: stress revision r4, procedure 1; standardisation on the 1882 documents of the v5 feature matrix. Median standardised shift Δz over the panel documents. Scale: the sign shows the direction of the feature shift, not of the decision. Abbreviations: Δz — shift in standard deviations of the feature; M01 — the standard deviation of the cosines of adjacent sentence pairs, not the mean similarity. Missing values: the interpretation ``both transformations change volume, so the decision collapses'' is withdrawn: the four format features shift identically under both, while the transformations diverge on M01. No correction for multiple comparisons was applied.
\end{minipage}
\end{table}

%% file: tables/tab-S16-limitations-groups.tex
\begin{table}[htbp]
\centering
\small
\caption{Twenty-six limitation items collapsed into the seven groups of the main text}
\label{tab:s16}
\begin{tabular}{>{\RaggedRight\hspace{0pt}\arraybackslash}p{0.302\linewidth}>{\RaggedRight\hspace{0pt}\arraybackslash}p{0.302\linewidth}>{\RaggedRight\hspace{0pt}\arraybackslash}p{0.302\linewidth}}
\toprule
Group & Items it folds in & What it tells the reviewer \\
\midrule
1. Identifiability of the design & 2, 12, 13 & two of three primary outcomes are unidentifiable, the negative control is not feasible, five holdouts are single-class and three of them coincide \\
2. Content validity of the features & 1, 17 & the index fully covers two of thirteen categories, fourteen features were not computed \\
3. Dependence on the procedure & 3, 4, 6, 7, 9, 10 & procedures 2–4 are exploratory, 1 and 2 are one family, net is not available for all, effects are small, Bonferroni correction, different contrast populations \\
4. Source and genre & 11, 14, 15, 19, 21 & the extraction defect is one-sided, missing values carry the source label, four machine clusters, level annotated by source by one person, the human part predates 2022 \\
5. Limited reproducibility of the judge & 8 & ICC(2,1) 0.477, fifteen values of the scale median, 85.9\% at two points \\
6. Absence of a prospective bridge set & 5 & all procedures on one corpus: alternative measurements, not independent samples \\
7. Boundaries of the stress test & 22, 23, 24, 25, 26 & tercile discretisation, format counters normalised to volume, denominator 10 of 16, composite volume transformations, the length sensitivity check rejected at preflight \\
\bottomrule
\end{tabular}
\begin{minipage}{\linewidth}\footnotesize\vspace{2pt}
\textit{Note.} The unit of analysis is a limitation group. Denominator: 26 items of the full list, seven groups in the main text. Data: the limitations list of 2026-07-29 with the additions of 31 July and 1 August. The table describes composition; it contains no run numbers. Scale: none. Abbreviations: ICC(2,1) — the intraclass correlation coefficient for random raters; bridge set — an independent sample for transferring the conclusions. Missing values: items 16, 18 and 20 did not enter the main text and remain in the full list only. No correction for multiple comparisons was applied.
\end{minipage}
\end{table}